\documentclass{article}

\usepackage{arxiv}

\usepackage[utf8]{inputenc}
\usepackage[T1]{fontenc}
\usepackage{hyperref}
\usepackage{url}
\usepackage{booktabs}
\usepackage{amsfonts}
\usepackage{nicefrac}
\usepackage{microtype}
\usepackage{xcolor}
\usepackage{amsmath}
\usepackage{amsthm}

\usepackage{graphicx}
\usepackage{enumitem}
\usepackage{wrapfig}
\usepackage{titletoc}
\usepackage{natbib}
\newcommand{\ci}[1]{{\tiny$\,\pm$#1}}

\definecolor{forgetdrop}{HTML}{1B5E20}
\definecolor{utilitydrop}{HTML}{B71C1C}

\title{LLM Unlearning Evaluation with TRIAGE}

\author{
 Danial Ataee \\
  Department of Computer Science\\
  University of Warwick\\
  \texttt{danial.ataee@warwick.ac.uk} \\
   \And
 Peter Triantafillou \\
  Department of Computer Science\\
  University of Warwick\\
  \texttt{p.triantafillou@warwick.ac.uk} \\
}

\begin{document}

\maketitle

\begin{abstract}
Large language models can memorize private or harmful information, motivating machine unlearning methods that remove targeted knowledge while preserving other capabilities. However, existing evaluations rely primarily on behavioral benchmarks, which assess \emph{whether} a model appears to forget but provide limited insight into \emph{how} unlearning changes the model or affects related knowledge. We introduce \textit{TRIAGE} (\textit{Tripartite Representation-internal Introspection for Adjacency Gap Evaluation}), a benchmark-agnostic evaluation framework for characterizing these changes. TRIAGE uses diagonal approximations of the Fisher information and Hessian to measure changes in parameter sensitivity and local curvature, and utilizes a Forget / \emph{Adjacent-Retain} / \emph{Generic-Retain} partition to quantify an \emph{adjacency gap} in semantically related knowledge. Based on the magnitude and distribution of these changes, TRIAGE further classifies each algorithm's update as \emph{no-op}, \emph{partially localized}, \emph{collateral
dominant}, or \emph{globally destructive}. Across 12 unlearning methods, four language models, and the WMDP, TOFU, and MUSE benchmarks, we find that methods with similar behavioral forgetting can produce substantially different internal changes and patterns of collateral damage. These signatures also vary across models and benchmarks, indicating that the effects of unlearning are not determined solely by the unlearning algorithm. TRIAGE can be applied alongside existing unlearning benchmarks to complement behavioral evaluation with a model-internal view of how unlearning reshapes the model's parameter space and affects retained knowledge. 
\footnote{Code available at \url{https://github.com/Dan-A2/Concept-Unlearning}.}
\end{abstract}

\section{Introduction}
\label{sec:introduction}

Today's LLMs are trained on massive internet-scale data that may include private, sensitive, erroneous, obsolete, poisoned, or harmful content. LLMs can reproduce copyrighted text \citep{karamolegkou2023copyright}, reveal personal information from training data \citep{carlini2021extracting, lukas2023analyzing}, and memorize content at rates that grow with model size \citep{carlini2023quantifying}. Training data can also be recovered after alignment \citep{nasr2023scalable}, while quantization can restore supposedly unlearned knowledge \citep{zhang2025catastrophic}. These risks, together with privacy regulation and copyright concerns \citep{henderson2023foundation}, have driven growing interest in machine unlearning as a way to suppress problematic knowledge.

Given the scale of modern LLMs, a key challenge is removing knowledge without full retraining \citep{shaik2024exploring, nguyen2025survey}. A diverse set of approaches has emerged, including gradient-based objectives \citep{jang2023knowledge, barbulescu2024each}, preference optimization \citep{rafailov2023direct, zhang2024negative, fan2026simplicity}, and parameter- or representation-level interventions \citep{bhaila2025soft, huu2024effects, cha2025towards}. However, recent work has questioned whether apparent forgetting reflects actual removal: forgotten knowledge can be recovered using benign data \citep{hu2025unlearning}, soft prompts \citep{schwinn2024soft}, similar facts \citep{deeb2024unlearning}, membership inference \citep{hayes2025inexact}, retain-set fine-tuning \citep{siddiqui2026dormant}, or even a single fine-tuning step \citep{bakman2026hair}. Probing attacks likewise reveal supposedly unlearned content \citep{burns2022discovering, qi2025safety}. Thus, model behavior alone may not establish successful unlearning.

This has motivated analysis of changes at the parameter and representation levels~\citep{hong2025intrinsic, xu2025unlearning, siddiqui2026dormant}. Yet existing evaluations primarily ask \emph{whether} unlearning succeeds, rather than \emph{what kind of change} produces the observed behavior. Similar forgetting scores may result from localized updates, negligible changes, behavioral suppression, or widespread collateral damage. In particular, existing evaluations do not directly assess whether changes are localized to the forget concept relative to semantically related knowledge that should be retained.

We introduce \textit{TRIAGE} to address this gap. It combines scalable diagonal approximations of the Fisher information and Hessian to characterize changes in parameter sensitivity and local curvature with a tripartite \emph{Forget}/\emph{Adjacent-Retain}/\emph{Generic-Retain} protocol that measures whether changes extend to semantically related knowledge. TRIAGE thus provides a model-internal view of unlearning that complements conventional behavioral evaluation.

\paragraph{Contributions.} We propose TRIAGE with four main contributions:

\begin{itemize}
\item \textbf{CoFi and CHess: model-internal diagnostics.}
We introduce scalable diagnostics based on diagonal approximations of the Empirical Fisher and Hessian, measuring changes in parameter sensitivity and local curvature.

\item \textbf{Tripartite evaluation with adjacent-retain control.}
We evaluate \emph{Forget}, \emph{Adjacent-Retain}, and \emph{Generic-Retain} content, integrating adjacent-retain evaluation~\citep{hu2025blur,cao2024rwku,chang2025retain} into model-internal and behavioral analysis.

\item \textbf{A four-way clustering of unlearning algorithms.}
We classify algorithms as \emph{no-op}, \emph{partially localized}, \emph{collateral dominant}, or \emph{globally destructive}, revealing failure modes that behavioral metrics can conflate across 12 methods, four LLMs, and three benchmarks.

\item \textbf{A test of benchmark validity.}
We use TRIAGE to assess whether unlearning benchmarks measure concept removal or primarily data untraining~\citep{triantafillou2026your}, identifying when structural change is expected to provide a meaningful signal for comparing unlearning methods.
\end{itemize}

\section{Related Work and TRIAGE Positioning}
\label{sec:related_work}

Machine unlearning predates large language models, where classical formulations often used retraining from scratch as a reference and evaluated success behaviorally~\citep{triantafillou2024we,sablayrolles2019white}. For modern LLMs, full retraining is generally infeasible, motivating a range of approximate unlearning methods and evaluation protocols. We organize recent work into three areas: unlearning algorithms, behavioral evaluation, and structural or parameter-space analysis. TRIAGE belongs to the third, but addresses a complementary question: beyond whether a model appears to forget, \emph{what kind of change does the unlearning update produce?}

\subsection{Unlearning Algorithms for LLMs}
LLM unlearning methods broadly operate by modifying the training objective or intervening directly in model representations. Objective-based approaches include Gradient Ascent, Selective Gradient Ascent, Gradient Difference, Direct Preference Optimization, Negative Preference Optimization, and SimNPO~\citep{jang2023knowledge,barbulescu2024each,rafailov2023direct,zhang2024negative,fan2026simplicity}. Representation-based approaches instead modify activations or parameters, including RMU and Adaptive-RMU, soft prompting, Fisher-initialized adapters, embedding-level interventions, membership-inference objectives, and stochastic perturbations~\citep{li2024wmdp,huu2024effects,bhaila2025soft,cha2025towards,spohn2025align,xu2025obliviate,tran2025tokens,huu2025improving}. Despite their differences, these methods are predominantly evaluated through behavioral metrics.

\paragraph{Two problem formulations.} Recent work distinguishes \emph{untraining}, which removes the influence of a specific forget set, from \emph{unlearning}, which aims to remove a broader concept or behavior~\citep{triantafillou2026your}. This distinction affects the expected structural footprint: removing a shallowly injected or well-generalized forget set may require little parameter change, whereas removing a distributed concept may require broader intervention. We use this distinction when interpreting our benchmarks (Section~\ref{sec:benchmarks}): WMDP primarily targets hazardous capabilities and is an
unlearning problem, while TOFU and MUSE-News are closer to untraining; MUSE-Books lies between the two removing memorized passages that nonetheless form a single coherent concept, the characters and plot of one fictional universe.

\subsection{Behavioral Benchmarks and Their Limits}

Existing benchmarks primarily evaluate model behavior. TOFU~\citep{maini2024tofu} studies synthetic biographies, WMDP~\citep{li2024wmdp} evaluates hazardous knowledge, and MUSE~\citep{shi2025muse} evaluates memorization, privacy leakage, utility, and related properties. BLUR~\citep{hu2025blur} is particularly relevant to TRIAGE because it examines behavior near the forget-retain boundary. However, these evaluations remain output-based, whereas TRIAGE examines whether the corresponding model update is localized in parameter space.

Behavioral evaluation can also be fragile: results may depend on prompting~\citep{feng2025existing}, forget and retain queries may be statistically dependent~\citep{thaker2025position}, and apparent forgetting can be reversed, bypassed, or reflect behavioral suppression rather than removal~\citep{shumailov2024ununlearning,cooper2026machine,lynch2024eight}. These limitations motivate complementary analysis of the unlearned model itself.

\subsection{Parameter- and Representation-Space Evaluation}
Recent work has begun to inspect unlearning beyond model outputs. ConceptVectors~\citep{hong2025intrinsic} studies localized concept-specific traces, while other work examines representation-level reversibility~\citep{xu2025unlearning} and resistance to relearning attacks~\citep{siddiqui2026dormant}. These approaches ask whether forgotten knowledge remains represented or recoverable.

TRIAGE addresses a different question: \emph{is the unlearning update localized?} Specifically, does it produce a larger change around the forget concept than around semantically adjacent knowledge that should be retained? To answer this, TRIAGE combines concept-level diagonal approximations of the Fisher information and Hessian with a tripartite \emph{Forget / Adjacent-Retain / Generic-Retain} evaluation. CoFi and CHess provide complementary measures of changes in parameter sensitivity and local curvature, while the adjacent-retain control exposes collateral changes that conventional Forget/Retain evaluations can obscure.

The use of Fisher and Hessian information is not itself novel; both have long-standing roles in optimization and have also been used in unlearning methods~\citep{cha2025towards,gu2024second}. Our contribution is to use tractable estimates of these quantities \emph{diagnostically and method-agnostically}, rather than as components of an unlearning algorithm, and to interpret them alongside a matched adjacent-retain control. This yields a model-internal characterization of unlearning updates that complements behavioral evaluation.

\section{The TRIAGE Evaluation Framework}
\label{sec:framework}

The TRIAGE framework has three components: two concept-level diagonal-curvature metrics, CoFi and CHess, that summarize change in parameter-space; a tripartite partition aiming to surface collateral damage on semantically-close knowledge; and a perplexity-based fluency check.

\subsection{Notation and setup}
\label{sec:notation}

Consider $\theta \in \mathbb{R}^P$, the parameters of an LLM where $P$ is the total number of parameters, and $\mathcal{L}(x;\theta) = -\sum_t \log p_\theta(x_t \mid x_{<t})$ its next-token loss on sequence $x$. Unlearning yields an updated set of parameters $\theta'$. Given a corpus $\mathcal{C}$ representing the knowledge targeted for removal, a concept or behaviour (e.g., WMDP capabilities), or specific memorized content (e.g., MUSE passages), we seek to quantifiably answer: \emph{how did unlearning shift the model's local geometry w.r.t. $\mathcal{C}$?}

Computing the full Fisher information or Hessian is infeasible at LLM scale, so we use diagonal estimates~\citep{kirkpatrick2017overcoming,lecun1989optimal,kunstner2019limitations,grosse2023studying}. We evaluate the target set $\mathcal{T}$ consisting of the seven attention and feed-forward projection matrices
$\{\texttt{q\_proj}, \texttt{k\_proj}, \texttt{v\_proj}, \texttt{o\_proj},
\texttt{gate\_proj}, \texttt{up\_proj}, \texttt{down\_proj}\}$ across all layers. These are the parameters modified by our LoRA-based \citep{hu2021lora} unlearning setup, so CoFi and CHess measure the footprint over the subspace on which the intervention acts. Let $N=|\mathcal{T}|$.

\subsection{Concept Fisher (CoFi)}
\label{sec:cofi}

The diagonal Empirical Fisher~\citep{kunstner2019limitations} on corpus $\mathcal{C}$ is
\begin{equation}
\hat{F}_i(\mathcal{C}; \theta) = \frac{1}{|\mathcal{C}|} \sum_{x \in \mathcal{C}} \left(\frac{\partial \mathcal{L}(x;\theta)}{\partial \theta_i}\right)^2, \quad i \in \mathcal{T}.
\label{eq:cofi-def}
\end{equation}
This is a per-parameter sensitivity measure: $\hat{F}_i$ is large (small) when parameter $i$ has a large (small) gradient signal under data from $\mathcal{C}$. The same object underpins EWC~\citep{kirkpatrick2017overcoming} for task-important weight identification and LoKU~\citep{cha2025towards} for unlearning-adapter initialization; we use it diagnostically rather than algorithmically. Equation~\ref{eq:cofi-def} is computed in log space with a small floor for numerical stability.

Stacking the per-parameter values over $\mathcal{T}$ gives the \textbf{Concept Fisher (CoFi)} vector $\hat{F}(\mathcal{C};\theta) = \big(\hat{F}_i(\mathcal{C};\theta)\big)_{i\in\mathcal{T}} \in \mathbb{R}^N$, i.e.\ the map $\hat{F}(\cdot;\theta)\colon \mathcal{C} \mapsto \mathbb{R}^N$. The norm $\|\cdot\|_F$ in Eq.~\ref{eq:cofi-shift} is the Euclidean norm of this $N$-vector (equivalently, the Frobenius norm of the per-layer diagonal blocks stacked together). Given a base model $\theta$ and an unlearned model $\theta'$, the relative CoFi shift on corpus $\mathcal{C}$ is
\begin{equation}
\Delta\mathrm{CoFi}(\mathcal{C}) = \frac{\big\| \hat{F}(\mathcal{C};\theta) - \hat{F}(\mathcal{C};\theta') \big\|_F}{\big\| \hat{F}(\mathcal{C};\theta) \big\|_F} \times 100\%,
\label{eq:cofi-shift}
\end{equation}
with a $1/\sqrt{N}$ normalization that makes Frobenius norms comparable across model scales. A large shift indicates that the model's parameter sensitivity to $\mathcal{C}$ changed; it is evidence of structural change, not proof that the underlying knowledge was removed. On retain corpora, a large shift indicates that the update reached content that was not intended to change.

\subsection{Concept Hessian (CHess)}
\label{sec:chess}

While CoFi captures first-order parameter sensitivity, it does not capture the local geometry of the loss landscape. Two models can have similar gradient-based sensitivity yet occupy different loss basins, ranging from sharp regions with strong structural commitment to flatter regions with weaker commitment. We therefore complement CoFi with a second-order diagnostic. The diagonal Hessian on corpus $\mathcal{C}$ is
\[
\hat{H}_i(\mathcal{C};\theta) =
\mathbb{E}_{x\sim\mathcal{C}}\left[
\frac{\partial^2 \mathcal{L}(x;\theta)}{\partial\theta_i^2}
\right].
\]
For its estimation, we use a Hutchinson-style diagonal estimator \citep{hutchinson1989stochastic} with Rademacher probes $z \in \{-1,+1\}^P$. Since $\mathbb{E}_z[z \odot Hz] = \mathrm{diag}(H)$, we approximate the Hessian-vector product via central finite differences of gradients:
\begin{equation}
\hat{H}_i(\mathcal{C};\theta) \approx \mathbb{E}_{z}\bigg[ z_i \cdot \frac{\nabla_i \mathcal{L}(\theta + \epsilon z) - \nabla_i \mathcal{L}(\theta - \epsilon z)}{2\epsilon} \bigg],
\label{eq:hutchinson}
\end{equation}
with $\epsilon = 10^{-3}$. Each probe requires two forward/backward passes. Four probes per batch are typically sufficient for stable estimates at the scales we study, and we apply the same log-space stabilization used for CoFi. A large $\Delta\mathrm{CHess}$ indicates that the local curvature around the model parameters has changed substantially with respect to that corpus.

Both estimators are well-established, and their diagonal forms provide an efficient way to perform parameter-level structural analysis at LLM scale \citep{kirkpatrick2017overcoming, matena2022merging, kunstner2019limitations, cha2025towards}. TRIAGE therefore uses diagonal approximations to retain useful sensitivity and curvature information without the computational cost of full matrices\citep{yao2020pyhessian, yao2021adahessian, ghorbani2019investigation, grosse2023studying}.

\subsection{The tripartite partition}
\label{sec:tripartite}

Standard evaluations contrast a forget corpus with generic retained data, making it difficult to distinguish changes on the target from collateral changes on semantically adjacent knowledge \citep{wei2026llms, ko2025probing}. Adjacent-retain controls have prior precedent in unlearning~\citep{hu2025blur,cao2024rwku,chang2025retain, amara2025erasing}; TRIAGE carries this control into parameter-space evaluation through three partitions:

\begin{itemize}
    \item \textbf{Forget} ($\mathcal{C}_F$): the corpus targeted for removal.
    \item \textbf{Adjacent-Retain} ($\mathcal{C}_A$): content from the same domain as $\mathcal{C}_F$ that must be preserved.
    \item \textbf{Generic-Retain} ($\mathcal{C}_G$): out-of-domain content, such as WikiText, that should be unaffected.
\end{itemize}

A method that unlearns $\mathcal{C}_F$ should create a model footprint that is largest on the forget corpus and much smaller on both retain corpora. Ideally, for either CoFi or CHess,
\begin{equation}
\Delta(\mathcal{C}_F) \;\gg\; \Delta(\mathcal{C}_A) \;\approx\; \Delta(\mathcal{C}_G),
\label{eq:ideal-ordering}
\end{equation}
or at minimum $\Delta(\mathcal{C}_F) > \Delta(\mathcal{C}_A)$ with a positive margin. A method that fails to localize impact, but bipartite evaluations would nonetheless regard as successful, instead exhibits
\begin{equation}
\Delta(\mathcal{C}_A) \;\approx\; \Delta(\mathcal{C}_F) \;\gg\; \Delta(\mathcal{C}_G),
\label{eq:collateral-ordering}
\end{equation}
a \emph{collateral-dominant} indicator.

We define
\[
\mathrm{AdjGap} = \Delta(\mathcal{C}_F) - \Delta(\mathcal{C}_A)
\]
the \emph{adjacency gap}. A positive gap indicates preferential movement on the forget set. A near-zero or negative gap indicates that the unlearning update has ``leaked'' into the adjacent-retain content. Construction criteria and pre-unlearning diagnostics for $\mathcal{C}_A$ are given in Appendix~\ref{app:adjacency}.

\subsection{Fluency check}
\label{sec:ppl}

We also report perplexity, as a fluency check, on each corpus, normalized to the base model:
\begin{equation}
\mathrm{PPL\text{-}ratio}(\mathcal{C}) = \mathrm{PPL}_{\theta'}(\mathcal{C}) / \mathrm{PPL}_{\theta}(\mathcal{C}).
\end{equation}
A ratio near one on retain corpora indicates preserved fluency, while an increase on the forget corpus is compatible with successful unlearning.

\subsection{The TRIAGE classes}
\label{sec:classes}

TRIAGE assigns each checkpoint to one of four classes using CoFi alone. CHess and perplexity provide complementary evidence but do not determine the class. Classification depends on update magnitude and the globality ratio:
\begin{equation}
\rho = \frac{\Delta\mathrm{CoFi}(\mathcal{C}_G)}
            {\max\!\big(\Delta\mathrm{CoFi}(\mathcal{C}_F),\,
                        \Delta\mathrm{CoFi}(\mathcal{C}_A)\big)},
\label{eq:rho}
\end{equation}
A small $\rho$ indicates that the footprint remains concentrated near the intervention domain, whereas $\rho$ near or above one indicates substantial global change. Applied in order:

\begin{itemize}
\item \textbf{No-op}: every per-corpus CoFi shift is below $1.5\%$. Nothing moved
anywhere, and the sign of any asymmetry is not interpretable.
\item \textbf{Globally destructive}: $\rho \geq \tau$ with
$\Delta\mathrm{CoFi}(\mathcal{C}_G) > 1\%$. General text is affected roughly as much as
the targeted domain.
\item \textbf{Partially localized}: $\Delta\mathrm{CoFi}(\mathcal{C}_A) <
\Delta\mathrm{CoFi}(\mathcal{C}_F)$. The footprint is concentrated on the forget corpus
relative to its semantic neighbourhood.
\item \textbf{Collateral dominant}: $\Delta\mathrm{CoFi}(\mathcal{C}_A) \geq
\Delta\mathrm{CoFi}(\mathcal{C}_F)$. Adjacent content moves at least as much as the
target.
\end{itemize}

We set $\tau=0.75$ from the empirical distribution rather than as a universal constant. Across 65 structurally active (method, model) pairs on WMDP, $\rho$ has a clear gap between $0.66$ and $0.85$, so any threshold in this interval gives the same partition. We first require sufficient update magnitude before interpreting the shape. The classes describe a footprint, and a footprint is a joint property of the algorithm,
the base model and the corpora, not a label attached to an algorithm. Methods move
between classes as any of the three changes, which we document across four models and
three benchmarks in Appendix~\ref{app:full-results}.

Overall, TRIAGE characterizes both the parameter-space footprint of unlearning and its localization relative to retained knowledge. It is therefore intended as a complement to behavioral evaluation rather than a replacement for it.

\section{Evaluating with TRIAGE}
\label{sec:experiments}

We evaluate TRIAGE across four LLMs and three benchmarks, asking whether unlearning produces a structural footprint, whether that footprint is localized, and how the model-internal view relates to behavioral evaluation.

\subsection{Experimental Setup}
\label{sec:setup}

\textbf{Models.} We evaluate Llama-3.2-3B, Zephyr-7B-$\beta$, Llama-3.1-8B, and Qwen3-32B.

\textbf{Benchmarks.} 
WMDP~\citep{li2024wmdp} targets hazardous bio and cyber knowledge; TOFU~\citep{maini2024tofu} evaluates fictitious biographies; and MUSE~\citep{shi2025muse} evaluates memorization of news and book content. For each benchmark, $\mathcal{C}_F$ and $\mathcal{C}_A$ follow the corresponding forget/retain construction, while $\mathcal{C}_G$ is WikiText.

\textbf{Methods.} We evaluate 12 recent algorithms: GA, GD, DPO, NPO, SimNPO, RMU, Adaptive-RMU, RSV, ATU, SPUL, Obliviate, and LoKU+FILA~\citep{jang2023knowledge,rafailov2023direct,zhang2024negative,fan2026simplicity,li2024wmdp,huu2024effects,huu2025improving,spohn2025align,bhaila2025soft,xu2025obliviate,cha2025towards}. We also evaluate eight RNA variants; because they do not materially alter the main structural or behavioral trends, they are reported in Appendix~\ref{app:full-results}. Hyperparameter details are in Appendix~\ref{app:hyperparams}.

\textbf{Configuration.} CoFi and CHess use the fixed diagnostic subset $\mathcal{T}$. For each corpus, three random subsets of 200 samples are evaluated and averaged with 95\% confidence intervals. We use batch sizes of 4 for CoFi and 8 for CHess with $K=4$ Hutchinson probes. For Qwen3-32B, CHess uses 50 samples, batch size 1, and one probe. Qwen3-32B is evaluated only on WMDP and MUSE-Books because a full TOFU and MUSE-News sweep exceeded our compute budget. Hyperparameters are in Appendix \ref{app:hyperparams}.

\subsection{Method Classification and the Adjacency Gap}
\label{sec:method-classification}

Table~\ref{tab:headline} reports CoFi, CHess, and PPL ratios for the 12 base methods on Llama-3.1-8B WMDP; full results are in Appendix~\ref{app:full-results}. Figure~\ref{fig:localization} visualizes the adjacency gap. Methods that move the forget concept more than its semantic neighborhood fall
below the diagonal and vice versa. The results reveal four distinct classes:

\begin{table}[t]
\centering
\small
\setlength{\tabcolsep}{3pt}
\caption{TRIAGE on Llama-3.1-8B WMDP. CoFi and CHess are relative shifts (\%), mean $\pm$
95\% CI over three random subsets of 200 samples; PPL ratio is unlearned/original
(1.0 = unchanged), geometric mean over the per-topic corpora. $\mathcal{C}_F$ and
$\mathcal{C}_A$ average the bio and cyber splits. Methods are grouped by their TRIAGE class.
Full per-topic values and the other three models are in Appendix~\ref{app:wmdp-results}.}
\label{tab:headline}
\begin{tabular}{l rrr rrr rrr}
\toprule
& \multicolumn{3}{c}{\textbf{CoFi (\%)}} & \multicolumn{3}{c}{\textbf{CHess (\%)}} & \multicolumn{3}{c}{\textbf{PPL ratio}} \\
\cmidrule(lr){2-4}\cmidrule(lr){5-7}\cmidrule(lr){8-10}
\textbf{Method} & $\mathcal{C}_F$ & $\mathcal{C}_A$ & $\mathcal{C}_G$ & $\mathcal{C}_F$ & $\mathcal{C}_A$ & $\mathcal{C}_G$ & $\mathcal{C}_F$ & $\mathcal{C}_A$ & $\mathcal{C}_G$ \\
\midrule
\multicolumn{10}{l}{\textit{No-op class}} \\
ATU           &  0.03\,\tiny{$\pm$0.00} &  0.03\,\tiny{$\pm$0.00} &  0.21\,\tiny{$\pm$0.17} &  22.3\,\tiny{$\pm$0.3} &  23.0\,\tiny{$\pm$1.2} &  26.7\,\tiny{$\pm$0.5} & 1.0 & 1.0 & 1.0 \\
Obliviate     &  0.08\,\tiny{$\pm$0.01} &  0.08\,\tiny{$\pm$0.01} &  0.93\,\tiny{$\pm$0.41} &  22.7\,\tiny{$\pm$0.4} &  23.5\,\tiny{$\pm$1.3} &  29.2\,\tiny{$\pm$0.6} & 1.1 & 1.1 & 1.7 \\
RSV           &  0.03\,\tiny{$\pm$0.00} &  0.03\,\tiny{$\pm$0.00} &  0.23\,\tiny{$\pm$0.18} &  22.3\,\tiny{$\pm$0.3} &  23.0\,\tiny{$\pm$1.2} &  26.9\,\tiny{$\pm$0.8} & 1.0 & 1.0 & 1.0 \\
\midrule
\multicolumn{10}{l}{\textit{Partially-localized class}} \\
RMU           &  7.05\,\tiny{$\pm$0.32} &  6.04\,\tiny{$\pm$1.19} &  0.24\,\tiny{$\pm$0.17} &  51.7\,\tiny{$\pm$1.6} &  43.7\,\tiny{$\pm$7.4} &  26.7\,\tiny{$\pm$0.5} & $10^{4}$ & $10^{2}$ & 1.0 \\
\midrule
\multicolumn{10}{l}{\textit{Collateral-dominant class}} \\
Adaptive-RMU  &  6.43\,\tiny{$\pm$0.18} &  6.62\,\tiny{$\pm$1.11} &  0.50\,\tiny{$\pm$1.06} &  48.9\,\tiny{$\pm$0.6} &  49.1\,\tiny{$\pm$1.2} &  27.7\,\tiny{$\pm$1.8} & $10^{5}$ & $10^{5}$ & 1.0 \\
NPO           &  6.06\,\tiny{$\pm$1.49} & 12.24\,\tiny{$\pm$5.11} &  1.85\,\tiny{$\pm$4.57} &  54.0\,\tiny{$\pm$3.2} &  65.0\,\tiny{$\pm$10.7} &  35.2\,\tiny{$\pm$9.0} & $10^{31}$ & $10^{30}$ & 1.1 \\
SimNPO        &  6.56\,\tiny{$\pm$1.49} & 11.62\,\tiny{$\pm$5.17} &  2.20\,\tiny{$\pm$4.01} &  55.1\,\tiny{$\pm$3.8} &  64.4\,\tiny{$\pm$9.6} &  35.1\,\tiny{$\pm$15.6} & $10^{30}$ & $10^{29}$ & 1.1 \\
DPO           &  7.60\,\tiny{$\pm$1.88} & 17.82\,\tiny{$\pm$2.09} &  0.22\,\tiny{$\pm$0.20} &  55.8\,\tiny{$\pm$3.1} &  71.2\,\tiny{$\pm$1.6} &  27.3\,\tiny{$\pm$0.2} & $10^{31}$ & $10^{28}$ & 1.0 \\
\midrule
\multicolumn{10}{l}{\textit{Globally destructive class}} \\
LoKU+FILA     & 14.21\,\tiny{$\pm$1.07} & 14.41\,\tiny{$\pm$1.96} & 12.20\,\tiny{$\pm$1.13} &  57.6\,\tiny{$\pm$0.6} &  55.2\,\tiny{$\pm$2.3} &  53.4\,\tiny{$\pm$2.4} & $10^{4}$ & $10^{4}$ & $10^{3}$ \\
GA            &  4.17\,\tiny{$\pm$2.65} &  6.86\,\tiny{$\pm$4.84} & 15.26\,\tiny{$\pm$16.1} &  56.0\,\tiny{$\pm$4.0} &  59.9\,\tiny{$\pm$3.0} &  63.4\,\tiny{$\pm$15.4} & $10^{72}$ & $10^{70}$ & $10^{3}$ \\
GD            &  2.81\,\tiny{$\pm$0.76} &  4.69\,\tiny{$\pm$2.83} & 11.40\,\tiny{$\pm$7.10} &  54.2\,\tiny{$\pm$2.5} &  56.6\,\tiny{$\pm$1.7} &  64.8\,\tiny{$\pm$9.9} & $10^{72}$ & $10^{70}$ & $10^{2}$ \\
SPUL          &  5.24\,\tiny{$\pm$0.28} &  5.54\,\tiny{$\pm$0.47} & 21.87\,\tiny{$\pm$8.34} &  57.0\,\tiny{$\pm$0.9} &  57.0\,\tiny{$\pm$1.6} &  66.2\,\tiny{$\pm$5.8} & $10^{33}$ & $10^{33}$ & $10^{29}$ \\
\bottomrule
\end{tabular}
\end{table}

\paragraph{No-op class.}
ATU, Obliviate, and RSV remain at the CoFi measurement floor ($0.03$--$0.93$), with little CHess or perplexity change. Matched behavioral evaluation likewise remains at the base-model level.

\paragraph{Partially-localized class.}
RMU is the only method in this class on Llama-3.1-8B WMDP, with $\Delta\mathrm{CoFi}(\mathcal{C}_F)=7.05\%$ versus $6.04\%$ on $\mathcal{C}_A$ and $0.24\%$ on $\mathcal{C}_G$. CHess shows the same ordering. The positive adjacency gap is therefore real but modest: even this relatively localized update substantially affects semantically adjacent content.

\paragraph{Collateral-dominant class.}
Adaptive-RMU, NPO, SimNPO, and DPO affect adjacent-retain at least as much as forget while leaving generic text relatively quiet. DPO provides the clearest example, with $17.82\%$ CoFi shift on $\mathcal{C}_A$ versus $7.60\%$ on $\mathcal{C}_F$. This pattern is invisible to a standard Forget--Generic Retain evaluation.

\paragraph{Globally destructive class.}
LoKU+FILA, GA, GD, and SPUL produce large changes on $\mathcal{C}_G$ ($11.40$--$21.87\%$), accompanied by severe perplexity inflation and large drops in general behavioral accuracy. Here the forget set is no longer the primary area of the update.

CHess is complementary rather than a second classifying signal: it captures changes in local curvature, while CoFi determines where the structural change is concentrated. Across models and benchmarks, class assignments are not fixed properties of algorithms. For example, RMU ranges from no-op on Qwen3-32B to partially localized on the Llama models and collateral dominant on Zephyr-7B-$\beta$, while NPO and SimNPO shift from collateral dominant on WMDP to partially localized on MUSE-Books. The full grid is reported in Appendix~\ref{app:full-results}.

\subsection{Beyond Methods: TRIAGE for Evaluating Benchmarks}
\label{sec:benchmarks}

TRIAGE also distinguishes benchmark behavior arising from the problem formulation. On TOFU, most runs produce near-zero CoFi shifts on Llama and Zephyr, with minimal CHess and near-unity PPL ratios \ref{app:tofu-results}; MUSE-News exhibits the same flat pattern \ref{app:muse-news-results}. These results are consistent with the \emph{untraining} formulation of these benchmarks~\citep{triantafillou2026your}: removing a specific fine-tuning footprint may legitimately require little structural change. Thus, a near-no-op structural result is ambiguous between correct untraining and failed concept removal and should be interpreted behaviorally.

MUSE-Books behaves differently \ref{app:muse-books-results}, producing clear structural footprints and recognizable localization patterns. For example, NPO yields $\Delta\mathrm{CoFi}(\mathcal{C}_F)=13.29\%$ versus $0.03\%$ on $\mathcal{C}_A$ on Llama-3.1-8b. The contrast with WMDP shows that localization depends jointly on the method, model, and distribution of the targeted knowledge rather than on the algorithm alone.

\subsection{Comparison with Matched Behavioral Evaluation}
\label{sec:agreement}

To test what TRIAGE adds, we compare it against a behavioural evaluation on the
\emph{same} tripartite partition. For WMDP, $\mathcal{C}_F$ is WMDP-Bio/Cyber,
$\mathcal{C}_A$ is the MMLU~\citep{hendrycks2020measuring} \texttt{virology},
\texttt{college\_biology} and \texttt{computer\_security} subsets, and $\mathcal{C}_G$ is
the remaining MMLU subsets. Writing $\delta_c$ for the relative accuracy drop on
partition $c$,
\[
\delta_c = 100 \cdot
\frac{\mathrm{Acc}_c^{\mathrm{base}} - \mathrm{Acc}_c^{\mathrm{post\text{-}UL}}}
     {\mathrm{Acc}_c^{\mathrm{base}}},
\qquad
\text{B-Gap} = \delta_F - \delta_A,
\]
the behavioural gap B-Gap is the direct analogue of
$\mathrm{AdjGap} = \Delta\mathrm{CoFi}(\mathcal{C}_F) -
\Delta\mathrm{CoFi}(\mathcal{C}_A)$. Table~\ref{tab:matched-tbe} puts them side by side;
full accuracies are in Appendix~\ref{app:tbe}.

\begin{table}[t]
\centering
\small
\setlength{\tabcolsep}{4pt}
\caption{Matched structural and behavioural adjacency gaps on Llama-3.1-8B WMDP,
computed on the same $\mathcal{C}_F$, $\mathcal{C}_A$ partition. ``Opposite'' marks
methods whose structural and behavioural gaps have opposite signs. ``Near zero'' marks
cases where both gaps are too small to interpret.}
\label{tab:matched-tbe}
\begin{tabular}{l rrr rrr l}
\toprule
& \multicolumn{3}{c}{\textbf{TRIAGE: CoFi (\%)}}
& \multicolumn{3}{c}{\textbf{TBE: relative drop (\%)}} & \\
\cmidrule(lr){2-4}\cmidrule(lr){5-7}
\textbf{Method} & $\mathcal{C}_F$ & $\mathcal{C}_A$ & AdjGap
& $\delta_F$ & $\delta_A$ & B-Gap & \textbf{Relation} \\
\midrule
\multicolumn{8}{l}{\textit{No-op class}} \\
ATU          & 0.03 & 0.03 & $0.00$    & 0.1  & 1.6  & $-1.5$  & near zero \\
Obliviate    & 0.08 & 0.08 & $0.00$    & 0.6  & 0.0  & $+0.6$  & near zero \\
RSV          & 0.03 & 0.03 & $0.00$    & $-0.5$ & 0.0 & $-0.5$ & near zero \\
\midrule
\multicolumn{8}{l}{\textit{Partially-localized class}} \\
RMU          & 7.05 & 6.04 & $+1.01$   & 48.4 & 27.9 & $+20.5$ & same sign \\
\midrule
\multicolumn{8}{l}{\textit{Collateral-dominant class}} \\
Adaptive-RMU & 6.43 & 6.62 & $-0.19$   & 53.9 & 49.2 & $+4.7$  & opposite \\
NPO          & 6.06 & 12.24 & $-6.18$  & 29.4 & 16.4 & $+13.0$ & opposite \\
SimNPO       & 6.56 & 11.62 & $-5.06$  & 32.0 & 22.9 & $+9.1$  & opposite \\
DPO          & 7.60 & 17.82 & $-10.22$ & 9.3  & 6.6  & $+2.7$  & opposite \\
\midrule
\multicolumn{8}{l}{\textit{Globally destructive class}} \\
LoKU+FILA    & 14.21 & 14.41 & $-0.20$ & 58.9 & 59.0 & $-0.1$  & near zero \\
GA           & 4.17 & 6.86 & $-2.69$   & 55.1 & 57.4 & $-2.3$  & same sign \\
GD           & 2.81 & 4.69 & $-1.88$   & 55.1 & 59.0 & $-3.9$  & same sign \\
SPUL         & 5.24 & 5.54 & $-0.30$   & 55.1 & 59.0 & $-3.9$  & same sign \\
\bottomrule
\end{tabular}
\end{table}

\begin{wrapfigure}{R}{0.45\textwidth}
\centering
\vspace{-10pt}
\includegraphics[width=\linewidth]{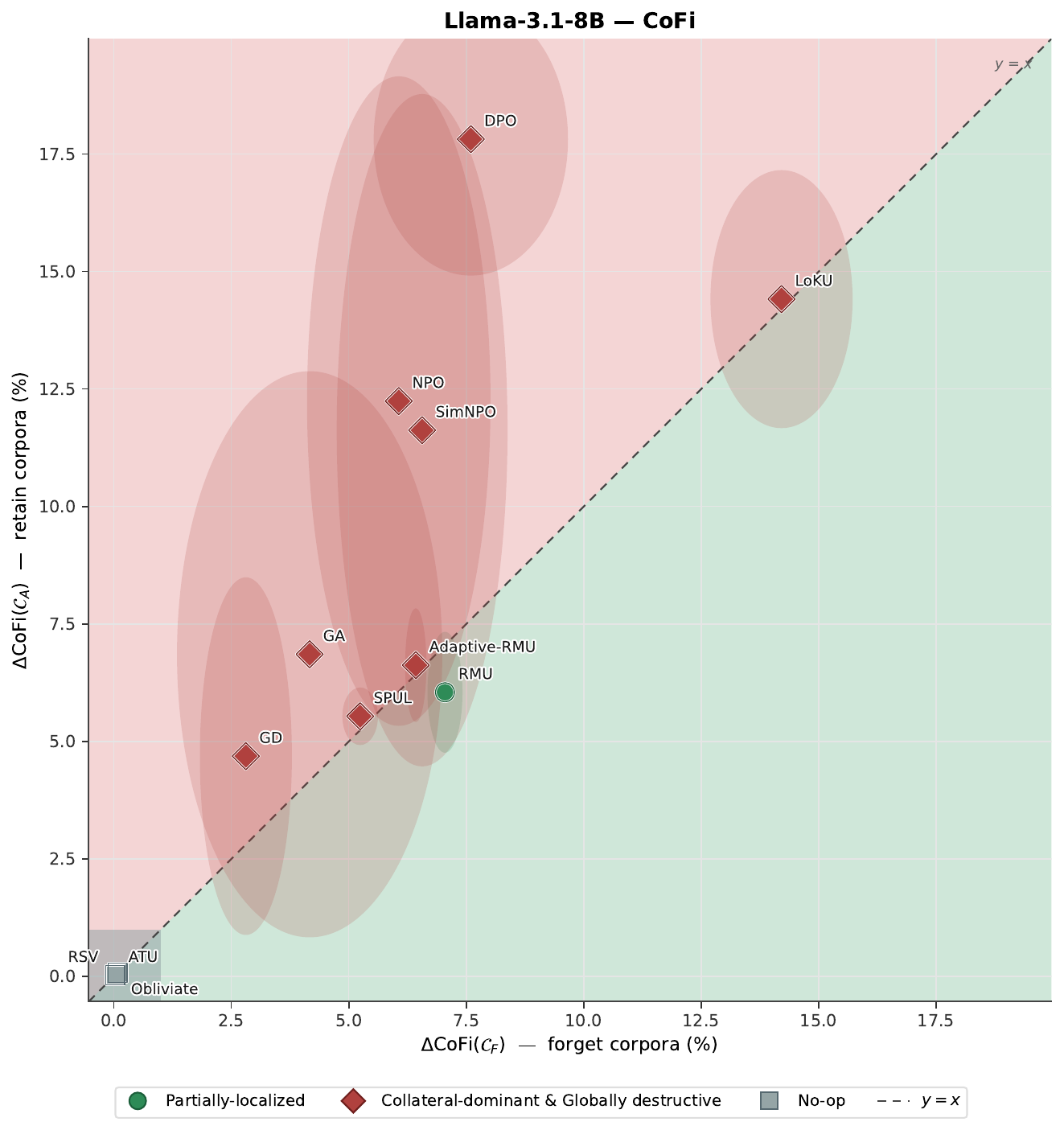}
\caption{Localization scatter on WMDP, Llama-3.1-8B. Each point is one method run. Methods below the diagonal target the forget concept more than the surrounding domain and vice versa. Other models are in Appendix~\ref{app:full-results}.}
\label{fig:localization}
\vspace{-10pt}
\end{wrapfigure}

\paragraph{The two signals are not interchangeable.}
The disagreement is concentrated in the collateral-dominant class: Adaptive-RMU, NPO, SimNPO, and DPO have positive behavioral gaps but negative structural gaps. Across all 12 methods, the two gaps are essentially uncorrelated ($r=-0.13$, $\rho\approx0$, $n=12$). In contrast, the generic-retain partition shows strong agreement: globally destructive methods exhibit both large CoFi shifts and severely degraded general accuracy. Thus, TRIAGE provides a signal that is not redundant with matched behavioral evaluation.

The matched evaluation produces four useful patterns. First, ATU, Obliviate, and RSV show little change under either evaluation, with TRIAGE explaining that the updates leave essentially no structural footprint. Second, LoKU+FILA, GA, GD, and SPUL reduce forget accuracy to chance while also damaging generic behavior, identifying collapse rather than targeted forgetting. Third, Adaptive-RMU, NPO, SimNPO, and DPO appear selective behaviorally but show collateral-dominant structural signatures; for DPO, $\Delta\mathrm{CoFi}(\mathcal{C}_A)=17.82\%$ despite only a $6.6\%$ adjacent accuracy drop. Fourth, the relearning attack shows that the collateral-dominant checkpoints are readily reversible, while the globally destructive checkpoints recover little because the model has already collapsed.

These comparisons do not establish that AdjGap predicts behavioral damage. When the two signals disagree, CoFi may be detecting internal non-selectivity without an immediate behavioral consequence, or behavioral evaluation may be missing latent adjacent damage. TRIAGE therefore serves as a diagnostic for identifying checkpoints that warrant further behavioral or relearning tests.

\subsection{Comparison with Other Structural Evaluations}
\label{sec:comparison}

Prior structural evaluations such as ConceptVectors, Reversibility, and Tamper-Resistance examine whether concepts remain recoverable or whether model weights have changed~\citep{hong2025intrinsic,xu2025unlearning,siddiqui2026dormant}. TRIAGE differs primarily in its matched adjacent-retain control, which allows structural localization to be distinguished from broad parameter deviation. The consistent tripartite evaluation also exposes cross-benchmark differences that isolated structural probes can miss; a detailed comparison is given in Appendix~\ref{app:structural_comparisons}.

Importantly, CoFi and CHess measure changes in parameter sensitivity and local curvature, not whether knowledge has actually been removed. A change may reflect removal, suppression, rerouting, or changes in how the knowledge is expressed. We therefore interpret TRIAGE by asking whether the structural change is concentrated on $\mathcal{C}_F$, extends to $\mathcal{C}_A$, or reaches $\mathcal{C}_G$.

\subsection{Response to a Relearning Attack}
\label{sec:relearning}

We ran a relearning attack applied to every Llama-3.1-8B WMDP checkpoint, measuring
\[
\Delta\mathrm{Acc}_F^{\mathrm{RL}}
= \mathrm{Acc}_F^{\mathrm{post\text{-}RL}} - \mathrm{Acc}_F^{\mathrm{post\text{-}UL}},
\]
averaged over the Bio and Cyber splits (details are in Appendix~\ref{app:hyperparams}). Higher values mean the forgotten content came
back. We interpret $\Delta\mathrm{CoFi}(\mathcal{C}_F)$ together with $\Delta\mathrm{CoFi}(\mathcal{C}_G)$ to distinguish target-conditioned change from global degradation.

\begin{table}[t]
\centering
\small
\setlength{\tabcolsep}{5pt}
\caption{The joint CoFi profile separates three relearning regimes on Llama-3.1-8B WMDP.
$\Delta\mathrm{Acc}_F^{\mathrm{RL}}$ is the accuracy recovered on the forget partition
under a fixed relearning budget.}
\label{tab:relearning-regimes}
\begin{tabular}{p{5.1cm} c c c}
\toprule
\textbf{Regime} & $\Delta\mathrm{CoFi}(\mathcal{C}_F)$ & $\Delta\mathrm{CoFi}(\mathcal{C}_G)$
& $\Delta\mathrm{Acc}_F^{\mathrm{RL}}$ \\
\midrule
No-op\newline \textit{\small ATU, RSV, Obliviate}
& $\leq 0.08\%$ & small & $-0.0017$ to $0.0056$ \\
\addlinespace
Active, not collapsed\newline \textit{\small RMU, Adaptive-RMU, NPO, SimNPO, DPO}
& large & small & $0.0586$ to $0.3020$ \\
\addlinespace
Globally destructive\newline \textit{\small GA, GD, LoKU+FILA}
& moderate--large & $11.40$--$15.26\%$ & $0.0000$ to $0.0116$ \\
\bottomrule
\end{tabular}
\end{table}

The resulting regimes are clear. No-op checkpoints show negligible structural change and negligible relearning response. Checkpoints with large $\Delta\mathrm{CoFi}(\mathcal{C}_F)$ but small $\Delta\mathrm{CoFi}(\mathcal{C}_G)$ recover substantial forgotten accuracy under a modest budget: RMU recovers $0.2738$ and Adaptive-RMU $0.3020$. Conversely, checkpoints with large generic CoFi shifts recover little because their general behavior has already collapsed. Thus, $\Delta\mathrm{CoFi}(\mathcal{C}_G)$ distinguishes apparent attack resistance from model collapse.

Across all 11 attacked checkpoints, $\Delta\mathrm{CoFi}(\mathcal{C}_F)$ correlates with relearning recovery (Spearman $\rho=0.65$, $p=0.030$). Among the eight non-destructive checkpoints the association is stronger (Pearson $r=0.76$, $p=0.028$; Spearman $\rho=0.68$, $p=0.062$). CHess shows a similar descriptive pattern ($r=0.72$), but we make no predictive claim for it. These results do not make a large CoFi shift a certificate of durable removal: rather, the shift contains information about how the checkpoint responds to subsequent intervention that point-in-time behavioral accuracy does not.

\section{Conclusion}
\label{sec:conclusion}

We introduce \textit{TRIAGE}, a benchmark-agnostic evaluation framework that complements behavioral unlearning evaluation with a model-internal view of parameter-space change. Combining CoFi and CHess with a \emph{Forget}/\emph{Adjacent-Retain}/\emph{Generic-Retain}
protocol, TRIAGE classifies each algorithm's update as \emph{no-op}, \emph{partially
localized}, \emph{collateral dominant}, or \emph{globally destructive}. Across 12 unlearning methods, four LLMs, and the WMDP, TOFU, and MUSE benchmarks, we find that similar behavioral forgetting can correspond to substantially different internal changes, with signatures varying across models and benchmarks. TRIAGE therefore provides a complementary diagnostic for distinguishing apparent forgetting from localized, collateral, or broadly destructive model changes.

The current evaluation also leaves important questions open. Since CoFi depends on the probing corpus, methods effective under unseen prompting styles may remain undetected; matched multi-format probing and paraphrased or extraction-based evaluation would extend its coverage. Such extensions require carefully designed tripartite benchmarks that preserve the target knowledge across partitions without introducing near-duplicates. Finally, because no ground truth exists for knowledge localization in parameter space, our null-intervention and noise-floor analyses establish that TRIAGE does not manufacture signal, but do not validate the inferred location itself.

\bibliography{references}
\bibliographystyle{plainnat}

\newpage

\appendix
\section{Appendix}

\section*{Appendix Contents}
\startcontents[appendix]
\printcontents[appendix]{}{1}{\setcounter{tocdepth}{2}}

\vspace{1em}

\section{Hyperparameters and Compute}
\label{app:hyperparams}

\textbf{Hardware.} All experiments were run on a single node with $3 \times$ NVIDIA A100
40GB GPUs, 8 CPU cores and 150GB RAM. Each unlearning run takes roughly 10--15 minutes
depending on method complexity, and each TRIAGE evaluation roughly 2--3 hours per
checkpoint for the three lighter models, and roughly 9--10 hours for Qwen3-32b model. The adjacent-retain diagnostics of
Appendix~\ref{app:adjacency} are far cheaper: they need one Fisher pass per corpus on the
reference model and run once per benchmark rather than once per checkpoint which takes less than 1 hour.

\textbf{LoRA configuration (the adapted subset).} All methods except SPUL use LoRA
adapters with $r = 16$, $\alpha = 32$ and dropout $0.05$, placed on the seven attention
and feed-forward projection families $\{\textit{q\_proj}, \textit{k\_proj},
\textit{v\_proj}, \textit{o\_proj}, \textit{gate\_proj}, \textit{up\_proj},
\textit{down\_proj}\}$ across all transformer layers. For these methods the optimizer is
AdamW with batch size 4 and 500 unlearning batches, per-method learning rates are given
in Table~\ref{tab:hyperparams}, and the saved checkpoint is the LoRA adapter.
SPUL~\citep{bhaila2025soft} instead runs in two phases. In phase~1, a LoRA adapter with
the same rank, scaling, dropout and target modules is fine-tuned on the forget and retain
data for 300 steps (AdamW, learning rate $10^{-4}$, $\beta_1 = 0.9$, $\beta_2 = 0.999$,
$\epsilon = 10^{-8}$, weight decay $0.01$, no scheduler, gradient checkpointing enabled)
and then merged into the base weights. In phase~2 the merged model is frozen and only a
P-tuning prompt encoder is trained: 20 virtual tokens reparameterized by an MLP with
hidden size 128, learning rate $10^{-3}$, batch size 4, the same AdamW settings and seed
42. The phase-2 objective is
$\mathcal{L} = -\mathrm{CE}(\mathcal{D}_{\text{forget}})
 + \alpha\,\mathrm{CE}(\mathcal{D}_{\text{retain}})
 + \beta\,\mathrm{KL}\big(p_{\text{prompted}} \,\|\, p_{\text{frozen}}\big)$,
with the KL term computed on retain data between the prompted model and the frozen,
unprompted model, and $\alpha = \beta = 1.0$ ($\alpha$ applied per topic). Phase~2 runs for
at most 500 steps. The step count is clipped to the number of batches the forget corpus
provides, so shorter corpora, MUSE-Books in particular, train for fewer steps. The saved
SPUL checkpoint is therefore a prompt-encoder adapter rather than a LoRA adapter. In the
released code, SPUL's phase-1 rank, scaling and dropout are fixed defaults, and only the
target modules can be set from the command line. Sequence length is set by the benchmark
rather than the method: 512 tokens for WMDP-Bio, TOFU and MUSE, and 768 for WMDP-Cyber,
with right-side truncation. Apart from SPUL's two-phase procedure, all methods share the
same projection subset, batch size and optimizer family.

\textbf{Diagnostic subset $\mathcal{T}$ (the measured subset).} CoFi and CHess are
computed over the same seven projection families across all layers. The adapted and
diagnostic subsets coincide here but are conceptually separate: the LoRA configuration
fixes where the update acts, $\mathcal{T}$ fixes where the footprint is measured.
$\mathcal{T}$ is held constant across every method, model and benchmark, so CoFi and
CHess values are comparable within it.

\textbf{Per-method hyperparameters} (Table~\ref{tab:hyperparams}).

\begin{table}[h]
\centering
\small
\setlength{\tabcolsep}{5pt}
\caption{Per-method hyperparameters. \texttt{lr} is learning rate; $\alpha$ is the
per-topic retain-loss weight; $\beta$ is the temperature for preference-style methods;
\texttt{coeff} is the steering coefficient where applicable.}
\label{tab:hyperparams}
\begin{tabular}{l l l l l}
\toprule
\textbf{Method} & \textbf{lr} & $\boldsymbol{\alpha}$ \textbf{(retain weight)} & \textbf{Other} & \textbf{Retain loss} \\
\midrule
GA              & $1\text{e-}5$ & 1.0  & ---                          & KL \\
GD              & $1\text{e-}5$ & 1.0  & ---                          & KL \\
NPO             & $2\text{e-}5$ & 50.0 & $\beta = 0.1$                & KL \\
SimNPO          & $2\text{e-}5$ & 50.0 & $\beta = 0.1$                & KL \\
DPO             & $5\text{e-}5$ & 50.0 & $\beta = 0.1$                & KL \\
RMU             & $5\text{e-}5$ & 1200 & coeff $=6.5$                 & MSE on retain activations \\
Adaptive-RMU    & $5\text{e-}5$ & 1200 & coeff $=1.0$, scale $=3.0$   & MSE on retain activations \\
RSV             & $5\text{e-}5$ & 5000 & coeff $=10$                  & MSE on retain activations \\
ATU             & $5\text{e-}5$ & 1.0  & threshold $=0.5$, align lr $=10^{-4}$ & KL to frozen model \\
Obliviate       & $1\text{e-}5$ & ---  & $\Lambda_1=0.2$, $\Lambda_2=0.7$ & MSE distill + CE retain \\
SPUL            & $1\text{e-}3$ & 1.0  & 20 prompt tokens, $\beta = 1.0$ & CE retain + KL retain \\
LoKU+FILA       & $1\text{e-}4$ & 1.0  & FILA Fisher init             & IHL forget + CE retain \\
\bottomrule
\end{tabular}
\end{table}

\textbf{RNA noise variants.} For methods admitting a stochastic-perturbation variant
(suffix $\nu$), we use $\nu = 0.001$ following \citet{huu2025improving}.

\textbf{TRIAGE evaluation parameters.} For each corpus we draw three random subsets of
200 samples, evaluate each metric independently on each, and report the mean with a 95\%
confidence interval. CoFi uses max sequence length 1024 and batch size 4; CHess uses max
sequence length 512, batch size 8, $K = 4$ Hutchinson probes per batch, and
finite-difference $\epsilon = 10^{-3}$. Perplexity uses a sliding window of 2048 tokens
with a stride of 512, capped at 50{,}000 tokens per corpus. Three exceptions apply. The Qwen3-32B is the only model loaded using 4-bit quantization for our calculations to fit in GPU VRAM, and CHess is evaluated on subsets of 50 samples with batch size 1 and a single
Hutchinson probe for this model to keep the computations tractable. For MUSE-Books, the three subsets coincide because the splits contain only 4, 12 and 13 documents, so we report point estimates there and give a confidence interval only for the WikiText column; see Appendix~\ref{app:muse-books-results} for the full explanation.

\textbf{Behavioral evaluation parameters.} The matched Tripartite Behavioural Evaluation
of Section~\ref{sec:agreement} is run through \texttt{lm-eval-harness} with batch size 8,
on the same partitions used by TRIAGE: $\mathcal{C}_F$ is WMDP-Bio and WMDP-Cyber,
$\mathcal{C}_A$ is the MMLU \texttt{virology}, \texttt{college\_biology} and
\texttt{computer\_security} subsets, and $\mathcal{C}_G$ is the remaining MMLU subsets.

\textbf{Relearning attack.} The fixed-budget attack of Section~\ref{sec:relearning}
fine-tunes each unlearned checkpoint on a small slice of the forget corpus and measures
how much forget-partition accuracy returns. We use AdamW at learning rate $1\text{e-}5$
for 5 epochs over 50 forget documents, with effective batch size 4, a linear schedule
with $3\%$ warmup followed by linear decay, and a global cap of 500 optimizer steps. The
attack updates LoRA adapters with $r = 16$, $\alpha = 32$ and dropout $0.05$ on the same
attention and feed-forward projections used during unlearning, so the attacker operates
in the same parameter subspace as the unlearning method.

These settings follow the conventions of the robust-unlearning literature, where
recovery is typically obtained at learning rates between $1\text{e-}5$ and $5\text{e-}5$
\citep{lynch2024eight, deeb2024unlearning}; $1\text{e-}5$ is the
conservative end of that range. The point of a relearning attack is that it is cheap:
limited data and few steps, so that recovery reflects how easily the forgotten behaviour
returns rather than how much compute was spent retrieving it. The identical budget is
applied to every checkpoint, which is what makes $\Delta\mathrm{Acc}_F^{\mathrm{RL}}$
comparable across methods. We report a single operating point and do not sweep attack
strength; a stronger attack would recover more from every checkpoint, and our claim
concerns the ordering across checkpoints at a fixed budget rather than any absolute
level of robustness.

\section{Sensitivity and Robustness of CoFi and CHess}
\label{app:sensitivity}

We vary three estimation settings one at a time: the number of concept samples $n$, the
Hutchinson probe count $K$ (CHess only), and the diagnostic subspace. The methods are
ATU, RMU and NPO, one per structurally distinct class --- no-op, partially localized and
collateral dominant. The globally destructive regime is not covered here. All numbers
are Llama-3.1-8B on the \emph{bio corpora only}, whereas Table~\ref{tab:headline} averages
$\mathcal{C}_F$ and $\mathcal{C}_A$ over bio and cyber, so absolute values should be
compared within this appendix rather than against the main table.

Three factors play different roles and we do not treat them as one grid. Sample size and
probe count control estimator precision at a fixed operating point ($n = 200$, $K = 4$),
chosen once and never tuned per method; this is the axis on which we do claim invariance.
The diagnostic subspace defines \emph{which} structural object is being measured, so
attention-only and FFN-only diagnostics are different measurements rather than noisier
readings of one quantity; we report subspace dependence as a characterisation, not as a
validity test. The diagonal approximation is the scalable design of the diagnostic rather
than a tunable knob.

\subsection{CoFi}

\begin{table}[h]
\centering
\small
\setlength{\tabcolsep}{6pt}
\caption{CoFi relative shift (\%) and class margin against sample size $n$, diagnostic
subset attention $+$ FFN. Margin $=$ CoFi(forget) $-$ CoFi(retain); its sign determines
the class.}
\label{tab:cofi-samplesize}
\begin{tabular}{l l rrr}
\toprule
\textbf{Method} & \textbf{Corpus} & $n = 52$ & $n = 100$ & $n = 200$ \\
\midrule
ATU & {[}F{]} bio-forget & 0.0053 & 0.0044 & 0.0052 \\
ATU & {[}R{]} bio-retain & 0.0147 & 0.0126 & 0.0106 \\
NPO & {[}F{]} bio-forget & 12.6013 & 12.0699 & 12.3285 \\
NPO & {[}R{]} bio-retain & 14.7573 & 15.1277 & 15.3290 \\
RMU & {[}F{]} bio-forget & 5.6918 & 5.6428 & 5.6584 \\
RMU & {[}R{]} bio-retain & 4.8750 & 4.8630 & 4.8058 \\
\midrule
\multicolumn{5}{l}{\textit{Class margin}} \\
ATU & & $-0.0094$ & $-0.0083$ & $-0.0054$ \\
NPO & & $-2.1559$ & $-3.0578$ & $-3.0005$ \\
RMU & & $+0.8168$ & $+0.7798$ & $+0.8526$ \\
\bottomrule
\end{tabular}
\end{table}

\begin{table}[h]
\centering
\small
\setlength{\tabcolsep}{6pt}
\caption{CoFi relative shift (\%) and class margin by diagnostic subspace, $n = 200$.}
\label{tab:cofi-subspace}
\begin{tabular}{l l rrr}
\toprule
\textbf{Method} & \textbf{Corpus} & attn & FFN & both \\
\midrule
ATU & {[}F{]} bio-forget & 0.0101 & 0.0029 & 0.0052 \\
ATU & {[}R{]} bio-retain & 0.0196 & 0.0069 & 0.0106 \\
NPO & {[}F{]} bio-forget & 16.6078 & 11.0684 & 12.3285 \\
NPO & {[}R{]} bio-retain & 19.5909 & 14.1260 & 15.3290 \\
RMU & {[}F{]} bio-forget & 9.4016 & 4.3122 & 5.6584 \\
RMU & {[}R{]} bio-retain & 8.1648 & 3.5668 & 4.8058 \\
\midrule
\multicolumn{5}{l}{\textit{Class margin}} \\
ATU & & $-0.0095$ & $-0.0040$ & $-0.0054$ \\
NPO & & $-2.9831$ & $-3.0576$ & $-3.0005$ \\
RMU & & $+1.2368$ & $+0.7454$ & $+0.8526$ \\
\bottomrule
\end{tabular}
\end{table}

\paragraph{Estimation budget.}
Cutting the budget roughly fourfold ($200 \to 52$) has little effect. NPO moves by at
most $0.6$ percentage points ($\approx 4\%$ relative), RMU by at most $0.07$ points
($< 1.3\%$), and ATU stays at the $10^{-2}$ point level, orders of magnitude below the
structurally active methods. All three class assignments hold at every budget. The least
stable entry is the \emph{magnitude} of NPO's margin at the smallest budget ($-2.16$ at
$n = 52$ against $-3.00$ at $n = 200$, a $28\%$ deviation), which is the reduction in
estimator precision we would expect; the sign, and therefore the class, is unaffected.

\paragraph{Diagnostic subspace.}
Absolute CoFi values depend on the subspace, and monotonically so: attention-only shifts
run roughly $1.5$--$2.2\times$ the FFN-only shifts for both active methods
(NPO $16.61$ against $11.07$; RMU $9.40$ against $4.31$), with the joint subset in
between. These are different structural subspaces rather than noisier readings of one
quantity, which is why we fix and state $\mathcal{T}$ explicitly. The classification-relevant quantity
survives: the margin sign is identical in all nine cells, and for NPO the margin is
nearly invariant ($-2.98$, $-3.06$, $-3.00$) despite the spread underneath it. A reader
should not compare absolute CoFi values across subspaces; the class assignment does carry
across.

\subsection{CHess}

CHess is not used to assign classes --- CoFi determines where an update is concentrated,
while CHess measures how much the loss landscape moved --- so the question here is
whether CHess values and the separation between regimes stay stable. We report the
forget corpus only, for the same reason.

\begin{table}[h]
\centering
\small
\setlength{\tabcolsep}{6pt}
\caption{CHess relative shift (\%) on Bio-Forget under three estimation settings.
Left: sample size at $K=4$. Middle: probe count at $n=200$. Right: diagnostic subspace at
$n=200$, $K=8$. All on the attention $+$ FFN subset unless stated.}
\label{tab:chess-sensitivity}
\begin{tabular}{l rrr @{\hspace{10pt}} rrr @{\hspace{10pt}} rrr}
\toprule
& \multicolumn{3}{c}{\textbf{Sample size}} & \multicolumn{3}{c}{\textbf{Probe count}}
& \multicolumn{3}{c}{\textbf{Subspace}} \\
\cmidrule(lr){2-4}\cmidrule(lr){5-7}\cmidrule(lr){8-10}
\textbf{Method} & $n{=}52$ & $n{=}100$ & $n{=}200$ & $K{=}2$ & $K{=}4$ & $K{=}8$
& attn & FFN & both \\
\midrule
ATU & 7.6525  & 7.6096  & 10.3733 & 7.4355  & 11.0664 & 11.2539 & 10.9362 & 11.3282 & 11.2539 \\
NPO & 62.2578 & 61.5353 & 61.2569 & 61.6975 & 61.2370 & 60.7847 & 64.5920 & 59.8425 & 60.7847 \\
RMU & 48.4860 & 48.4151 & 48.1047 & 48.2289 & 48.0467 & 47.6811 & 51.2432 & 46.7930 & 47.6811 \\
\bottomrule
\end{tabular}
\end{table}

\paragraph{Estimation budget.}
CHess is stable for the structurally active methods. Reducing $n$ fourfold changes NPO by
$1.0$ percentage point ($1.6\%$ relative) and RMU by $0.4$ points ($0.8\%$). The
separation between regimes holds at every budget: the no-op checkpoint stays an order of
magnitude below the active methods, and RMU and NPO stay clearly apart from each other.

\paragraph{Probe count.}
Going from $K = 2$ to $K = 8$ changes NPO by $0.9$ points ($1.5\%$ relative) and RMU by
$0.5$ points ($1.1\%$), so our operating point of $K = 4$ is not a boundary case: halving
and doubling the budget both land in the same place. ATU is the exception and is
informative about the estimator rather than the method: its value is unstable at
$K = 2$ ($7.44$) and settles from $K = 4$ onward ($11.07$, $11.25$), indicating that two
probes are not enough for a checkpoint whose true shift is near the floor.

\paragraph{Diagnostic subspace.}
Absolute CHess values depend on the subspace but noticeably less than CoFi does.
Attention-only runs about $1.08\times$ FFN-only for NPO ($64.59$ against $59.84$) and
$1.10\times$ for RMU ($51.24$ against $46.79$), against the $1.5$--$2.2\times$ spread seen
for CoFi. This is consistent with the reading in Section~\ref{sec:chess}: CHess tracks
the overall curvature of the loss basin, which shows up similarly across parameter
subsets, whereas CoFi is built to localize where the change sits and is therefore more
sensitive to which parameters are being examined.

\paragraph{On the diagonal approximation.}
A full Fisher or Hessian reference is infeasible at this scale and would not give a
practical diagnostic in any case; block-diagonal or low-rank alternatives would
substitute different structural assumptions rather than supply ground truth. We therefore
validate the diagonal form empirically rather than against an intractable reference:
through its non-redundancy with matched behavioural evaluation
(Section~\ref{sec:agreement}), its association with relearning response
(Section~\ref{sec:relearning}), and the stability reported above.

\section{Constructing and Validating the Adjacent-Retain Corpus}
\label{app:adjacency}

AdjGap is only as meaningful as the corpus it is computed on. We set out three criteria
for constructing $\mathcal{C}_A$, together with two diagnostics that need only the
reference model and the corpus definitions, so that the quality of the adjacent corpus
can be checked independently of any unlearning method.

\paragraph{1. Deletion-scope disjointness.}
No item in $\mathcal{C}_A$ may require information whose removal was requested. If it
does, degradation on $\mathcal{C}_A$ is intended behaviour rather than collateral
movement, and a negative AdjGap is the correct outcome rather than a warning. This is a
semantic property of the deletion specification. It has to be established by benchmark
construction or annotation and cannot be diagnosed after the fact from the model.

\paragraph{2. Reference-model headroom.}
The pre-unlearning model must perform well enough above chance on $\mathcal{C}_A$ for the
retain score to have somewhere to fall; otherwise a preserved $\mathcal{C}_A$ tells us
nothing. On WMDP, where chance is $0.25$, Llama-3.1-8B scores $0.775$ and $0.750$ on the
Bio- and Cyber-adjacent sets and Zephyr-7B-$\beta$ scores $0.525$ and $0.650$. For TOFU
and MUSE the analogous requirement is a finite, meaningful reference-model perplexity on
$\mathcal{C}_A$.

\paragraph{3. Model-side adjacency evidence.}
Whether the candidate $\mathcal{C}_A$ shares more concept-conditioned high-Fisher
parameter support with $\mathcal{C}_F$ than $\mathcal{C}_G$ does. This one needs care:
Fisher overlap is itself model- and representation-dependent, and a semantically
legitimate $\mathcal{C}_A$ may show weak measured overlap. It does not replace semantic
adjacency. What it determines is how confidently a structural AdjGap can be read as
localization \emph{with respect to that particular model}.

The first two criteria establish semantic and behavioural validity; the third is a
model-side diagnostic rather than a condition for a valid adjacent-retain set. We test
adjacency through two independent mechanisms, one content-only and one model-internal.
They can disagree, and where they do the disagreement is informative.

\paragraph{Diagnostic (a): semantic similarity.}
Mean pairwise cosine similarity under \texttt{all-MiniLM-L6-v2}, a widely used default
sentence embedder. This is model-independent and computed once per benchmark. The test is
$\mathrm{sim}(\mathcal{C}_F, \mathcal{C}_A) > \mathrm{sim}(\mathcal{C}_A, \mathcal{C}_G)$:
the adjacent set must sit closer to the forget target than to generic text.

\begin{table}[h]
\centering
\small
\setlength{\tabcolsep}{6pt}
\caption{Semantic similarity between partitions. All three benchmarks pass.
Absolute values depend on writing style and corpus length and are not comparable across
benchmarks; only the within-benchmark ordering is meaningful.}
\label{tab:adjacency-semantic}
\begin{tabular}{l rrr r c}
\toprule
\textbf{Benchmark} & $\mathrm{sim}(\mathcal{C}_F,\mathcal{C}_A)$
& $\mathrm{sim}(\mathcal{C}_F,\mathcal{C}_G)$
& $\mathrm{sim}(\mathcal{C}_A,\mathcal{C}_G)$
& \textbf{Ratio} & \textbf{Passes} \\
\midrule
WMDP        & 0.1012 & 0.0013 & 0.0080 & $12.6\times$ & Yes \\
TOFU        & 0.2424 & 0.0390 & 0.0419 & $5.8\times$  & Yes \\
MUSE-Books  & 0.3540 & 0.0477 & 0.0623 & $5.7\times$  & Yes \\
MUSE-News   & 0.1299 & 0.0343 & 0.0360 & $3.6\times$  & Yes \\
\bottomrule
\end{tabular}
\end{table}

All four satisfy the criterion by a clear margin: $\mathrm{sim}(\mathcal{C}_F,
\mathcal{C}_A)$ is $3.6$ to $12.6$ times larger than $\mathrm{sim}(\mathcal{C}_A,
\mathcal{C}_G)$, and consistently larger than $\mathrm{sim}(\mathcal{C}_F,
\mathcal{C}_G)$. $\mathcal{C}_A$ sits where it is meant to, between the forget and
generic corpora.

\paragraph{Diagnostic (b): reference-model Fisher overlap.}
For each corpus we compute the diagonal empirical Fisher of the reference model $\theta$,
take the top-$k$ parameter indices by Fisher mass, and report
\[
\mathrm{overlap}(X, Y) = \frac{|\mathrm{top}_k(X) \cap \mathrm{top}_k(Y)|}{k},
\qquad k \in \{10^3, 10^4, 10^5\},
\]
summarising model-side adjacency by the margin
$M_k = \mathrm{overlap}(\mathcal{C}_F, \mathcal{C}_A) -
\mathrm{overlap}(\mathcal{C}_F, \mathcal{C}_G)$. The reference is the pretrained base
model for WMDP and the finetuned checkpoint for TOFU and MUSE, matching the model against
which CoFi shifts are measured. We report the margin rather than absolute overlap because
absolute overlap is not monotone in $k$ and can be dominated at intermediate $k$ by a
corpus-agnostic high-Fisher backbone. A positive margin is evidence that $\mathcal{C}_A$
shares more concept-conditioned high-Fisher support with $\mathcal{C}_F$ than generic text
does.

\begin{table}[h]
\centering
\small
\setlength{\tabcolsep}{8pt}
\caption{Reference-model Fisher top-$k$ overlap margins
$M_k = \mathrm{overlap}(\mathcal{C}_F,\mathcal{C}_A) -
\mathrm{overlap}(\mathcal{C}_F,\mathcal{C}_G)$. Positive values support model-side
adjacency.}
\label{tab:adjacency-fisher}
\begin{tabular}{l l rrr}
\toprule
\textbf{Model} & \textbf{Benchmark} & $M(10^3)$ & $M(10^4)$ & $M(10^5)$ \\
\midrule
Llama-3.1-8B & WMDP       & $+0.069$ & $+0.012$ & $+0.119$ \\
Llama-3.1-8B & TOFU       & $+0.113$ & $+0.026$ & $+0.154$ \\
Llama-3.1-8B & MUSE-Books & $+0.071$ & $-0.001$ & $+0.073$ \\
Llama-3.1-8B & MUSE-News  & $+0.273$ & $+0.039$ & $+0.159$ \\
\midrule
Zephyr-7B-$\beta$ & WMDP       & $+0.165$ & $+0.042$ & $+0.085$ \\
Zephyr-7B-$\beta$ & TOFU       & $+0.196$ & $+0.071$ & $+0.139$ \\
Zephyr-7B-$\beta$ & MUSE-Books & $-0.019$ & $-0.034$ & $-0.019$ \\
Zephyr-7B-$\beta$ & MUSE-News  & $+0.122$ & $+0.061$ & $+0.127$ \\
\bottomrule
\end{tabular}
\end{table}

The margin is positive in 20 of 24 cells, with a 21st at $-0.001$, and positive at all
three values of $k$ for 6 of 8 model--benchmark pairs. WMDP, TOFU and MUSE-News therefore
have consistent model-side adjacency evidence on both models, so AdjGap there is computed
against a corpus that is adjacent in the model's own parameter geometry and not only in
our description of it.

MUSE-Books is the informative exception, and the two diagnostics come apart there. It has
the highest semantic similarity of any benchmark ($0.354$), which is unsurprising when
both splits come from the same finetuning corpus, yet its Fisher margin is negative at
every $k$ on Zephyr-7B-$\beta$ and effectively zero at $k = 10^4$ on Llama. At that same
$k$, $\mathrm{overlap}(\mathcal{C}_A, \mathcal{C}_G)$ exceeds
$\mathrm{overlap}(\mathcal{C}_F, \mathcal{C}_A)$ on both models ($0.748$ against $0.535$
on Llama; $0.693$ against $0.529$ on Zephyr): the adjacent split shares more high-Fisher
support with generic English than with the forget split. Semantic relatedness does not
imply parameter-level coupling. We do not read this as invalidating the benchmark, but we
do qualify it: a clean AdjGap on MUSE-Books is weaker evidence of selective unlearning
than the same result on WMDP, where both forms of adjacency hold.

This is also consistent with one reading of our benchmark-level results, though it does
not establish it. WMDP's forget and adjacent corpora share extensive high-Fisher support,
so leaving $\mathcal{C}_A$ untouched may be structurally hard, which would fit how few
methods achieve a positive gap there. MUSE-Books shares substantially less, so an
apparently clean gap on it may partly reflect weaker model-side coupling rather than
localization alone. We offer this as a supported interpretation, not a demonstrated
mechanism.

\paragraph{Protocol for new benchmarks.}
To apply TRIAGE beyond WMDP, TOFU and MUSE we recommend reporting both
$\mathrm{sim}(\mathcal{C}_F, \mathcal{C}_A)$ against
$\mathrm{sim}(\mathcal{C}_A, \mathcal{C}_G)$ and $M_k$ alongside AdjGap. A failed semantic
test should trigger reconstruction of the corpus; a non-positive Fisher margin should
trigger caution in reading AdjGap as structural localization for that model. Both run on
the reference model alone and cost one Fisher pass per corpus.

Two limitations, stated plainly. Fisher overlap cannot establish deletion-scope
disjointness, which stays a semantic property of the deletion specification and is the one
criterion no diagnostic can automate. And with eight model--benchmark pairs we report
these margins descriptively; we fit no threshold beyond their sign, and we would not want
$M_k$ used as a pass/fail gate on this evidence base.

\section{Detailed Clustering Analysis and Localization Scatters}
\label{app:localization_clustering}

\begin{figure}[p]
\centering
\includegraphics[width=1.0\textwidth]{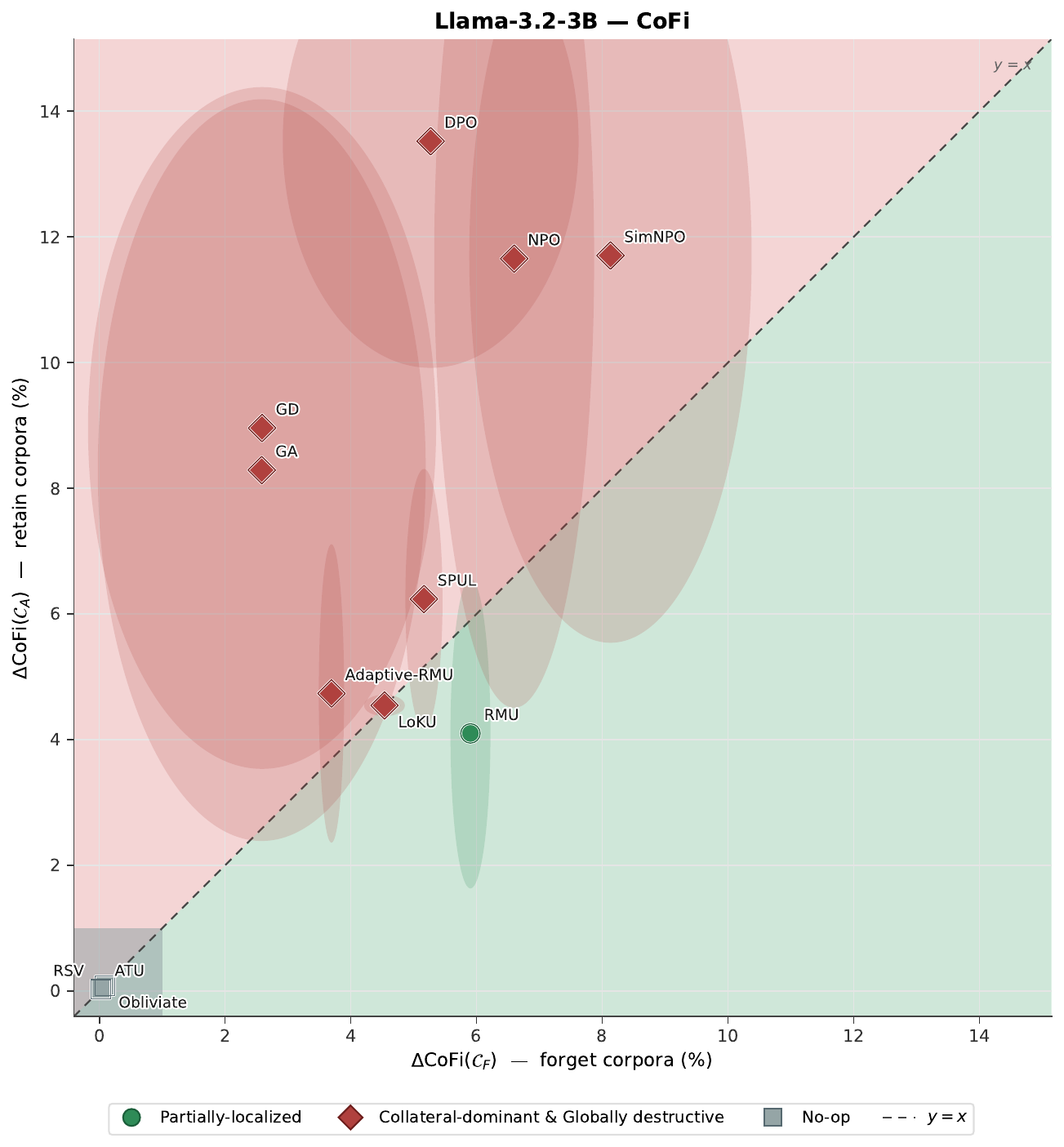}
\caption{Localization scatter on WMDP, Llama-3.2-3B. As on Llama-3.1-8B, RMU is the only
method below the diagonal. Ellipses are 95\% confidence intervals over three random
subsets of 200 samples.}
\label{fig:localization-wmdp-llama3b}
\end{figure}

\begin{figure}[p]
\centering
\includegraphics[width=1.0\textwidth]{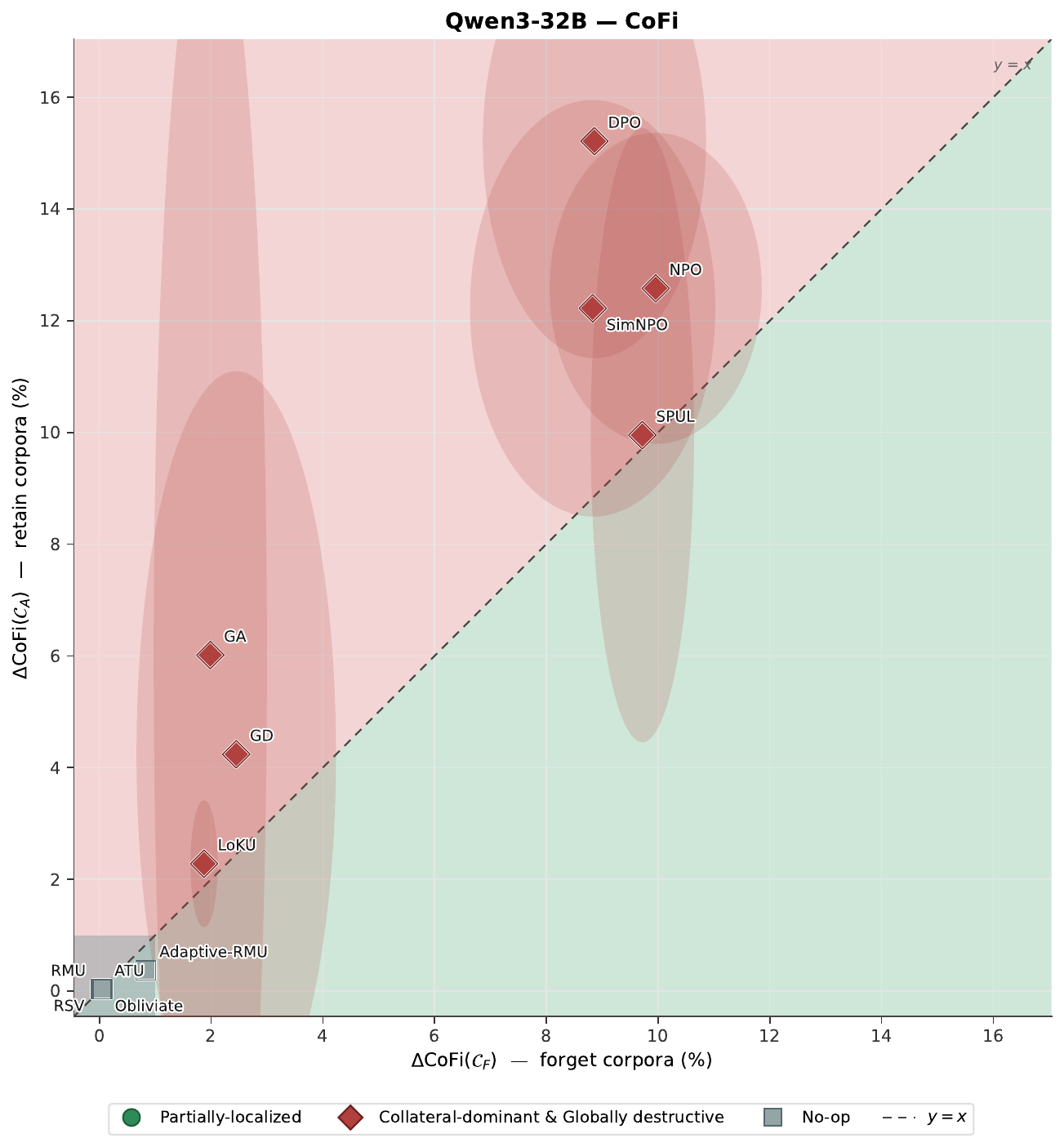}
\caption{Localization scatter on WMDP, Qwen3-32B. No method falls below the diagonal on
this model: every structurally active update moves the adjacent corpus at least as much as
the forget corpus.}
\label{fig:localization-wmdp-qwen}
\end{figure}

\begin{figure}[p]
\centering
\includegraphics[width=1.0\textwidth]{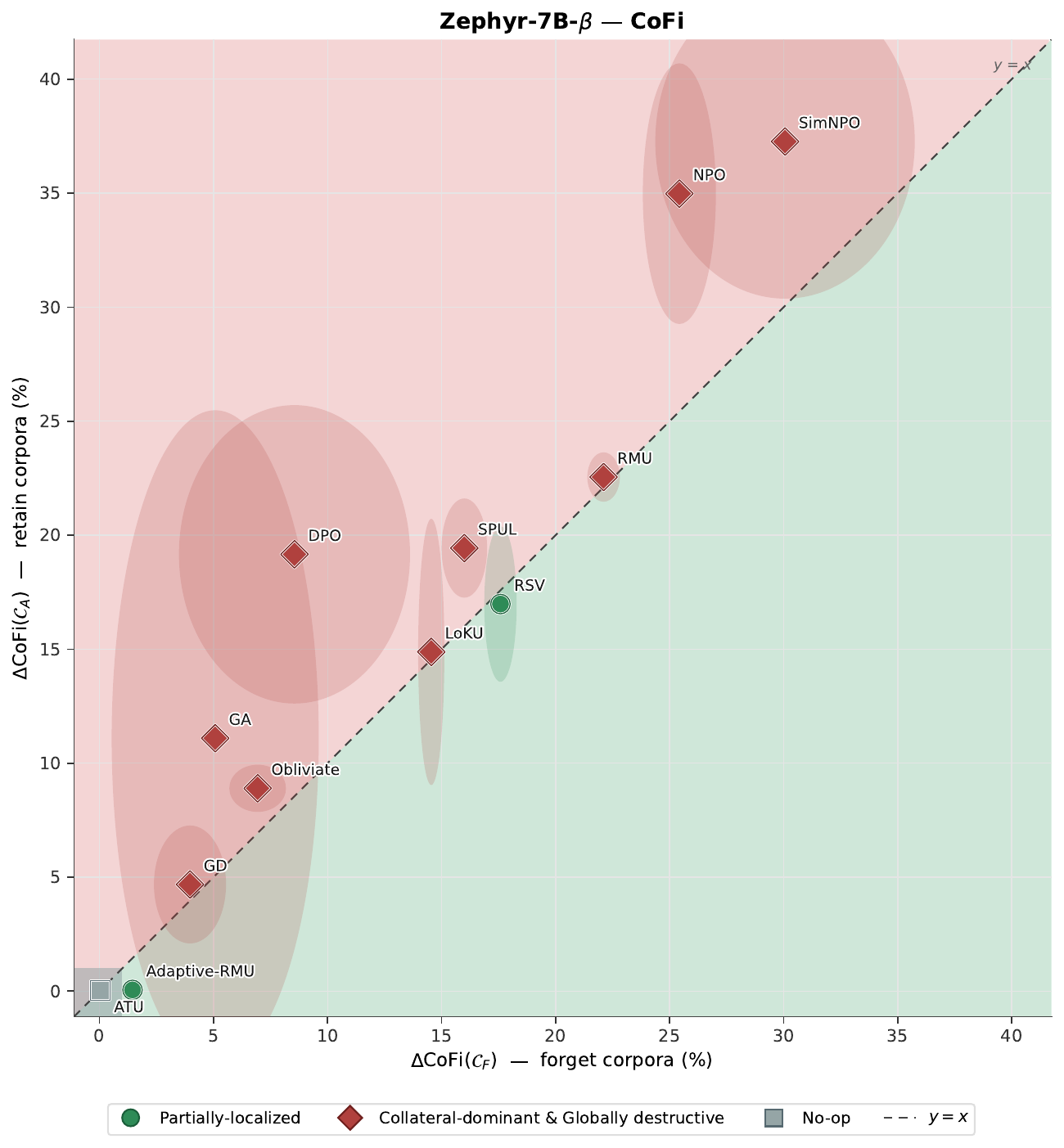}
\caption{Localization scatter on WMDP, Zephyr-7B-$\beta$. Note the axis range, roughly
twice that of the other models. Obliviate migrates from the no-op corner into the
collateral-dominant zone, and RSV moves out to a nominally forget-leading position whose
gap is inside its confidence interval.}
\label{fig:localization-wmdp-zephyr}
\end{figure}

\begin{figure}[p]
\centering
\includegraphics[width=1.0\textwidth]{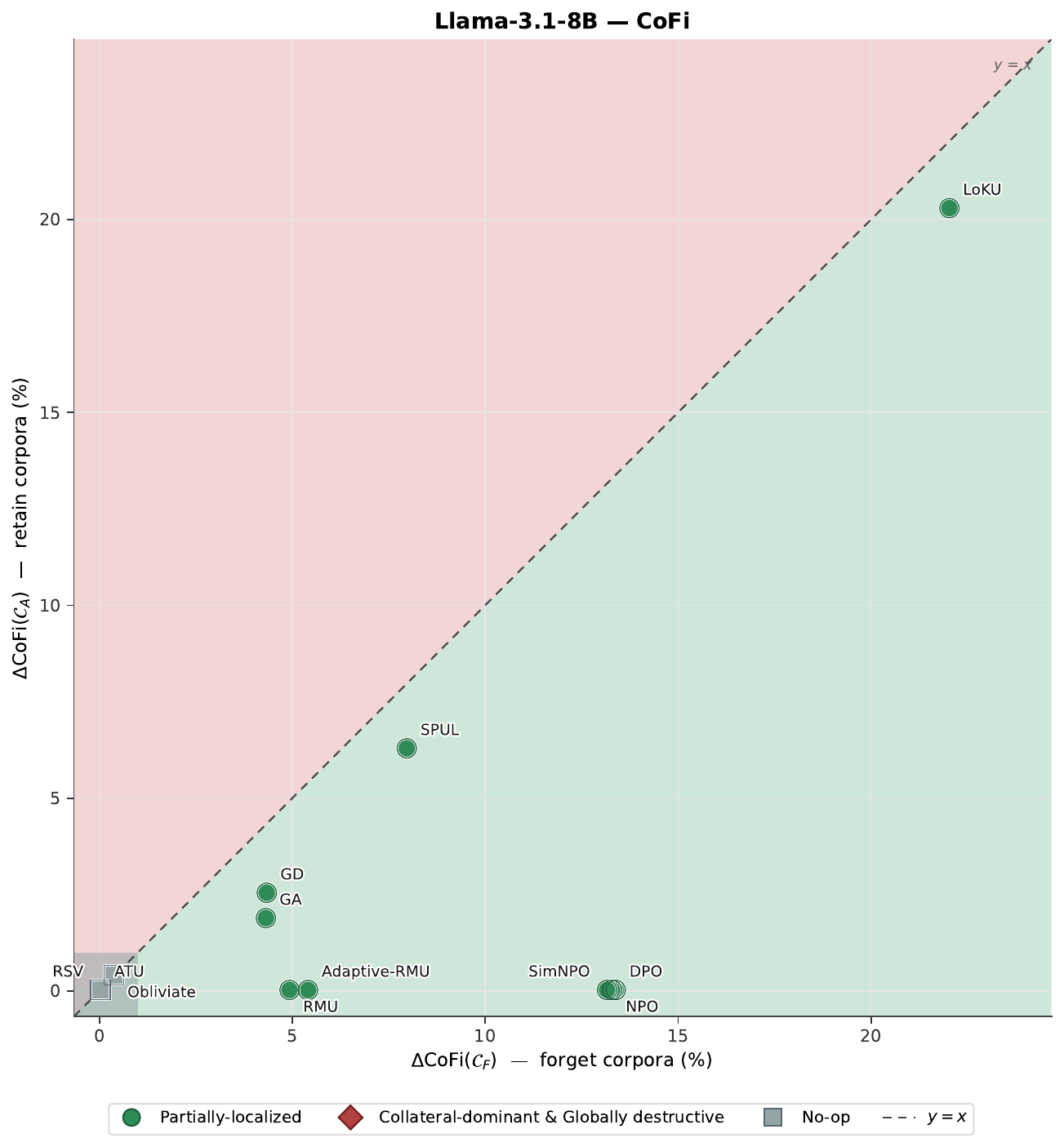}
\caption{Localization scatter on MUSE-Books, Llama-3.1-8B. NPO, SimNPO, DPO, RMU and
Adaptive-RMU rest on the $x$-axis with adjacent-retain at the measurement floor. Four of
the points below the diagonal --- GA, GD, SPUL and LoKU+FILA --- are globally destructive
on this benchmark, which this projection cannot show. No confidence intervals are shown
for MUSE-Books; see Appendix~\ref{app:muse-books-results}.}
\label{fig:localization-books-llama8b}
\end{figure}

\begin{figure}[p]
\centering
\includegraphics[width=1.0\textwidth]{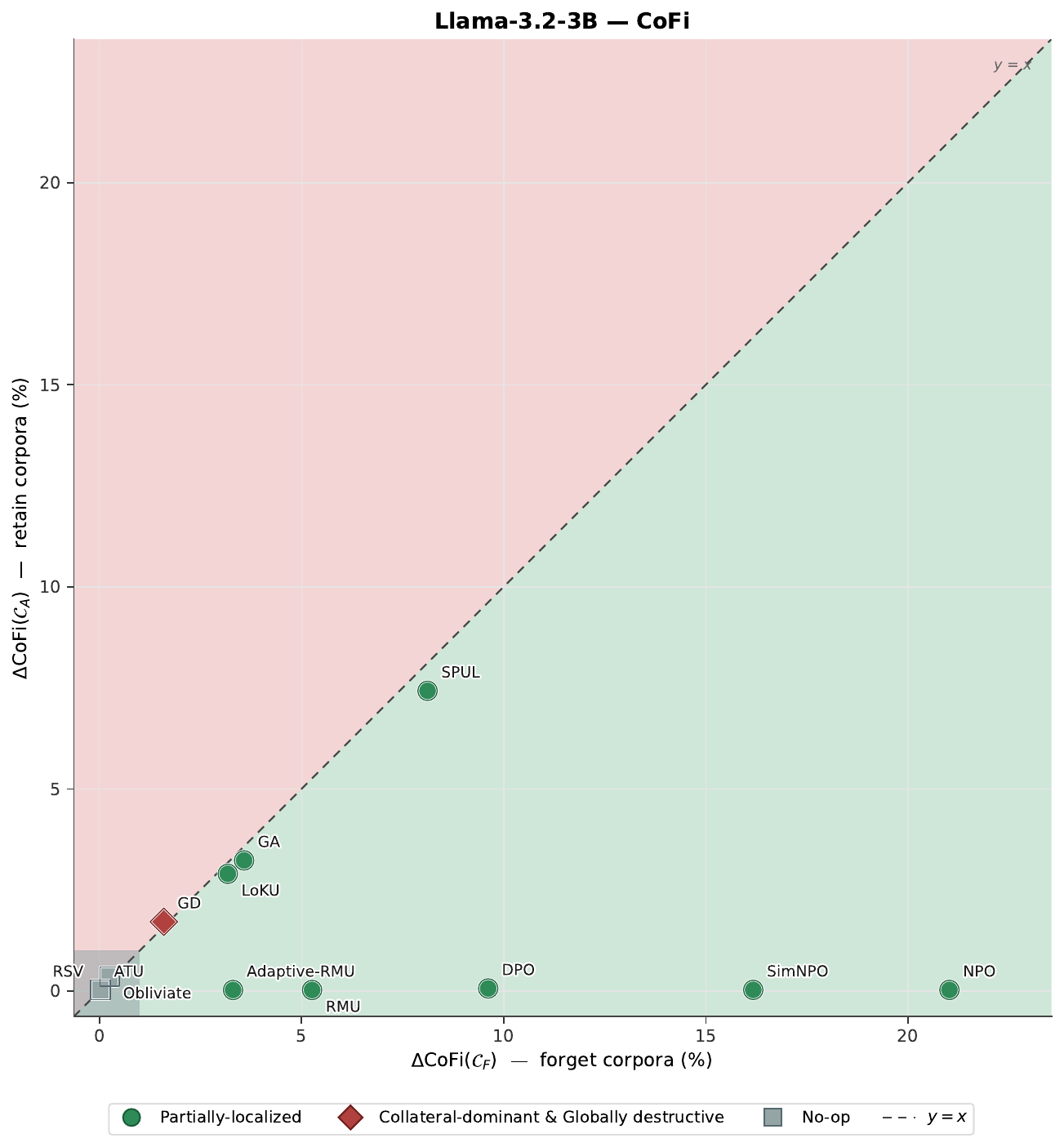}
\caption{Localization scatter on MUSE-Books, Llama-3.2-3B. NPO reaches $21.03\%$ on the
forget corpus against $0.03\%$ on adjacent-retain. GD is the only method above the
diagonal.}
\label{fig:localization-books-llama3b}
\end{figure}

\begin{figure}[p]
\centering
\includegraphics[width=1.0\textwidth]{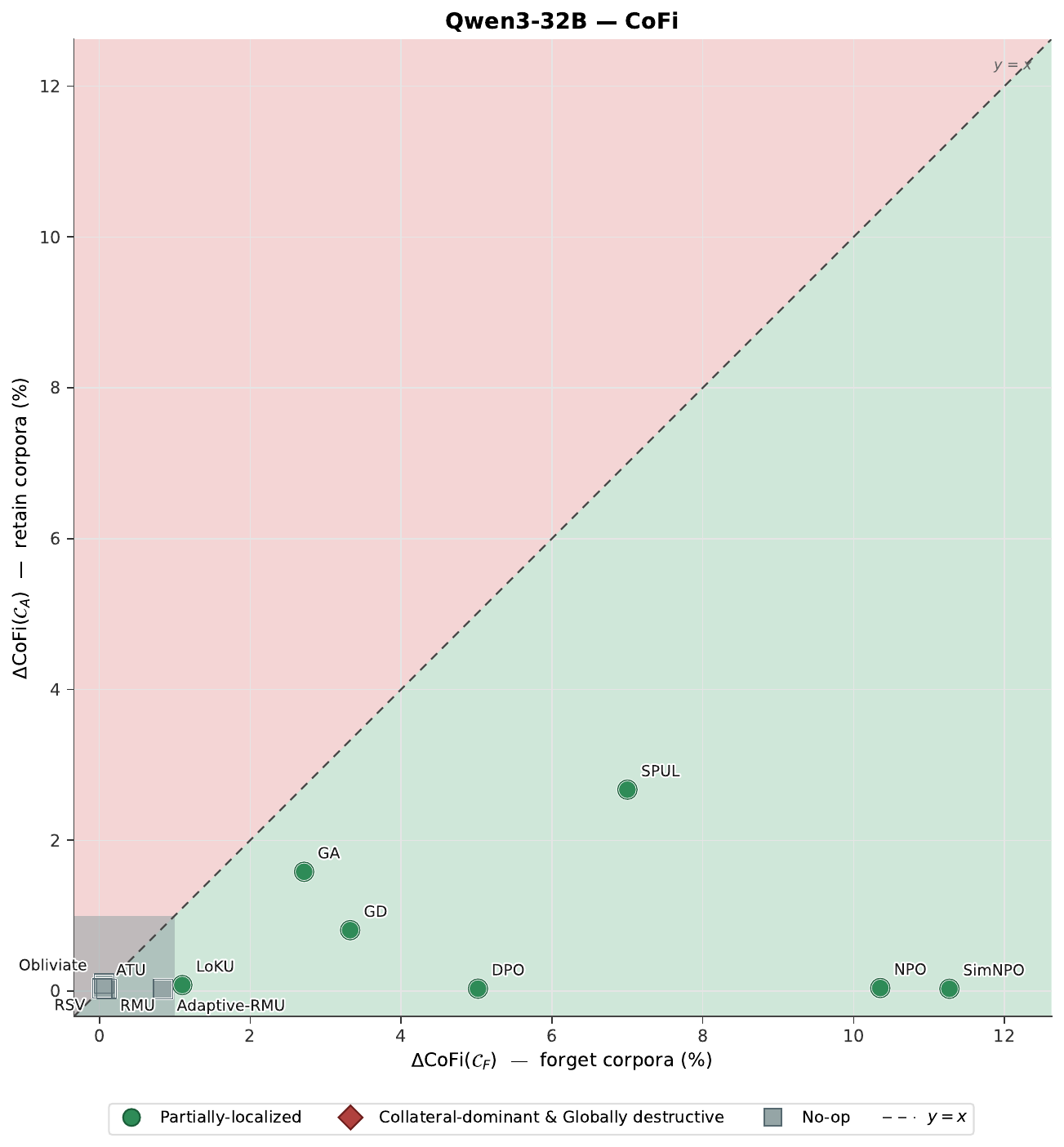}
\caption{Localization scatter on MUSE-Books, Qwen3-32B. Nine of the twenty configurations
are no-ops on this model, and no method lies above the diagonal.}
\label{fig:localization-books-qwen}
\end{figure}

\begin{figure}[p]
\centering
\includegraphics[width=1.0\textwidth]{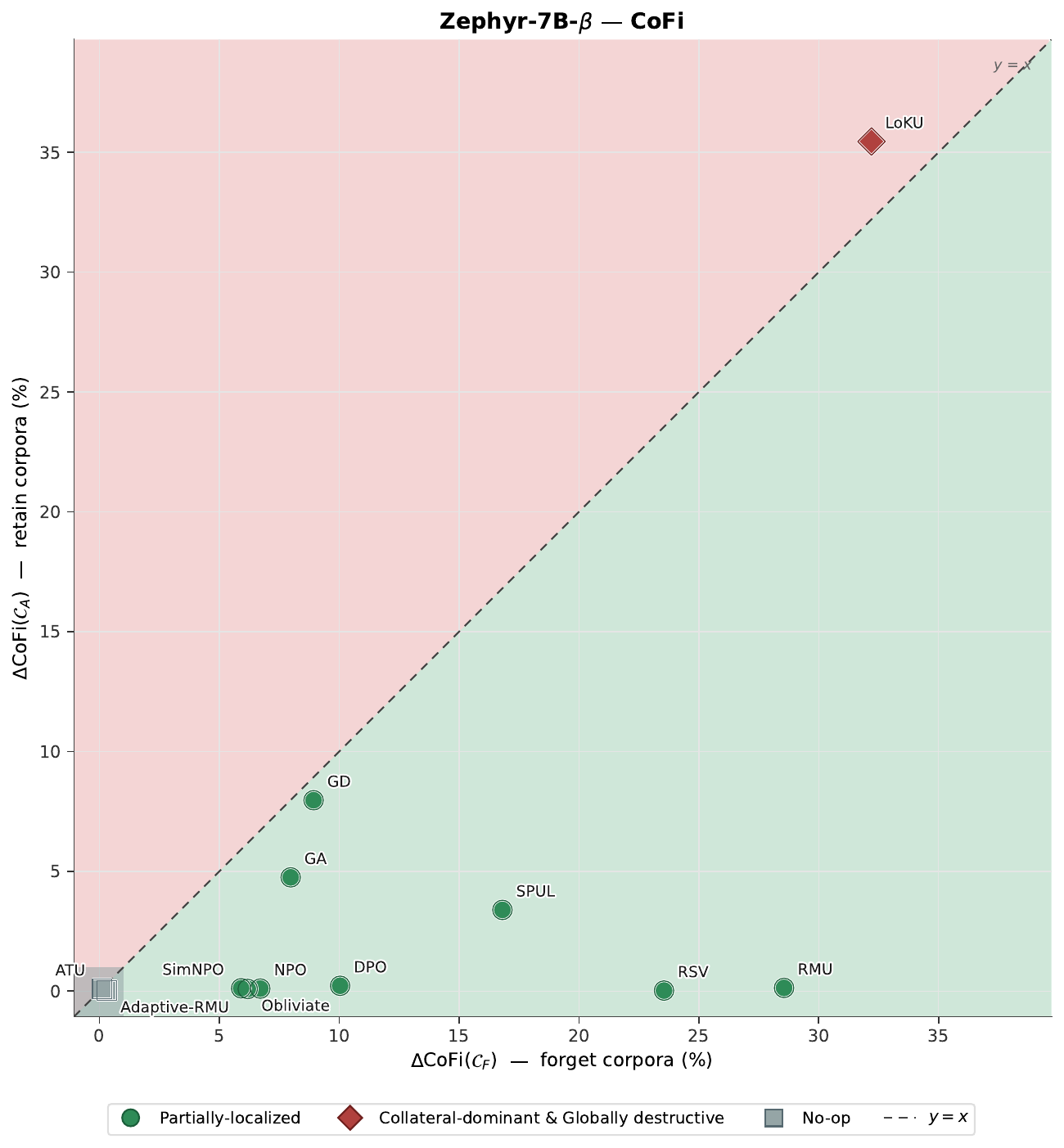}
\caption{Localization scatter on MUSE-Books, Zephyr-7B-$\beta$. LoKU+FILA is the only
method above the diagonal. RMU and RSV reach $28.56\%$ and $23.55\%$ on the forget corpus
while leaving adjacent-retain at the floor.}
\label{fig:localization-books-zephyr}
\end{figure}

Plotting $\Delta\mathrm{CoFi}(\mathcal{C}_F)$ against $\Delta\mathrm{CoFi}(\mathcal{C}_A)$
makes the adjacency gap visible at a glance: methods below the diagonal moved the forget
corpus more than its semantic neighbourhood, methods on or above it did not, and methods
in the shaded corner did not move anything. Figures~\ref{fig:localization-wmdp-llama3b}
to~\ref{fig:localization-wmdp-zephyr} give the WMDP scatters for the three models not
shown in Figure~\ref{fig:localization}, and
Figures~\ref{fig:localization-books-llama8b} to~\ref{fig:localization-books-zephyr} the
MUSE-Books scatters for all four.

\paragraph{What these plots cannot show.}
The scatter has two axes and the taxonomy has three quantities, so it separates the
partially localized class from the collateral-dominant one but cannot distinguish either
from the globally destructive class: a method that wrecks general text still has to be
plotted somewhere relative to the diagonal. This is why the legend groups
collateral-dominant and globally destructive under one marker, and it is not a cosmetic
point. On MUSE-Books with Llama-3.1-8B (Figure~\ref{fig:localization-books-llama8b}), GA,
GD, SPUL and LoKU+FILA all sit below the diagonal and would read as localized, while
their WikiText shifts run from $14.76\%$ to $27.00\%$ and place all four in the globally
destructive class. The scatter is a view of the adjacency gap, not of the classification;
the class assignments are in Table~\ref{tab:books-class-summary} and the corresponding
WMDP tables.

\paragraph{WMDP: the base model changes the picture.}
On Llama-3.1-8B (Figure~\ref{fig:localization}) only RMU sits below the diagonal, at
$7.05\%$ against $6.04\%$, and Llama-3.2-3B repeats the pattern with RMU at $5.90\%$
against $4.10\%$ (Figure~\ref{fig:localization-wmdp-llama3b}). Qwen3-32B is the extreme
case (Figure~\ref{fig:localization-wmdp-qwen}): no method falls below the diagonal at
all, and every method that does anything measurable moves its adjacent corpus at least as
much as its target. Zephyr-7B-$\beta$ (Figure~\ref{fig:localization-wmdp-zephyr}) is the
opposite extreme in magnitude --- the axes run to $40\%$ rather than $18\%$ --- and the
membership changes with it. Obliviate moves from the no-op corner on the other three
models into the collateral-dominant zone here ($6.93\%$ against $8.90\%$), and RSV, also a
no-op elsewhere, moves out to $17.58\%$ against $16.98\%$. The Zephyr panel is showing that the same twelve updates produce footprints several times larger on this model than on any other.

\paragraph{MUSE-Books: the adjacency gap opens up.}
The MUSE-Books scatters look qualitatively different. On Llama-3.1-8B
(Figure~\ref{fig:localization-books-llama8b}) the preference-based methods sit on the
$x$-axis: NPO reaches $13.29\%$ on the forget corpus against $0.03\%$ on adjacent-retain,
with SimNPO at $13.16\%$ and DPO at $13.38\%$ against the same floor. RMU and Adaptive-RMU
do the same at smaller magnitudes. The pattern repeats on Llama-3.2-3B
(Figure~\ref{fig:localization-books-llama3b}, NPO at $21.03\%$ against $0.03\%$),
Qwen3-32B (Figure~\ref{fig:localization-books-qwen}, SimNPO at $11.27\%$ against
$0.03\%$), and Zephyr-7B-$\beta$ (Figure~\ref{fig:localization-books-zephyr}, RMU at
$28.56\%$ against $0.14\%$). On Zephyr, LoKU+FILA is the only method above the diagonal,
at $32.20\%$ against $35.46\%$.

\paragraph{Why the two benchmarks differ.}
The contrast comes from how $\mathcal{C}_A$ is constructed, not from a change in how the
methods behave. WMDP targets pretraining-distributed knowledge: $\mathcal{C}_F$ is
biosecurity or cybersecurity text and $\mathcal{C}_A$ is general bioscience or programming
content, and the two share parameter pathways, so movement on one propagates to the other.
MUSE uses a finetune-then-unlearn protocol in which $\mathcal{C}_A$ comes from the same
finetuning corpus as $\mathcal{C}_F$ --- a different portion of the same fictional
universe --- and the update concentrates on the comparatively narrow region that
finetuning modified. A narrowly defined adjacent set drawn from that same region has
little to lose. The reading we take from this is not that NPO and SimNPO are well
localized in general, but that they are well localized \emph{when both corpora occupy the
same finetuning-targeted region}, and that the much larger adjacent movement on WMDP shows
they do not localize when the target is distributed through pretraining. This is also the
benchmark where our Fisher-overlap diagnostic is least supportive
(Appendix~\ref{app:adjacency}): on MUSE-Books the adjacent split shares more high-Fisher
support with generic English than with the forget split, so a clean gap there is weaker
evidence of selective unlearning than the same gap on WMDP. Localization is not an
intrinsic property of an algorithm; it emerges from the interaction between the method,
the model, and how the targeted knowledge is distributed.

\section{Full TRIAGE Results}
\label{app:full-results}

This section contains complete tabular TRIAGE results for every (method, model,
benchmark) combination, including the noise-augmented ($\nu$) variants. Each benchmark
is given its own subsection; within a subsection, results are reported per model and
per corpus, with CoFi and CHess as means $\pm$ 95\% confidence intervals over three
random subsets of 200 samples. Section~\ref{app:heatmaps} collects the per-model
heatmaps for all benchmarks in one place. 

Qwen3-32B is evaluated only on WMDP and MUSE-Books, as its scale makes a full evaluation on TOFU and MUSE-News prohibitively expensive. The remaining three models are evaluated on all three benchmarks.

\paragraph{Reading the tables.}
CoFi and CHess are relative shifts, so an increase on a forget corpus is the intended
effect of unlearning and any increase on a retain corpus or on the general-text corpus
is collateral. Class membership is determined by the CoFi aggregates alone: CHess and
perplexity are reported alongside to give deeper insight on how the model's parameter space has altered and as a sanity check to see how fluent the model is, respectively. Within each per-model table, methods are
grouped by the class they occupy \emph{on that model and that benchmark}, so the
grouping is not identical from one table to the next. For cross-model comparison at a
glance, each subsection also provides a grid of class assignments in which the row
order is held fixed.

\paragraph{Perplexity is a sanity check, reported on a log scale.}
The unlearned/original perplexity ratio spans many orders of magnitude across the
method set, from $1.00$ for methods that leave the model untouched to values far beyond
any scale on which a model still produces text. We therefore report $\log_{10}$ of the
ratio, where $0.00$ denotes unchanged fluency. We omit subset confidence intervals on
this metric because at these magnitudes the linear-scale interval is of the same order
as the mean itself. The ratio should be read as an order-of-magnitude indicator of
whether a corpus is still modelled at all, and it serves here to confirm that the
structural picture CoFi reports is not an artefact of the probe: a method that CoFi
places in the no-op class should leave perplexity at $1.00$, and one CoFi places in the
globally destructive class should show general-text degradation. It is not a
classification input.

\paragraph{Reading the heatmaps.}
The heatmaps in Section~\ref{app:heatmaps} show the same numbers as the tables, with
one row per method, one column per corpus, and the three metrics side by side. Colour
scales are set per metric panel, so brightness is comparable down a panel but not
across panels. Each class has a visual signature in the CoFi panel, which is the panel
the classification is made from. No-op methods are uniformly dark: every corpus sits at
the measurement floor. Partially localized methods brighten on the two forget columns
and fade across the two retain columns, giving a left-to-right decay. Collateral
dominant methods invert that gradient, with the retain columns at least as bright as
the forget columns, but the WikiText column stays dark --- the damage spreads to
neighbours of the forget set and stops there. Globally destructive methods are the only
ones whose WikiText column lights up, and the fifth column is therefore the single
fastest way to read the taxonomy off the figure. The CHess panel is not a repeat of the CoFi panel at a different contrast: it registers
distributional reshaping that leaves the ranking intact, so a row that is
dark in CoFi and bright in CHess is telling us something real about the depth of the
intervention. The PPL panel has the narrower job of confirming that a bright CoFi row
corresponds to a model whose fluency has genuinely degraded.

\paragraph{Class membership is a property of the (method, model, benchmark) triple.}
The TRIAGE classes describe a structural footprint, and that footprint is a joint
property of the algorithm, the base model and the evaluation corpora rather than an
intrinsic label attached to the algorithm. A method is expected to migrate between
classes as any of the three changes, and the tables below show that it does: the same
optimizer that leaves a large model essentially untouched can destroy a smaller one,
and a method whose damage is confined to the retain corpora on one model can reach
general text on another. Very little is stable. Only the two extremes of our method set
hold their class across every model we evaluate, and everything between them moves at
least once. This is the intended behaviour of the taxonomy rather than an instability
in it. The diagnostic answers ``what did this run do to this model'', so a practitioner
must re-run it whenever the model, the forget set or the evaluation suite changes, and
the per-benchmark class assignments reported here should not be read as method-level
verdicts.

\paragraph{Noise augmentation ($\nu$).}
Every method that supports the RNA mechanism is reported twice, as a base run and as a
noise-augmented run marked ($\nu$). Because the effect of noise augmentation depends on
the stability of the underlying model and on the forget objective, we analyse the two
variants separately within each benchmark subsection rather than drawing a single
conclusion here.

\subsection{WMDP}
\label{app:wmdp-results}

\begin{figure}[p]
\centering
\includegraphics[width=\textwidth]{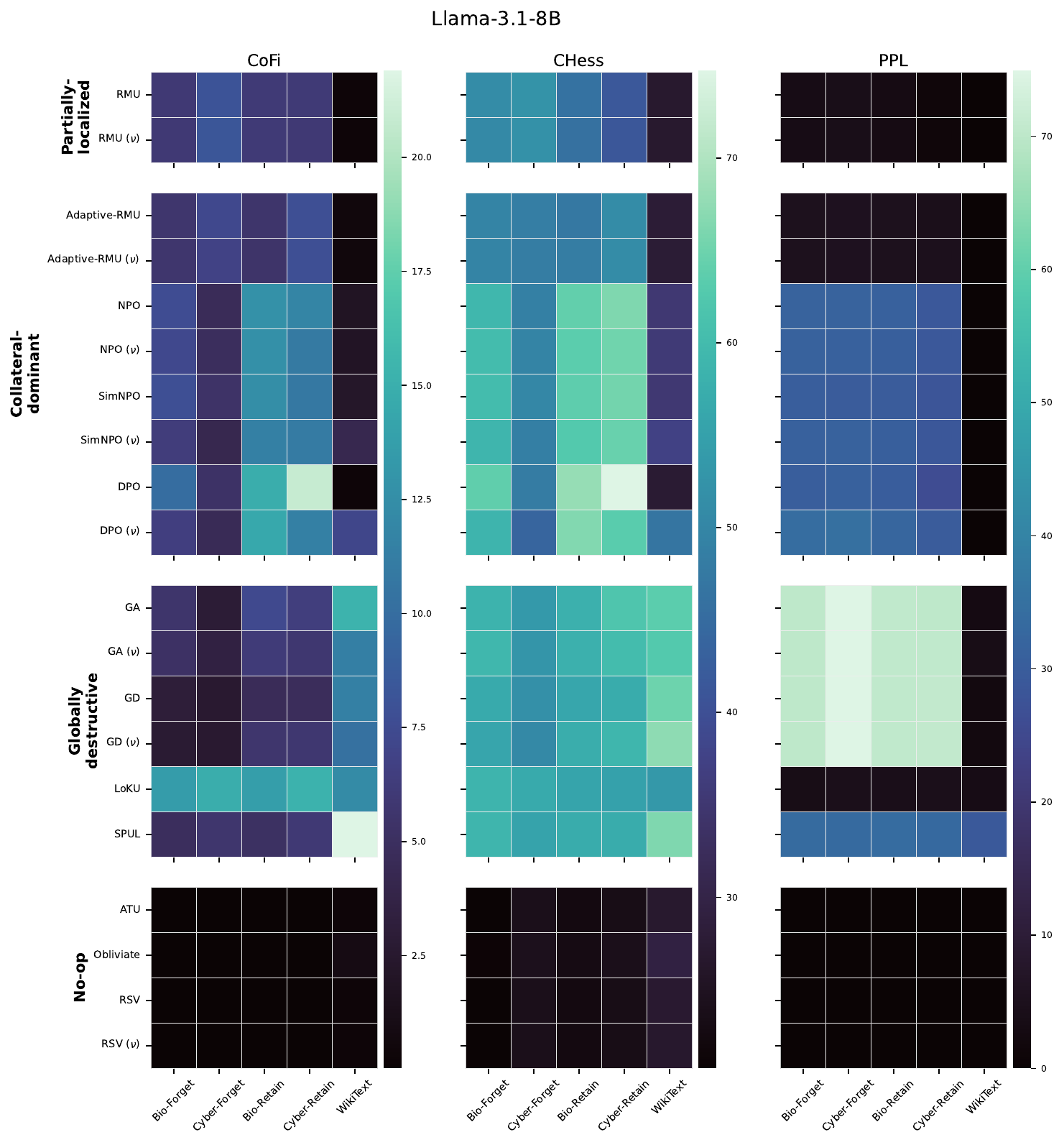}
\caption{WMDP TRIAGE heatmap for Llama-3.1-8B. Rows are grouped by TRIAGE class;
columns are the five evaluation corpora. Colour scales are independent per metric
panel. PPL is shown as $\log_{10}$ of the unlearned/original ratio.}
\label{fig:heatmap-wmdp-llama8b}
\end{figure}

\begin{figure}[p]
\centering
\includegraphics[width=\textwidth]{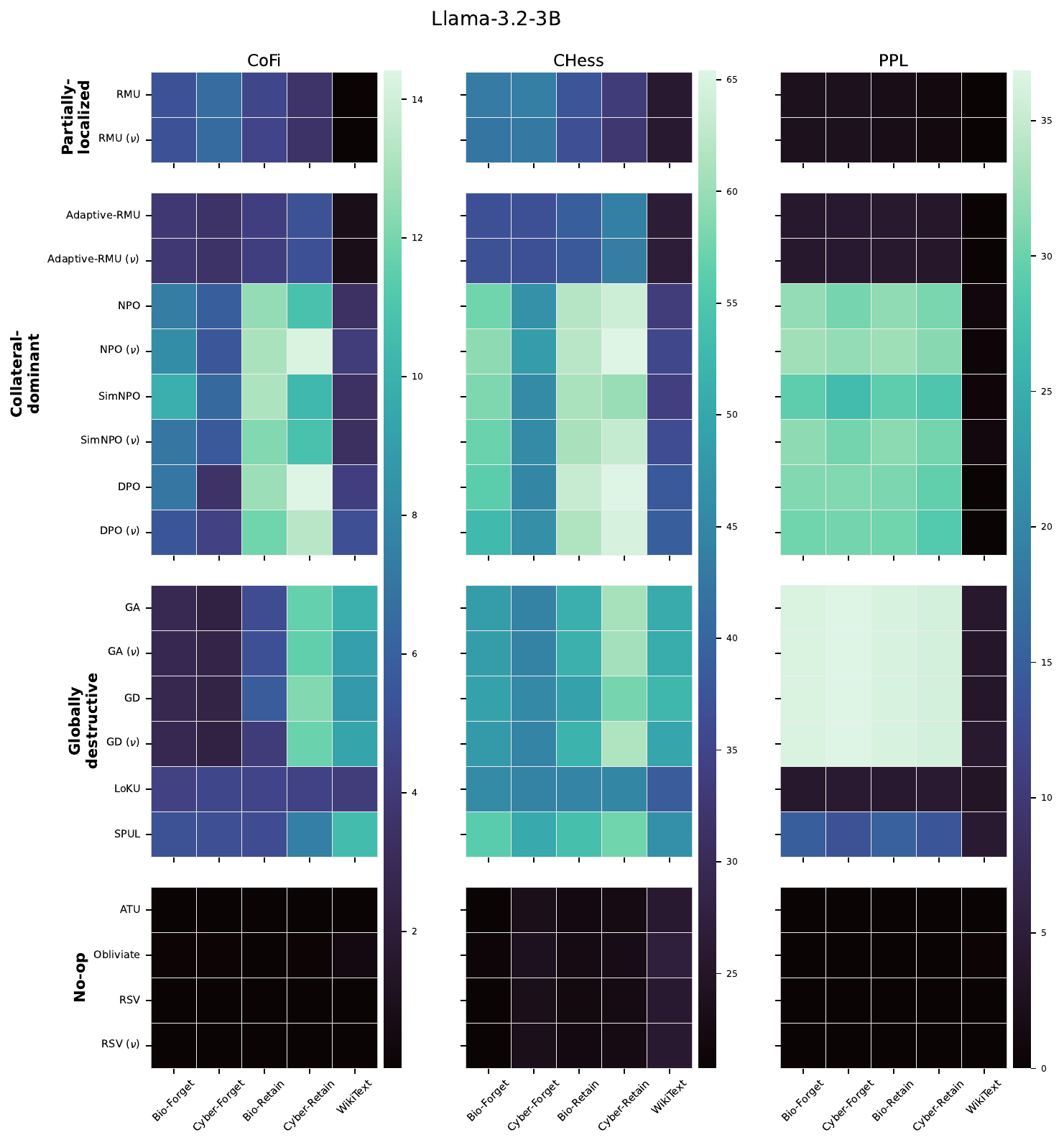}
\caption{WMDP TRIAGE heatmap for Llama-3.2-3B. Rows are grouped by TRIAGE class;
columns are the five evaluation corpora. Colour scales are independent per metric
panel. PPL is shown as $\log_{10}$ of the unlearned/original ratio.}
\label{fig:heatmap-wmdp-llama3b}
\end{figure}

\begin{figure}[p]
\centering
\includegraphics[width=\textwidth]{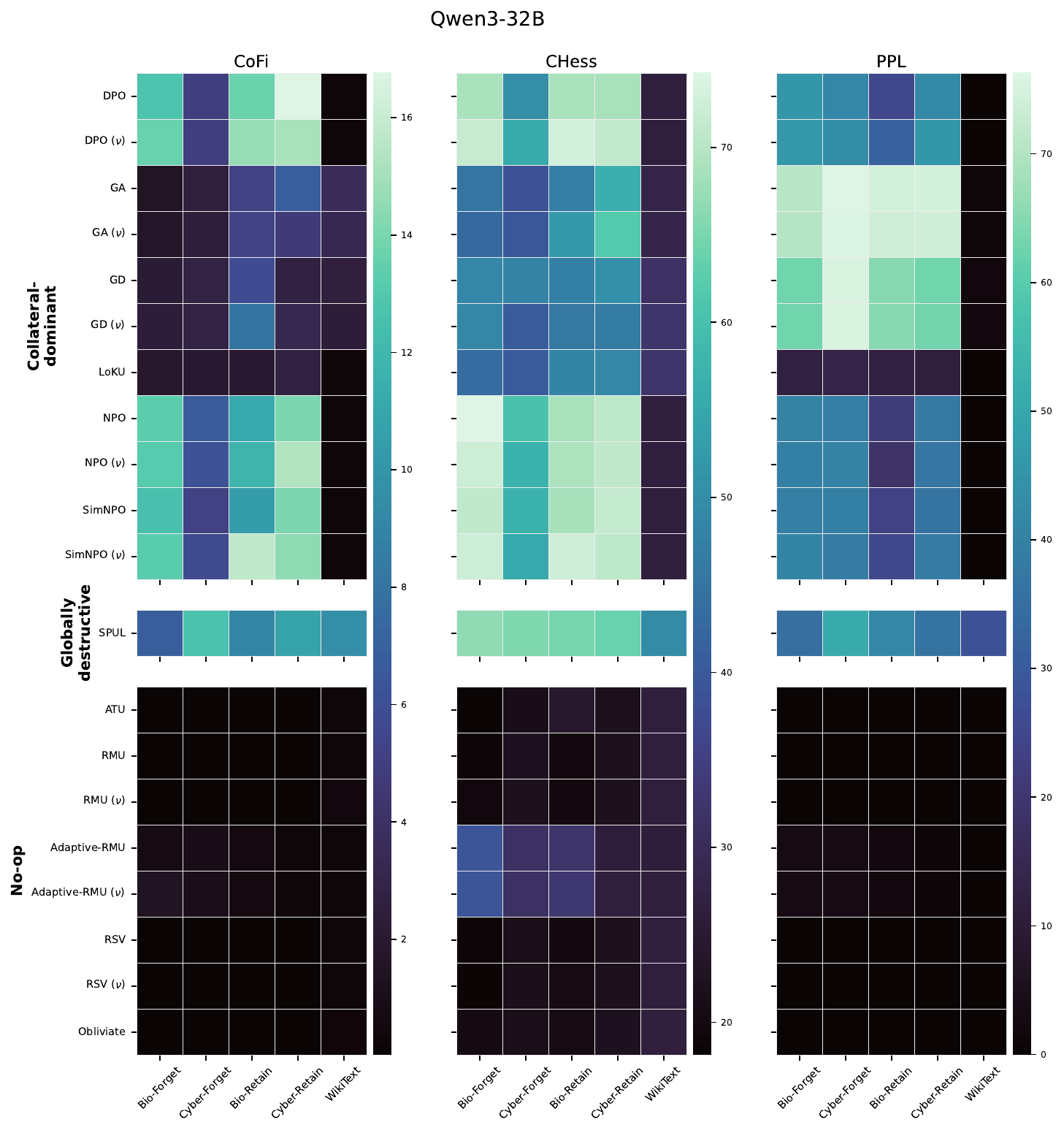}
\caption{WMDP TRIAGE heatmap for Qwen3-32b. Rows are grouped by TRIAGE class;
columns are the five evaluation corpora. Colour scales are independent per metric
panel. PPL is shown as $\log_{10}$ of the unlearned/original ratio.}
\label{fig:heatmap-wmdp-qwen}
\end{figure}

\begin{figure}[p]
\centering
\includegraphics[width=\textwidth]{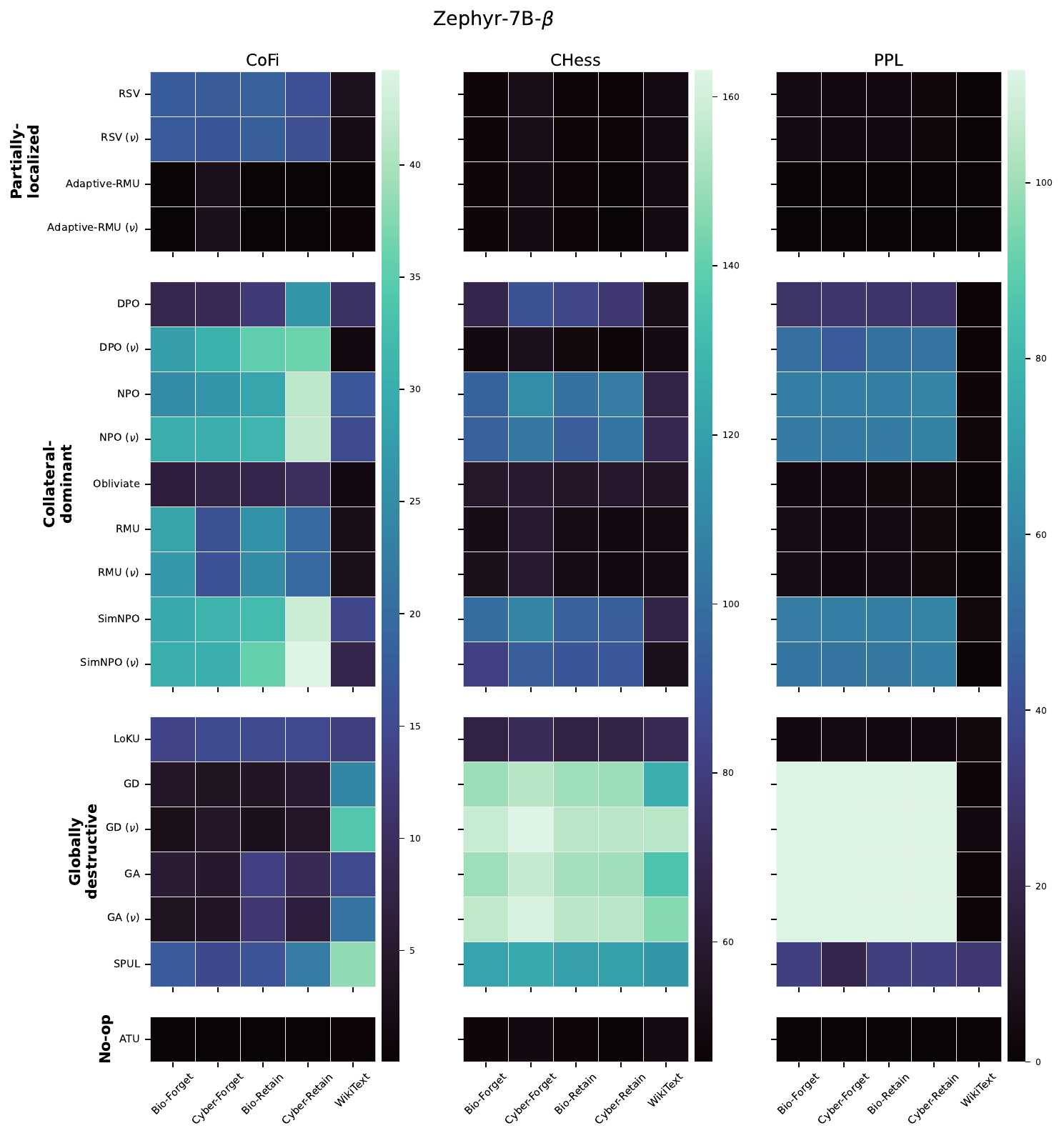}
\caption{WMDP TRIAGE heatmap for Zephyr-7B-$\beta$. Rows are grouped by TRIAGE class;
columns are the five evaluation corpora. Colour scales are independent per metric
panel. PPL is shown as $\log_{10}$ of the unlearned/original ratio.}
\label{fig:heatmap-wmdp-zephyr}
\end{figure}

Tables~\ref{tab:wmdp-cofi-chess-llama8b}--\ref{tab:wmdp-cofi-chess-zephyr} give the
per-topic CoFi and CHess relative shifts for Llama-3.1-8B, Llama-3.2-3B, Qwen3-32B and
Zephyr-7B-$\beta$ respectively. Tables~\ref{tab:wmdp-ppl-1} and~\ref{tab:wmdp-ppl-2}
give the corresponding perplexity ratios,
Table~\ref{tab:wmdp-aggregate-other} repeats the aggregated
$\mathcal{C}_F/\mathcal{C}_A/\mathcal{C}_G$ view of Table~\ref{tab:headline} for the
three models not shown in the main text, and Table~\ref{tab:triage-class-summary}
collects the class assignments in a single grid. The corresponding heatmaps are
Figures~\ref{fig:heatmap-wmdp-llama8b}--\ref{fig:heatmap-wmdp-zephyr}. The largest
perplexity ratios we observe on this benchmark, around $10^{112}$ for gradient ascent
and gradient difference on Zephyr-7B-$\beta$, are the reason for the logarithmic
reporting described above.

\paragraph{Class migration across models.}
WMDP illustrates the model dependence of the taxonomy directly, and the migrations run
in both directions. RMU registers as a no-op on Qwen3-32B, is partially localized on
both Llama models, and becomes collateral dominant on Zephyr-7B-$\beta$, where it
shifts the forget and adjacent retain corpuses by more than twenty percent. Obliviate is a no-op on the three models but collateral dominant on Zephyr. Adaptive-RMU spans three classes on
its own: collateral dominant on the two Llama models, partially localized on
Zephyr-7B-$\beta$, and a no-op on Qwen3-32B, where it moves no corpus by more than
$0.93\%$. Gradient ascent and gradient difference are globally destructive on three of
the four models but fall just short on Qwen3-32B, with globality ratios of $\rho = 0.59$
and $0.60$ against a threshold of $0.75$; these are the two largest $\rho$ values among
all methods we do not place in the globally destructive class, and their WikiText
perplexity ratios there (roughly $10$ and $29$) confirm that the general-text damage is
real, but small. Qwen3-32B is also the only model on which no method is
partially localized: every method that does anything at all to it spreads that effect
to the retain corpora. Only ATU, which never leaves the no-op class, and SPUL, which is
globally destructive on all four models, hold a single class across the grid. The
preference-optimization family (NPO, SimNPO, DPO) is collateral dominant on every
model, though the magnitude of that collateral damage varies by more than a factor of
three between Llama-3.1-8B and Zephyr-7B-$\beta$.

\paragraph{Effect of noise augmentation.}
Noise augmentation leaves the structural reading of WMDP essentially untouched: base
and $\nu$ variants receive the same TRIAGE class in all 32 method--model pairs for
which both were run. The agreement in magnitude is close but not uniform. The median
across pairs of the largest per-corpus CoFi discrepancy is $1.53$ percentage points,
comfortably inside the subset confidence intervals, and the two Llama models account
for most of the tightest pairs. The exceptions are concentrated on the more
fragile Zephyr-7B-$\beta$ and on DPO in particular: DPO($\nu$) differs from DPO by
$23.1$ percentage points on the Zephyr bio-retain corpus ($12.22\%$ against $35.30\%$)
and by comparable margins on the two forget corpora, while its WikiText shift falls
from $10.62\%$ to $1.52\%$. On Llama-3.1-8B the same pair diverges mainly on general
text, where CoFi rises from $0.22\%$ to $7.06\%$. On WMDP, then, RNA can change how
hard a method hits a given model, sometimes substantially, without changing the shape
of the footprint it leaves.

\paragraph{Configuration notes.}
Two entries require comment. LoKU is run with its FILA component on Llama-3.1-8B,
Llama-3.2-3B and Zephyr-7B-$\beta$, but in the FILA-free configuration on Qwen3-32B due to 4-bit quantization of this model.
The two are listed as separate rows in Table~\ref{tab:triage-class-summary} and are not
directly comparable: LoKU+FILA is globally destructive on all three models where it is
run, whereas the FILA-free configuration is collateral dominant. Second, both RMU and
Adaptive-RMU land in the no-op class on Qwen3-32B. RMU leaves CoFi at the measurement
floor ($\leq 0.09\%$) on every corpus, with perplexity ratios between $1.00$ and
$1.35$; Adaptive-RMU moves CoFi slightly further ($\mathcal{C}_F = 0.83\%$) but not
enough to separate it from the untouched baselines, even though its forget-corpus
perplexity rises by three orders of magnitude. That disagreement between the two probes
is exactly the kind of discrepancy the heatmaps make visible, and we return to it in
Section~\ref{app:heatmaps}.

\begin{table}[t]
\centering
\footnotesize
\setlength{\tabcolsep}{1.5pt}
\caption{WMDP per-topic CoFi and CHess relative shifts (\%) on \textbf{Llama-3.1-8B}, mean $\pm$ 95\% CI over three random subsets of 200 samples. $\uparrow$ on the forget splits = stronger unlearning; $\downarrow$ on the retain splits and on \textit{wiki} = less collateral damage. Methods are grouped by the TRIAGE class they fall into \emph{for this model}; the assignment uses the CoFi aggregates alone. ($\nu$) marks the randomly-perturbed variant.}
\label{tab:wmdp-cofi-chess-llama8b}
\resizebox{\textwidth}{!}{%
\begin{tabular}{l rrrrr @{\hspace{4pt}} rrrrr}
\toprule
& \multicolumn{5}{c}{\textbf{CoFi (\%)}} & \multicolumn{5}{c}{\textbf{CHess (\%)}} \\
\cmidrule(lr){2-6}\cmidrule(lr){7-11}
\textbf{Method} & \textit{bio-F} & \textit{cyber-F} & \textit{bio-R} & \textit{cyber-R} & \textit{wiki} & \textit{bio-F} & \textit{cyber-F} & \textit{bio-R} & \textit{cyber-R} & \textit{wiki} \\
\midrule
\multicolumn{11}{l}{\textit{No-op class}} \\
ATU                    & 0.03\ci{0.00} & 0.03\ci{0.00} & 0.03\ci{0.00} & 0.03\ci{0.00} & 0.21\ci{0.17} & 20.7\ci{0.4} & 23.9\ci{0.6} & 22.6\ci{1.0} & 23.4\ci{2.2} & 26.7\ci{0.5} \\
Obliviate              & 0.07\ci{0.00} & 0.08\ci{0.01} & 0.07\ci{0.01} & 0.09\ci{0.02} & 0.93\ci{0.41} & 21.1\ci{0.5} & 24.2\ci{0.5} & 22.9\ci{1.2} & 24.0\ci{2.3} & 29.2\ci{0.6} \\
RSV                    & 0.03\ci{0.00} & 0.03\ci{0.00} & 0.03\ci{0.00} & 0.03\ci{0.00} & 0.23\ci{0.18} & 20.8\ci{0.4} & 23.9\ci{0.6} & 22.6\ci{1.0} & 23.4\ci{2.2} & 26.9\ci{0.8} \\
RSV ($\nu$)            & 0.03\ci{0.01} & 0.03\ci{0.00} & 0.03\ci{0.00} & 0.03\ci{0.00} & 0.22\ci{0.15} & 20.7\ci{0.4} & 23.9\ci{0.6} & 22.6\ci{1.0} & 23.4\ci{2.2} & 26.6\ci{0.6} \\
\midrule
\multicolumn{11}{l}{\textit{Partially-localized class}} \\
RMU                    & 5.98\ci{0.11} & 8.12\ci{0.63} & 6.03\ci{0.19} & 6.06\ci{2.38} & 0.24\ci{0.17} & 50.9\ci{1.2} & 52.4\ci{3.0} & 46.1\ci{2.5} & 41.4\ci{14.7} & 26.7\ci{0.5} \\
RMU ($\nu$)            & 5.99\ci{0.11} & 8.22\ci{0.59} & 6.01\ci{0.20} & 5.96\ci{2.66} & 0.21\ci{0.14} & 50.6\ci{0.9} & 52.1\ci{1.2} & 45.8\ci{1.9} & 41.2\ci{12.2} & 26.8\ci{0.5} \\
\midrule
\multicolumn{11}{l}{\textit{Collateral-dominant class}} \\
Adaptive-RMU           & 5.65\ci{0.11} & 7.20\ci{0.34} & 5.50\ci{0.20} & 7.75\ci{2.22} & 0.50\ci{1.06} & 49.5\ci{0.7} & 48.3\ci{0.9} & 47.2\ci{2.1} & 50.9\ci{1.2} & 27.7\ci{1.8} \\
Adaptive-RMU ($\nu$)   & 5.61\ci{0.12} & 6.82\ci{0.37} & 5.48\ci{0.21} & 7.73\ci{0.80} & 0.51\ci{1.00} & 49.5\ci{0.8} & 48.2\ci{1.9} & 48.2\ci{3.8} & 50.9\ci{1.7} & 27.5\ci{1.6} \\
NPO                    & 7.54\ci{2.74} & 4.59\ci{1.17} & 12.71\ci{9.04} & 11.77\ci{4.78} & 1.85\ci{4.57} & 59.1\ci{5.5} & 48.9\ci{3.4} & 63.9\ci{21.2} & 66.0\ci{2.9} & 35.2\ci{9.0} \\
NPO ($\nu$)            & 7.22\ci{2.24} & 4.82\ci{0.67} & 12.59\ci{9.60} & 10.88\ci{6.66} & 1.96\ci{3.93} & 59.9\ci{5.5} & 49.5\ci{2.8} & 63.5\ci{21.8} & 65.0\ci{4.8} & 35.6\ci{11.4} \\
SimNPO                 & 7.78\ci{2.89} & 5.34\ci{0.74} & 12.47\ci{8.65} & 10.77\ci{5.66} & 2.20\ci{4.01} & 60.1\ci{7.2} & 50.2\ci{2.7} & 63.6\ci{18.8} & 65.2\ci{4.2} & 35.1\ci{15.6} \\
SimNPO ($\nu$)         & 6.30\ci{1.21} & 4.16\ci{0.67} & 11.42\ci{9.34} & 10.96\ci{5.62} & 4.13\ci{4.74} & 58.7\ci{5.2} & 48.7\ci{2.2} & 62.6\ci{21.3} & 64.5\ci{4.0} & 37.3\ci{11.0} \\
DPO                    & 9.95\ci{3.73} & 5.25\ci{0.41} & 14.84\ci{2.37} & 20.80\ci{3.44} & 0.22\ci{0.20} & 63.7\ci{5.5} & 48.0\ci{2.7} & 67.7\ci{1.7} & 74.6\ci{2.8} & 27.3\ci{0.2} \\
DPO ($\nu$)            & 6.44\ci{3.31} & 4.54\ci{0.38} & 14.37\ci{10.10} & 11.40\ci{4.82} & 7.06\ci{4.87} & 58.5\ci{6.1} & 43.8\ci{4.4} & 66.3\ci{21.2} & 63.2\ci{5.9} & 46.7\ci{3.4} \\
\midrule
\multicolumn{11}{l}{\textit{Globally destructive class}} \\
LoKU+FILA              & 13.56\ci{1.49} & 14.86\ci{1.54} & 13.66\ci{3.17} & 15.17\ci{2.32} & 12.20\ci{1.13} & 58.5\ci{1.0} & 56.7\ci{0.6} & 55.5\ci{4.4} & 54.9\ci{1.2} & 53.4\ci{2.4} \\
GA                     & 5.49\ci{5.29} & 2.84\ci{0.36} & 7.34\ci{9.27} & 6.37\ci{2.78} & 15.26\ci{16.14} & 58.3\ci{7.8} & 53.6\ci{1.9} & 57.9\ci{5.6} & 61.8\ci{2.1} & 63.4\ci{15.4} \\
GA ($\nu$)             & 5.19\ci{9.56} & 3.52\ci{2.00} & 6.09\ci{5.94} & 5.80\ci{3.20} & 11.31\ci{5.08} & 59.2\ci{5.3} & 52.6\ci{1.3} & 57.8\ci{2.8} & 60.0\ci{1.8} & 62.5\ci{6.5} \\
GD                     & 3.09\ci{1.42} & 2.53\ci{0.54} & 4.61\ci{5.06} & 4.76\ci{2.55} & 11.40\ci{7.10} & 56.7\ci{4.9} & 51.8\ci{0.3} & 56.1\ci{2.0} & 57.1\ci{2.8} & 64.8\ci{9.9} \\
GD ($\nu$)             & 2.67\ci{0.83} & 2.56\ci{0.30} & 5.59\ci{5.54} & 5.80\ci{2.90} & 10.25\ci{4.30} & 55.8\ci{5.2} & 50.4\ci{0.6} & 57.3\ci{3.7} & 59.0\ci{5.9} & 67.2\ci{8.7} \\
SPUL                   & 4.86\ci{0.27} & 5.62\ci{0.49} & 5.11\ci{0.38} & 5.96\ci{0.85} & 21.87\ci{8.34} & 58.7\ci{1.5} & 55.4\ci{0.8} & 57.0\ci{2.2} & 57.0\ci{2.4} & 66.2\ci{5.8} \\
\bottomrule
\end{tabular}}
\end{table}

\begin{table}[t]
\centering
\footnotesize
\setlength{\tabcolsep}{1.5pt}
\caption{WMDP per-topic CoFi and CHess relative shifts (\%) on \textbf{Llama-3.2-3B}, mean $\pm$ 95\% CI over three random subsets of 200 samples. $\uparrow$ on the forget splits = stronger unlearning; $\downarrow$ on the retain splits and on \textit{wiki} = less collateral damage. Methods are grouped by the TRIAGE class they fall into \emph{for this model}; the assignment uses the CoFi aggregates alone. ($\nu$) marks the randomly-perturbed variant.}
\label{tab:wmdp-cofi-chess-llama3b}
\resizebox{\textwidth}{!}{%
\begin{tabular}{l rrrrr @{\hspace{4pt}} rrrrr}
\toprule
& \multicolumn{5}{c}{\textbf{CoFi (\%)}} & \multicolumn{5}{c}{\textbf{CHess (\%)}} \\
\cmidrule(lr){2-6}\cmidrule(lr){7-11}
\textbf{Method} & \textit{bio-F} & \textit{cyber-F} & \textit{bio-R} & \textit{cyber-R} & \textit{wiki} & \textit{bio-F} & \textit{cyber-F} & \textit{bio-R} & \textit{cyber-R} & \textit{wiki} \\
\midrule
\multicolumn{11}{l}{\textit{No-op class}} \\
ATU                    & 0.03\ci{0.00} & 0.03\ci{0.00} & 0.03\ci{0.00} & 0.03\ci{0.01} & 0.06\ci{0.06} & 20.8\ci{0.3} & 23.5\ci{0.5} & 22.2\ci{0.5} & 22.6\ci{1.4} & 25.9\ci{0.3} \\
Obliviate              & 0.09\ci{0.00} & 0.09\ci{0.02} & 0.08\ci{0.02} & 0.09\ci{0.02} & 0.48\ci{0.18} & 21.2\ci{0.3} & 23.9\ci{0.5} & 22.5\ci{0.5} & 22.9\ci{1.3} & 27.3\ci{0.2} \\
RSV                    & 0.03\ci{0.00} & 0.03\ci{0.00} & 0.03\ci{0.00} & 0.03\ci{0.01} & 0.06\ci{0.06} & 20.8\ci{0.3} & 23.5\ci{0.6} & 22.1\ci{0.5} & 22.6\ci{1.5} & 25.8\ci{0.3} \\
RSV ($\nu$)            & 0.03\ci{0.00} & 0.03\ci{0.00} & 0.03\ci{0.01} & 0.03\ci{0.00} & 0.08\ci{0.11} & 20.8\ci{0.3} & 23.5\ci{0.5} & 22.2\ci{0.5} & 22.6\ci{1.3} & 25.9\ci{0.3} \\
\midrule
\multicolumn{11}{l}{\textit{Partially-localized class}} \\
RMU                    & 5.30\ci{0.10} & 6.51\ci{0.53} & 4.63\ci{0.41} & 3.57\ci{4.52} & 0.07\ci{0.06} & 43.2\ci{2.7} & 43.7\ci{1.6} & 37.5\ci{4.3} & 33.4\ci{9.6} & 25.8\ci{0.3} \\
RMU ($\nu$)            & 5.28\ci{0.10} & 6.48\ci{0.54} & 4.61\ci{0.44} & 3.55\ci{4.58} & 0.08\ci{0.11} & 42.2\ci{1.7} & 42.8\ci{1.0} & 36.5\ci{4.0} & 32.5\ci{9.7} & 25.8\ci{0.2} \\
\midrule
\multicolumn{11}{l}{\textit{Collateral-dominant class}} \\
Adaptive-RMU           & 3.85\ci{0.09} & 3.54\ci{0.31} & 4.26\ci{2.47} & 5.21\ci{2.28} & 0.83\ci{1.79} & 36.7\ci{2.2} & 36.7\ci{0.9} & 38.9\ci{11.1} & 43.8\ci{2.0} & 26.5\ci{1.8} \\
Adaptive-RMU ($\nu$)   & 3.89\ci{0.08} & 3.50\ci{0.31} & 4.26\ci{2.53} & 5.14\ci{2.44} & 0.83\ci{1.85} & 37.0\ci{1.2} & 36.7\ci{1.5} & 38.1\ci{10.0} & 43.4\ci{1.6} & 26.9\ci{2.1} \\
NPO                    & 7.36\ci{2.36} & 5.84\ci{0.18} & 12.51\ci{9.84} & 10.80\ci{4.45} & 3.39\ci{5.07} & 57.5\ci{6.6} & 46.6\ci{2.1} & 61.8\ci{16.7} & 63.7\ci{2.5} & 33.5\ci{17.8} \\
NPO ($\nu$)            & 8.13\ci{2.94} & 5.47\ci{0.38} & 13.02\ci{10.34} & 14.23\ci{4.74} & 4.14\ci{10.48} & 59.1\ci{6.1} & 48.3\ci{2.1} & 62.1\ci{17.3} & 65.2\ci{2.2} & 35.4\ci{18.3} \\
SimNPO                 & 9.86\ci{4.20} & 6.40\ci{0.29} & 13.07\ci{8.26} & 10.34\ci{4.06} & 3.43\ci{8.26} & 58.3\ci{6.6} & 45.6\ci{2.1} & 61.1\ci{15.0} & 59.9\ci{2.2} & 34.1\ci{17.7} \\
SimNPO ($\nu$)         & 7.03\ci{2.19} & 5.58\ci{0.19} & 12.19\ci{9.97} & 10.77\ci{4.93} & 3.33\ci{4.00} & 57.0\ci{5.4} & 45.6\ci{2.4} & 60.9\ci{17.8} & 63.0\ci{4.4} & 36.2\ci{21.3} \\
DPO                    & 6.99\ci{4.04} & 3.55\ci{0.67} & 12.65\ci{4.65} & 14.39\ci{2.56} & 4.24\ci{3.25} & 56.0\ci{9.8} & 44.8\ci{4.0} & 63.1\ci{6.1} & 65.3\ci{1.4} & 38.1\ci{9.2} \\
DPO ($\nu$)            & 5.46\ci{3.65} & 4.41\ci{1.02} & 11.82\ci{4.17} & 13.35\ci{4.50} & 5.11\ci{5.36} & 52.9\ci{9.6} & 46.2\ci{4.1} & 61.4\ci{5.4} & 64.5\ci{2.8} & 39.0\ci{11.2} \\
\midrule
\multicolumn{11}{l}{\textit{Globally destructive class}} \\
LoKU+FILA              & 4.42\ci{0.39} & 4.65\ci{0.26} & 4.58\ci{0.13} & 4.51\ci{0.21} & 4.16\ci{0.57} & 45.6\ci{1.4} & 44.2\ci{0.5} & 44.7\ci{3.4} & 44.9\ci{2.4} & 38.4\ci{1.1} \\
GA                     & 2.85\ci{4.19} & 2.31\ci{1.03} & 4.97\ci{2.50} & 11.61\ci{9.30} & 9.93\ci{10.56} & 48.4\ci{5.8} & 44.3\ci{4.0} & 51.3\ci{8.5} & 60.8\ci{10.3} & 50.6\ci{13.9} \\
GA ($\nu$)             & 2.79\ci{5.07} & 2.47\ci{1.39} & 5.16\ci{2.25} & 11.57\ci{9.88} & 9.06\ci{10.15} & 48.3\ci{6.3} & 44.4\ci{1.1} & 51.4\ci{5.0} & 60.5\ci{9.2} & 50.7\ci{13.2} \\
GD                     & 2.75\ci{4.43} & 2.42\ci{1.10} & 5.74\ci{0.80} & 12.17\ci{10.05} & 8.78\ci{8.13} & 49.0\ci{4.8} & 45.3\ci{7.8} & 49.0\ci{2.5} & 57.8\ci{7.5} & 52.5\ci{8.7} \\
GD ($\nu$)             & 2.78\ci{4.29} & 2.27\ci{0.84} & 4.12\ci{1.75} & 11.75\ci{9.78} & 9.29\ci{10.10} & 48.0\ci{6.0} & 44.5\ci{0.3} & 51.8\ci{3.6} & 61.4\ci{12.3} & 49.7\ci{11.4} \\
SPUL                   & 5.24\ci{0.50} & 5.09\ci{0.09} & 5.01\ci{0.61} & 7.46\ci{3.53} & 10.48\ci{8.32} & 55.9\ci{3.3} & 50.3\ci{0.9} & 53.7\ci{0.9} & 57.4\ci{4.6} & 46.4\ci{12.9} \\
\bottomrule
\end{tabular}}
\end{table}

\begin{table}[t]
\centering
\footnotesize
\setlength{\tabcolsep}{1.5pt}
\caption{WMDP per-topic CoFi and CHess relative shifts (\%) on \textbf{Qwen3-32B}, mean $\pm$ 95\% CI over three random subsets of 200 samples. $\uparrow$ on the forget splits = stronger unlearning; $\downarrow$ on the retain splits and on \textit{wiki} = less collateral damage. Methods are grouped by the TRIAGE class they fall into \emph{for this model}; the assignment uses the CoFi aggregates alone. ($\nu$) marks the randomly-perturbed variant.}
\label{tab:wmdp-cofi-chess-qwen32b}
\resizebox{\textwidth}{!}{%
\begin{tabular}{l rrrrr @{\hspace{4pt}} rrrrr}
\toprule
& \multicolumn{5}{c}{\textbf{CoFi (\%)}} & \multicolumn{5}{c}{\textbf{CHess (\%)}} \\
\cmidrule(lr){2-6}\cmidrule(lr){7-11}
\textbf{Method} & \textit{bio-F} & \textit{cyber-F} & \textit{bio-R} & \textit{cyber-R} & \textit{wiki} & \textit{bio-F} & \textit{cyber-F} & \textit{bio-R} & \textit{cyber-R} & \textit{wiki} \\
\midrule
\multicolumn{11}{l}{\textit{No-op class}} \\
ATU                    & 0.03\ci{0.01} & 0.03\ci{0.00} & 0.03\ci{0.00} & 0.04\ci{0.02} & 0.24\ci{0.04} & 18.1\ci{4.0} & 21.1\ci{0.5} & 24.2\ci{20.8} & 21.8\ci{1.5} & 26.4\ci{4.4} \\
Obliviate              & 0.04\ci{0.03} & 0.03\ci{0.00} & 0.03\ci{0.01} & 0.03\ci{0.01} & 0.32\ci{0.05} & 20.2\ci{5.0} & 21.3\ci{0.9} & 20.5\ci{2.4} & 22.1\ci{1.5} & 26.5\ci{3.2} \\
RSV                    & 0.03\ci{0.01} & 0.05\ci{0.05} & 0.03\ci{0.01} & 0.04\ci{0.04} & 0.24\ci{0.05} & 18.8\ci{5.2} & 21.2\ci{0.4} & 19.6\ci{2.7} & 21.9\ci{1.6} & 26.5\ci{4.6} \\
RSV ($\nu$)            & 0.03\ci{0.01} & 0.03\ci{0.00} & 0.03\ci{0.00} & 0.03\ci{0.02} & 0.24\ci{0.05} & 18.5\ci{5.2} & 21.6\ci{1.9} & 20.4\ci{2.2} & 21.8\ci{1.6} & 26.4\ci{4.4} \\
RMU                    & 0.03\ci{0.01} & 0.09\ci{0.04} & 0.03\ci{0.01} & 0.03\ci{0.01} & 0.24\ci{0.05} & 18.7\ci{3.8} & 22.2\ci{1.1} & 19.6\ci{2.1} & 21.9\ci{1.2} & 26.4\ci{4.5} \\
RMU ($\nu$)            & 0.03\ci{0.02} & 0.09\ci{0.02} & 0.04\ci{0.04} & 0.03\ci{0.01} & 0.36\ci{0.48} & 19.3\ci{3.3} & 22.0\ci{0.5} & 19.5\ci{2.5} & 22.0\ci{1.5} & 26.4\ci{4.4} \\
Adaptive-RMU           & 0.73\ci{0.08} & 0.93\ci{0.02} & 0.50\ci{0.09} & 0.23\ci{0.11} & 0.24\ci{0.04} & 39.0\ci{8.9} & 31.2\ci{2.4} & 32.2\ci{2.2} & 26.2\ci{2.5} & 25.8\ci{1.8} \\
Adaptive-RMU ($\nu$)   & 1.40\ci{2.91} & 0.95\ci{0.02} & 0.53\ci{0.08} & 0.26\ci{0.07} & 0.24\ci{0.05} & 39.0\ci{5.3} & 31.1\ci{1.6} & 32.9\ci{1.9} & 26.3\ci{2.4} & 26.4\ci{4.5} \\
\midrule
\multicolumn{11}{l}{\textit{Collateral-dominant class}} \\
NPO                    & 13.26\ci{2.57} & 6.66\ci{1.22} & 11.12\ci{4.92} & 14.04\ci{0.65} & 0.24\ci{0.05} & 74.2\ci{16.6} & 60.0\ci{3.0} & 68.6\ci{5.3} & 70.7\ci{4.6} & 26.5\ci{4.5} \\
NPO ($\nu$)            & 13.14\ci{3.53} & 6.17\ci{5.10} & 11.75\ci{3.94} & 15.36\ci{7.02} & 0.24\ci{0.05} & 72.1\ci{5.6} & 57.0\ci{0.7} & 69.1\ci{6.0} & 70.7\ci{2.0} & 26.4\ci{4.3} \\
SimNPO                 & 12.48\ci{2.40} & 5.18\ci{1.99} & 10.39\ci{5.34} & 14.06\ci{2.12} & 0.24\ci{0.05} & 70.8\ci{5.8} & 57.0\ci{10.3} & 68.3\ci{2.0} & 71.2\ci{4.3} & 26.3\ci{4.1} \\
SimNPO ($\nu$)         & 13.17\ci{2.13} & 5.66\ci{2.21} & 15.72\ci{18.38} & 14.44\ci{2.06} & 0.24\ci{0.05} & 72.1\ci{9.4} & 55.1\ci{5.0} & 72.3\ci{9.8} & 70.7\ci{3.1} & 26.4\ci{4.3} \\
DPO                    & 12.73\ci{1.93} & 4.99\ci{2.06} & 13.67\ci{4.59} & 16.76\ci{3.18} & 0.24\ci{0.05} & 68.8\ci{1.0} & 50.3\ci{8.7} & 68.9\ci{2.9} & 68.7\ci{3.5} & 26.4\ci{4.4} \\
DPO ($\nu$)            & 13.58\ci{1.89} & 4.89\ci{1.88} & 14.64\ci{3.41} & 15.12\ci{2.40} & 0.24\ci{0.05} & 71.4\ci{1.8} & 55.5\ci{3.3} & 72.8\ci{7.6} & 71.0\ci{12.6} & 26.4\ci{4.5} \\
LoKU (no FILA)         & 1.82\ci{0.27} & 1.93\ci{0.22} & 1.96\ci{0.87} & 2.60\ci{1.40} & 0.23\ci{0.13} & 43.4\ci{6.8} & 40.2\ci{1.9} & 48.3\ci{6.6} & 48.9\ci{7.8} & 32.3\ci{11.9} \\
GA                     & 1.46\ci{0.48} & 2.51\ci{1.54} & 5.18\ci{10.16} & 6.85\ci{16.93} & 3.57\ci{2.12} & 45.2\ci{12.8} & 38.3\ci{3.7} & 47.0\ci{13.2} & 56.6\ci{15.9} & 27.9\ci{4.9} \\
GA ($\nu$)             & 1.62\ci{0.49} & 2.42\ci{2.19} & 5.24\ci{8.79} & 4.65\ci{2.42} & 3.27\ci{2.96} & 43.1\ci{2.3} & 39.6\ci{8.9} & 52.4\ci{13.0} & 61.6\ci{23.9} & 28.1\ci{4.7} \\
GD                     & 2.17\ci{0.75} & 2.73\ci{2.82} & 5.80\ci{12.63} & 2.67\ci{1.09} & 2.54\ci{3.13} & 48.8\ci{4.4} & 47.9\ci{32.0} & 47.1\ci{4.9} & 50.4\ci{17.1} & 31.4\ci{12.2} \\
GD ($\nu$)             & 2.29\ci{0.73} & 2.75\ci{2.53} & 8.01\ci{10.28} & 3.22\ci{1.48} & 2.30\ci{2.16} & 48.9\ci{5.1} & 40.1\ci{4.6} & 45.7\ci{4.7} & 46.6\ci{8.6} & 32.2\ci{10.2} \\
\midrule
\multicolumn{11}{l}{\textit{Globally destructive class}} \\
SPUL                   & 6.84\ci{0.65} & 12.60\ci{1.20} & 9.10\ci{3.56} & 10.80\ci{7.44} & 9.54\ci{2.16} & 66.8\ci{4.8} & 65.3\ci{6.0} & 64.6\ci{1.2} & 63.5\ci{0.7} & 49.3\ci{19.7} \\
\bottomrule
\end{tabular}}
\end{table}

\begin{table}[t]
\centering
\footnotesize
\setlength{\tabcolsep}{1.5pt}
\caption{WMDP per-topic CoFi and CHess relative shifts (\%) on \textbf{Zephyr-7B-$\beta$}, mean $\pm$ 95\% CI over three random subsets of 200 samples. $\uparrow$ on the forget splits = stronger unlearning; $\downarrow$ on the retain splits and on \textit{wiki} = less collateral damage. Methods are grouped by the TRIAGE class they fall into \emph{for this model}; the assignment uses the CoFi aggregates alone. ($\nu$) marks the randomly-perturbed variant.}
\label{tab:wmdp-cofi-chess-zephyr}
\resizebox{\textwidth}{!}{%
\begin{tabular}{l rrrrr @{\hspace{4pt}} rrrrr}
\toprule
& \multicolumn{5}{c}{\textbf{CoFi (\%)}} & \multicolumn{5}{c}{\textbf{CHess (\%)}} \\
\cmidrule(lr){2-6}\cmidrule(lr){7-11}
\textbf{Method} & \textit{bio-F} & \textit{cyber-F} & \textit{bio-R} & \textit{cyber-R} & \textit{wiki} & \textit{bio-F} & \textit{cyber-F} & \textit{bio-R} & \textit{cyber-R} & \textit{wiki} \\
\midrule
\multicolumn{11}{l}{\textit{No-op class}} \\
ATU                    & 0.04\ci{0.01} & 0.04\ci{0.00} & 0.04\ci{0.00} & 0.04\ci{0.00} & 0.28\ci{0.31} & 47.8\ci{0.9} & 49.9\ci{3.4} & 47.0\ci{2.4} & 45.8\ci{5.3} & 50.4\ci{2.9} \\
\midrule
\multicolumn{11}{l}{\textit{Partially-localized class}} \\
RSV                    & 17.69\ci{0.66} & 17.47\ci{0.76} & 18.17\ci{0.99} & 15.79\ci{5.83} & 3.00\ci{0.04} & 48.0\ci{0.8} & 51.8\ci{3.4} & 47.6\ci{0.8} & 46.3\ci{5.1} & 50.7\ci{2.3} \\
RSV ($\nu$)            & 17.38\ci{0.32} & 16.63\ci{0.58} & 17.87\ci{1.13} & 15.63\ci{6.71} & 2.25\ci{1.09} & 47.9\ci{0.8} & 52.2\ci{2.6} & 48.0\ci{1.7} & 46.8\ci{3.7} & 50.7\ci{2.6} \\
Adaptive-RMU           & 0.06\ci{0.01} & 2.83\ci{0.13} & 0.05\ci{0.01} & 0.06\ci{0.03} & 0.20\ci{0.16} & 47.7\ci{0.8} & 50.1\ci{2.6} & 46.8\ci{1.7} & 45.9\ci{4.1} & 50.4\ci{2.8} \\
Adaptive-RMU ($\nu$)   & 0.08\ci{0.00} & 2.92\ci{0.14} & 0.06\ci{0.04} & 0.07\ci{0.03} & 0.28\ci{0.21} & 47.8\ci{0.7} & 50.4\ci{1.4} & 46.8\ci{1.1} & 45.8\ci{4.0} & 50.4\ci{2.8} \\
\midrule
\multicolumn{11}{l}{\textit{Collateral-dominant class}} \\
Obliviate              & 6.27\ci{0.95} & 7.59\ci{1.53} & 7.81\ci{0.96} & 9.99\ci{1.11} & 1.50\ci{0.63} & 58.0\ci{1.2} & 59.1\ci{0.6} & 57.5\ci{0.7} & 58.5\ci{4.1} & 56.2\ci{0.5} \\
RMU                    & 28.30\ci{0.13} & 15.90\ci{1.29} & 25.74\ci{1.31} & 19.36\ci{0.86} & 2.53\ci{2.20} & 51.5\ci{3.5} & 59.4\ci{1.9} & 51.7\ci{3.3} & 49.9\ci{7.7} & 50.7\ci{2.5} \\
RMU ($\nu$)            & 26.86\ci{0.04} & 16.19\ci{1.47} & 24.66\ci{1.29} & 19.31\ci{1.36} & 2.73\ci{3.99} & 52.8\ci{2.9} & 59.3\ci{5.0} & 51.6\ci{1.8} & 49.5\ci{9.3} & 50.7\ci{2.3} \\
NPO                    & 24.66\ci{0.31} & 26.19\ci{2.89} & 28.66\ci{8.78} & 41.29\ci{2.65} & 16.92\ci{9.45} & 95.3\ci{2.3} & 112.5\ci{7.1} & 100.6\ci{14.5} & 105.4\ci{6.8} & 66.2\ci{27.1} \\
NPO ($\nu$)            & 30.02\ci{0.94} & 30.01\ci{9.93} & 31.22\ci{6.52} & 41.60\ci{5.02} & 14.92\ci{9.64} & 94.0\ci{10.9} & 103.4\ci{6.4} & 93.3\ci{12.0} & 102.3\ci{1.4} & 68.3\ci{22.5} \\
SimNPO                 & 29.16\ci{0.24} & 30.95\ci{11.14} & 32.16\ci{8.51} & 42.37\ci{5.27} & 14.31\ci{9.61} & 98.9\ci{13.9} & 108.7\ci{16.1} & 94.0\ci{16.4} & 93.2\ci{8.6} & 65.6\ci{31.7} \\
SimNPO ($\nu$)         & 30.07\ci{4.82} & 30.12\ci{4.07} & 35.56\ci{7.78} & 44.15\ci{1.65} & 7.95\ci{27.61} & 81.5\ci{7.2} & 93.4\ci{8.8} & 90.2\ci{10.8} & 90.7\ci{15.2} & 53.5\ci{3.8} \\
DPO                    & 8.22\ci{1.91} & 8.87\ci{8.22} & 12.22\ci{8.43} & 26.10\ci{4.66} & 10.62\ci{6.35} & 66.6\ci{7.3} & 88.2\ci{10.8} & 83.5\ci{12.1} & 76.9\ci{22.7} & 52.3\ci{6.3} \\
DPO ($\nu$)            & 27.57\ci{0.42} & 30.69\ci{3.93} & 35.30\ci{15.76} & 36.13\ci{11.06} & 1.52\ci{2.42} & 49.6\ci{3.8} & 53.5\ci{4.1} & 48.7\ci{2.7} & 47.5\ci{4.7} & 50.8\ci{2.6} \\
\midrule
\multicolumn{11}{l}{\textit{Globally destructive class}} \\
LoKU+FILA              & 14.03\ci{1.08} & 15.06\ci{0.07} & 15.03\ci{6.82} & 14.73\ci{4.86} & 12.84\ci{1.13} & 64.3\ci{1.9} & 70.4\ci{0.4} & 64.4\ci{3.3} & 65.6\ci{2.9} & 69.6\ci{8.3} \\
GA                     & 5.43\ci{5.81} & 4.71\ci{3.27} & 13.34\ci{18.87} & 8.85\ci{9.90} & 14.74\ci{28.37} & 149.7\ci{1.8} & 156.6\ci{8.4} & 150.3\ci{5.7} & 150.1\ci{8.6} & 134.4\ci{3.4} \\
GA ($\nu$)             & 3.58\ci{2.86} & 4.04\ci{1.96} & 11.63\ci{24.24} & 6.29\ci{3.40} & 20.98\ci{8.84} & 156.0\ci{5.0} & 161.0\ci{9.8} & 154.9\ci{1.1} & 154.5\ci{9.8} & 145.1\ci{6.4} \\
GD                     & 4.32\ci{2.84} & 3.61\ci{0.32} & 4.03\ci{1.78} & 5.33\ci{3.39} & 23.84\ci{20.35} & 148.9\ci{1.3} & 154.2\ci{4.1} & 149.3\ci{4.5} & 148.9\ci{7.2} & 126.0\ci{6.7} \\
GD ($\nu$)             & 2.74\ci{1.07} & 4.18\ci{1.35} & 2.89\ci{0.33} & 4.24\ci{0.51} & 34.01\ci{7.43} & 157.8\ci{2.9} & 163.0\ci{4.3} & 154.4\ci{2.8} & 155.0\ci{2.6} & 154.6\ci{0.6} \\
SPUL                   & 17.40\ci{0.69} & 14.58\ci{1.31} & 16.39\ci{0.95} & 22.47\ci{3.40} & 38.13\ci{3.82} & 120.8\ci{1.5} & 122.9\ci{3.7} & 119.1\ci{2.0} & 120.4\ci{4.2} & 115.5\ci{2.8} \\
\bottomrule
\end{tabular}}
\end{table}

\begin{table}[t]
\centering
\small
\setlength{\tabcolsep}{4pt}
\caption{WMDP perplexity ratios (unlearned/original), reported as $\log_{10}$ of the ratio, for Llama-3.1-8B and Llama-3.2-3B. $0.00$ means fluency is unchanged; $\uparrow$ on the forget splits is intended, while any large value on the retain splits or on \textit{wiki} is pure damage. Perplexity is reported as a fluency sanity check and takes no part in the TRIAGE class assignment, which is made from CoFi. Subset CIs are omitted because the raw ratio spans up to $10^{112}$ and its linear-scale CI is of the same order as the mean; the corresponding CoFi/CHess CIs are given in Tables~\ref{tab:wmdp-cofi-chess-llama8b}--\ref{tab:wmdp-cofi-chess-llama3b}. Rows follow a fixed method order so the two models can be read off against each other; the TRIAGE class each method falls into is listed in Table~\ref{tab:triage-class-summary}.}
\label{tab:wmdp-ppl-1}
\begin{tabular}{l rrrrr @{\hspace{7pt}} rrrrr}
\toprule
& \multicolumn{5}{c}{\textbf{Llama-3.1-8B}} & \multicolumn{5}{c}{\textbf{Llama-3.2-3B}} \\
\cmidrule(lr){2-6}\cmidrule(lr){7-11}
\textbf{Method} & \textit{bio-F} & \textit{cyber-F} & \textit{bio-R} & \textit{cyber-R} & \textit{wiki} & \textit{bio-F} & \textit{cyber-F} & \textit{bio-R} & \textit{cyber-R} & \textit{wiki} \\
\midrule
ATU                    & 0.00 & 0.00 & 0.00 & 0.00 & 0.00 & 0.00 & 0.00 & 0.00 & 0.00 & 0.00 \\
Obliviate              & 0.06 & 0.04 & 0.05 & 0.03 & 0.24 & 0.06 & 0.04 & 0.05 & 0.04 & 0.21 \\
RSV                    & 0.00 & 0.00 & 0.00 & 0.00 & 0.00 & 0.00 & 0.00 & 0.00 & 0.00 & 0.00 \\
RSV ($\nu$)            & 0.00 & 0.00 & 0.00 & 0.00 & 0.00 & 0.00 & 0.00 & 0.00 & 0.00 & 0.00 \\
RMU                    & 3.40 & 3.80 & 3.08 & 1.41 & 0.00 & 2.55 & 2.55 & 1.89 & 1.09 & 0.00 \\
RMU ($\nu$)            & 3.38 & 3.81 & 3.05 & 1.35 & 0.00 & 2.56 & 2.51 & 1.91 & 1.09 & 0.00 \\
Adaptive-RMU           & 5.03 & 5.37 & 5.18 & 4.65 & 0.01 & 3.89 & 4.18 & 4.10 & 3.88 & 0.01 \\
Adaptive-RMU ($\nu$)   & 5.05 & 5.36 & 5.17 & 4.70 & 0.01 & 3.90 & 4.18 & 4.10 & 3.88 & 0.02 \\
NPO                    & 31.33 & 31.40 & 30.79 & 28.84 & 0.03 & 31.92 & 30.50 & 31.83 & 30.75 & 0.85 \\
NPO ($\nu$)            & 31.02 & 30.70 & 30.53 & 28.73 & 0.04 & 32.66 & 32.00 & 32.58 & 31.40 & 0.35 \\
SimNPO                 & 30.10 & 29.74 & 29.69 & 27.81 & 0.04 & 29.20 & 26.86 & 29.18 & 28.10 & 0.71 \\
SimNPO ($\nu$)         & 30.87 & 31.14 & 30.42 & 28.54 & 0.06 & 31.62 & 30.35 & 31.49 & 30.46 & 0.91 \\
DPO                    & 30.14 & 30.94 & 30.03 & 25.95 & 0.00 & 31.06 & 31.08 & 30.86 & 29.46 & 0.09 \\
DPO ($\nu$)            & 34.08 & 34.61 & 32.30 & 29.65 & 0.02 & 30.26 & 30.33 & 30.31 & 28.51 & 0.02 \\
LoKU+FILA              & 3.79 & 4.59 & 4.18 & 4.54 & 3.39 & 3.97 & 4.34 & 4.22 & 4.35 & 3.40 \\
GA                     & 69.98 & 74.80 & 70.15 & 70.11 & 2.80 & 36.47 & 36.77 & 36.29 & 35.84 & 3.89 \\
GA ($\nu$)             & 70.03 & 74.80 & 70.19 & 70.20 & 3.73 & 36.47 & 36.79 & 36.29 & 35.84 & 3.73 \\
GD                     & 70.05 & 74.68 & 70.14 & 70.47 & 2.06 & 36.44 & 36.75 & 36.27 & 35.81 & 3.67 \\
GD ($\nu$)             & 70.06 & 74.73 & 70.14 & 70.50 & 2.09 & 36.47 & 36.74 & 36.29 & 35.84 & 4.11 \\
SPUL                   & 33.32 & 33.02 & 33.71 & 32.96 & 29.14 & 14.81 & 13.22 & 15.22 & 13.89 & 4.38 \\
\bottomrule
\end{tabular}
\end{table}

\begin{table}[t]
\centering
\small
\setlength{\tabcolsep}{4pt}
\caption{WMDP perplexity ratios (unlearned/original), reported as $\log_{10}$ of the ratio, for Qwen3-32B and Zephyr-7B-$\beta$. $0.00$ means fluency is unchanged; $\uparrow$ on the forget splits is intended, while any large value on the retain splits or on \textit{wiki} is pure damage. Perplexity is reported as a fluency sanity check and takes no part in the TRIAGE class assignment, which is made from CoFi. Subset CIs are omitted because the raw ratio spans up to $10^{112}$ and its linear-scale CI is of the same order as the mean; the corresponding CoFi/CHess CIs are given in Tables~\ref{tab:wmdp-cofi-chess-qwen32b}--\ref{tab:wmdp-cofi-chess-zephyr}. Rows follow a fixed method order so the two models can be read off against each other; the TRIAGE class each method falls into is listed in Table~\ref{tab:triage-class-summary}.}
\label{tab:wmdp-ppl-2}
\begin{tabular}{l rrrrr @{\hspace{7pt}} rrrrr}
\toprule
& \multicolumn{5}{c}{\textbf{Qwen3-32B}} & \multicolumn{5}{c}{\textbf{Zephyr-7B-$\beta$}} \\
\cmidrule(lr){2-6}\cmidrule(lr){7-11}
\textbf{Method} & \textit{bio-F} & \textit{cyber-F} & \textit{bio-R} & \textit{cyber-R} & \textit{wiki} & \textit{bio-F} & \textit{cyber-F} & \textit{bio-R} & \textit{cyber-R} & \textit{wiki} \\
\midrule
ATU                    & 0.00 & 0.00 & 0.00 & 0.00 & 0.00 & 0.00 & 0.00 & 0.00 & 0.00 & 0.00 \\
Obliviate              & 0.03 & 0.01 & 0.02 & 0.02 & 0.11 & 3.53 & 3.64 & 3.29 & 3.06 & 0.28 \\
RSV                    & 0.00 & 0.00 & 0.00 & 0.00 & 0.00 & 4.54 & 3.79 & 3.75 & 2.48 & 0.02 \\
RSV ($\nu$)            & 0.00 & 0.00 & 0.00 & 0.00 & 0.00 & 4.42 & 3.81 & 3.56 & 2.60 & 0.03 \\
RMU                    & 0.00 & 0.13 & 0.00 & 0.00 & 0.00 & 5.48 & 3.63 & 4.45 & 3.17 & 0.04 \\
RMU ($\nu$)            & 0.00 & 0.13 & 0.00 & 0.00 & 0.00 & 5.52 & 3.61 & 4.50 & 3.14 & 0.03 \\
Adaptive-RMU           & 3.21 & 3.05 & 1.83 & 0.55 & 0.00 & 0.01 & 0.16 & 0.00 & 0.00 & 0.00 \\
Adaptive-RMU ($\nu$)   & 3.21 & 3.06 & 1.93 & 0.61 & 0.00 & 0.01 & 0.18 & 0.00 & 0.00 & 0.00 \\
NPO                    & 40.12 & 39.16 & 21.57 & 37.51 & 0.08 & 57.88 & 58.24 & 58.71 & 60.45 & 1.95 \\
NPO ($\nu$)            & 39.75 & 40.24 & 18.64 & 37.03 & 0.06 & 56.34 & 56.21 & 56.89 & 59.25 & 2.44 \\
SimNPO                 & 39.27 & 39.55 & 23.29 & 36.24 & 0.08 & 57.63 & 58.33 & 58.40 & 60.27 & 2.90 \\
SimNPO ($\nu$)         & 41.07 & 38.25 & 24.76 & 38.24 & 0.08 & 54.23 & 53.77 & 55.57 & 58.21 & 0.39 \\
DPO                    & 45.81 & 41.51 & 24.89 & 42.07 & 0.05 & 27.60 & 29.00 & 27.82 & 27.72 & 0.71 \\
DPO ($\nu$)            & 45.93 & 42.94 & 31.25 & 45.42 & 0.06 & 51.77 & 44.38 & 53.11 & 53.78 & 0.52 \\
LoKU+FILA              & \textendash & \textendash & \textendash & \textendash & \textendash & 3.78 & 4.63 & 3.74 & 3.80 & 3.10 \\
LoKU (no FILA)         & 11.68 & 13.06 & 11.83 & 11.27 & -0.05 & \textendash & \textendash & \textendash & \textendash & \textendash \\
GA                     & 70.60 & 76.25 & 74.44 & 74.49 & 1.02 & 112.05 & 112.20 & 112.28 & 112.64 & 1.12 \\
GA ($\nu$)             & 70.13 & 75.68 & 73.58 & 73.85 & 1.03 & 112.06 & 112.20 & 112.30 & 112.65 & 1.49 \\
GD                     & 62.75 & 75.50 & 65.01 & 62.58 & 1.46 & 112.08 & 112.16 & 112.30 & 112.64 & 1.56 \\
GD ($\nu$)             & 62.81 & 75.61 & 65.13 & 62.95 & 1.83 & 112.08 & 112.13 & 112.31 & 112.64 & 3.66 \\
SPUL                   & 35.07 & 50.78 & 41.62 & 36.40 & 27.75 & 33.48 & 19.83 & 33.56 & 33.45 & 30.08 \\
\bottomrule
\end{tabular}
\end{table}

\begin{table}[t]
\centering
\small
\setlength{\tabcolsep}{3pt}
\caption{TRIAGE aggregates for the three models not shown in Table~\ref{tab:headline}. Conventions are identical: CoFi and CHess are relative shifts (\%), mean $\pm$ 95\% CI over three random subsets of 200 samples; the PPL ratio is unlearned/original (1.0 = unchanged), geometric mean over the per-topic corpora. $\mathcal{C}_F$ and $\mathcal{C}_A$ average the bio and cyber splits, $\mathcal{C}_G$ is WikiText. Methods are grouped by their TRIAGE class, which is assigned per model from the CoFi aggregates alone.}
\label{tab:wmdp-aggregate-other}
\begin{tabular}{l rrr rrr rrr}
\toprule
& \multicolumn{3}{c}{\textbf{CoFi (\%)}} & \multicolumn{3}{c}{\textbf{CHess (\%)}} & \multicolumn{3}{c}{\textbf{PPL ratio}} \\
\cmidrule(lr){2-4}\cmidrule(lr){5-7}\cmidrule(lr){8-10}
\textbf{Method} & $\mathcal{C}_F$ & $\mathcal{C}_A$ & $\mathcal{C}_G$ & $\mathcal{C}_F$ & $\mathcal{C}_A$ & $\mathcal{C}_G$ & $\mathcal{C}_F$ & $\mathcal{C}_A$ & $\mathcal{C}_G$ \\
\midrule
\multicolumn{10}{c}{\textbf{Llama-3.2-3B}} \\
\midrule
\multicolumn{10}{l}{\textit{No-op class}} \\
ATU                    & 0.03\ci{0.00} & 0.03\ci{0.01} & 0.06\ci{0.06} & 22.1\ci{0.3} & 22.4\ci{0.7} & 25.9\ci{0.3} & 1.0 & 1.0 & 1.0 \\
Obliviate              & 0.09\ci{0.01} & 0.08\ci{0.01} & 0.48\ci{0.18} & 22.6\ci{0.3} & 22.7\ci{0.7} & 27.3\ci{0.2} & 1.1 & 1.0 & 1.6 \\
RSV                    & 0.03\ci{0.00} & 0.03\ci{0.01} & 0.06\ci{0.06} & 22.1\ci{0.3} & 22.4\ci{0.8} & 25.8\ci{0.3} & 1.0 & 1.0 & 1.0 \\
\multicolumn{10}{l}{\textit{Partially-localized class}} \\
RMU                    & 5.90\ci{0.27} & 4.10\ci{2.27} & 0.07\ci{0.06} & 43.5\ci{1.6} & 35.4\ci{5.3} & 25.8\ci{0.3} & $10^{3}$ & $10^{1}$ & 1.0 \\
\multicolumn{10}{l}{\textit{Collateral-dominant class}} \\
Adaptive-RMU           & 3.70\ci{0.16} & 4.73\ci{1.68} & 0.83\ci{1.79} & 36.7\ci{1.2} & 41.3\ci{5.6} & 26.5\ci{1.8} & $10^{4}$ & $10^{4}$ & 1.0 \\
NPO                    & 6.60\ci{1.18} & 11.66\ci{5.40} & 3.39\ci{5.07} & 52.0\ci{3.5} & 62.8\ci{8.4} & 33.5\ci{17.8} & $10^{31}$ & $10^{31}$ & 7.1 \\
SimNPO                 & 8.13\ci{2.10} & 11.71\ci{4.60} & 3.43\ci{8.26} & 51.9\ci{3.5} & 60.5\ci{7.6} & 34.1\ci{17.7} & $10^{28}$ & $10^{29}$ & 5.2 \\
DPO                    & 5.27\ci{2.05} & 13.52\ci{2.65} & 4.24\ci{3.25} & 50.4\ci{5.3} & 64.2\ci{3.1} & 38.1\ci{9.2} & $10^{31}$ & $10^{30}$ & 1.2 \\
\multicolumn{10}{l}{\textit{Globally destructive class}} \\
LoKU+FILA              & 4.54\ci{0.23} & 4.54\ci{0.12} & 4.16\ci{0.57} & 44.9\ci{0.7} & 44.8\ci{2.1} & 38.4\ci{1.1} & $10^{4}$ & $10^{4}$ & $10^{3}$ \\
GA                     & 2.58\ci{2.16} & 8.29\ci{4.82} & 9.93\ci{10.56} & 46.3\ci{3.5} & 56.0\ci{6.7} & 50.6\ci{13.9} & $10^{37}$ & $10^{36}$ & $10^{4}$ \\
GD                     & 2.58\ci{2.28} & 8.96\ci{5.04} & 8.78\ci{8.13} & 47.2\ci{4.6} & 53.4\ci{4.0} & 52.5\ci{8.7} & $10^{37}$ & $10^{36}$ & $10^{4}$ \\
SPUL                   & 5.17\ci{0.25} & 6.23\ci{1.79} & 10.48\ci{8.32} & 53.1\ci{1.7} & 55.6\ci{2.3} & 46.4\ci{12.9} & $10^{14}$ & $10^{15}$ & $10^{4}$ \\
\midrule
\multicolumn{10}{c}{\textbf{Qwen3-32B}} \\
\midrule
\multicolumn{10}{l}{\textit{No-op class}} \\
ATU                    & 0.03\ci{0.01} & 0.04\ci{0.01} & 0.24\ci{0.04} & 19.6\ci{2.0} & 23.0\ci{10.4} & 26.4\ci{4.4} & 1.0 & 1.0 & 1.0 \\
Obliviate              & 0.04\ci{0.01} & 0.03\ci{0.01} & 0.32\ci{0.05} & 20.8\ci{2.5} & 21.3\ci{1.4} & 26.5\ci{3.2} & 1.0 & 1.0 & 1.3 \\
RSV                    & 0.04\ci{0.03} & 0.04\ci{0.02} & 0.24\ci{0.05} & 20.0\ci{2.6} & 20.7\ci{1.6} & 26.5\ci{4.6} & 1.0 & 1.0 & 1.0 \\
RMU                    & 0.06\ci{0.02} & 0.03\ci{0.01} & 0.24\ci{0.05} & 20.5\ci{2.0} & 20.7\ci{1.2} & 26.4\ci{4.5} & 1.2 & 1.0 & 1.0 \\
Adaptive-RMU           & 0.83\ci{0.04} & 0.36\ci{0.07} & 0.24\ci{0.04} & 35.1\ci{4.6} & 29.2\ci{1.7} & 25.8\ci{1.8} & $10^{3}$ & $10^{1}$ & 1.0 \\
\multicolumn{10}{l}{\textit{Collateral-dominant class}} \\
NPO                    & 9.96\ci{1.42} & 12.58\ci{2.48} & 0.24\ci{0.05} & 67.1\ci{8.4} & 69.6\ci{3.5} & 26.5\ci{4.5} & $10^{40}$ & $10^{30}$ & 1.2 \\
SimNPO                 & 8.83\ci{1.56} & 12.23\ci{2.87} & 0.24\ci{0.05} & 63.9\ci{5.9} & 69.8\ci{2.3} & 26.3\ci{4.1} & $10^{39}$ & $10^{30}$ & 1.2 \\
DPO                    & 8.86\ci{1.41} & 15.21\ci{2.79} & 0.24\ci{0.05} & 59.5\ci{4.4} & 68.8\ci{2.3} & 26.4\ci{4.4} & $10^{44}$ & $10^{33}$ & 1.0 \\
LoKU (no FILA)         & 1.88\ci{0.17} & 2.28\ci{0.82} & 0.23\ci{0.13} & 41.8\ci{3.5} & 48.6\ci{5.1} & 32.3\ci{11.9} & $10^{12}$ & $10^{12}$ & 1.0 \\
GA                     & 1.98\ci{0.81} & 6.01\ci{9.87} & 3.57\ci{2.12} & 41.8\ci{6.7} & 51.8\ci{10.3} & 27.9\ci{4.9} & $10^{73}$ & $10^{74}$ & $10^{1}$ \\
GD                     & 2.45\ci{1.46} & 4.23\ci{6.34} & 2.54\ci{3.13} & 48.3\ci{16.1} & 48.7\ci{8.9} & 31.4\ci{12.2} & $10^{69}$ & $10^{64}$ & $10^{1}$ \\
\multicolumn{10}{l}{\textit{Globally destructive class}} \\
SPUL                   & 9.72\ci{0.68} & 9.95\ci{4.12} & 9.54\ci{2.16} & 66.0\ci{3.9} & 64.1\ci{0.7} & 49.3\ci{19.7} & $10^{43}$ & $10^{39}$ & $10^{28}$ \\
\midrule
\multicolumn{10}{c}{\textbf{Zephyr-7B-$\beta$}} \\
\midrule
\multicolumn{10}{l}{\textit{No-op class}} \\
ATU                    & 0.04\ci{0.01} & 0.04\ci{0.00} & 0.28\ci{0.31} & 48.8\ci{1.8} & 46.4\ci{2.9} & 50.4\ci{2.9} & 1.0 & 1.0 & 1.0 \\
\multicolumn{10}{l}{\textit{Partially-localized class}} \\
RSV                    & 17.58\ci{0.50} & 16.98\ci{2.96} & 3.00\ci{0.04} & 49.9\ci{1.7} & 47.0\ci{2.6} & 50.7\ci{2.3} & $10^{4}$ & $10^{3}$ & 1.0 \\
Adaptive-RMU           & 1.45\ci{0.07} & 0.06\ci{0.02} & 0.20\ci{0.16} & 48.9\ci{1.3} & 46.3\ci{2.2} & 50.4\ci{2.8} & 1.2 & 1.0 & 1.0 \\
\multicolumn{10}{l}{\textit{Collateral-dominant class}} \\
Obliviate              & 6.93\ci{0.90} & 8.90\ci{0.73} & 1.50\ci{0.63} & 58.6\ci{0.7} & 58.0\ci{2.1} & 56.2\ci{0.5} & $10^{4}$ & $10^{3}$ & 1.9 \\
RMU                    & 22.10\ci{0.65} & 22.55\ci{0.78} & 2.53\ci{2.20} & 55.5\ci{2.0} & 50.8\ci{4.2} & 50.7\ci{2.5} & $10^{5}$ & $10^{4}$ & 1.0 \\
NPO                    & 25.43\ci{1.45} & 34.98\ci{4.59} & 16.92\ci{9.45} & 103.9\ci{3.7} & 103.0\ci{8.0} & 66.2\ci{27.1} & $10^{58}$ & $10^{60}$ & $10^{2}$ \\
SimNPO                 & 30.05\ci{5.57} & 37.27\ci{5.00} & 14.31\ci{9.61} & 103.8\ci{10.6} & 93.6\ci{9.3} & 65.6\ci{31.7} & $10^{58}$ & $10^{59}$ & $10^{3}$ \\
DPO                    & 8.54\ci{4.22} & 19.16\ci{4.82} & 10.62\ci{6.35} & 77.4\ci{6.5} & 80.2\ci{12.9} & 52.3\ci{6.3} & $10^{28}$ & $10^{28}$ & 5.2 \\
\multicolumn{10}{l}{\textit{Globally destructive class}} \\
LoKU+FILA              & 14.54\ci{0.54} & 14.88\ci{4.19} & 12.84\ci{1.13} & 67.4\ci{1.0} & 65.0\ci{2.2} & 69.6\ci{8.3} & $10^{4}$ & $10^{4}$ & $10^{3}$ \\
GA                     & 5.07\ci{3.33} & 11.09\ci{10.65} & 14.74\ci{28.37} & 153.1\ci{4.3} & 150.2\ci{5.2} & 134.4\ci{3.4} & $10^{112}$ & $10^{112}$ & $10^{1}$ \\
GD                     & 3.96\ci{1.43} & 4.68\ci{1.91} & 23.84\ci{20.35} & 151.5\ci{2.1} & 149.1\ci{4.2} & 126.0\ci{6.7} & $10^{112}$ & $10^{112}$ & $10^{2}$ \\
SPUL                   & 15.99\ci{0.74} & 19.43\ci{1.77} & 38.13\ci{3.82} & 121.8\ci{2.0} & 119.7\ci{2.3} & 115.5\ci{2.8} & $10^{27}$ & $10^{34}$ & $10^{30}$ \\
\bottomrule
\end{tabular}
\end{table}

\begin{table}[t]
\centering
\small
\caption{TRIAGE class assigned to each method on each model. \textsc{n} = no-op, \textsc{p} = partially localized, \textsc{c} = collateral dominant, \textsc{g} = globally destructive; \textendash{} = not run. Classes are assigned from the CoFi aggregates of Tables~\ref{tab:wmdp-cofi-chess-llama8b}--\ref{tab:wmdp-cofi-chess-zephyr}; CHess and perplexity take no part in the assignment.}
\label{tab:triage-class-summary}
\begin{tabular}{l cccc}
\toprule
\textbf{Method} & \textbf{Llama-3.1-8B} & \textbf{Llama-3.2-3B} & \textbf{Qwen3-32B} & \textbf{Zephyr-7B-$\beta$} \\
\midrule
ATU                    & \textsc{n} & \textsc{n} & \textsc{n} & \textsc{n} \\
Obliviate              & \textsc{n} & \textsc{n} & \textsc{n} & \textsc{c} \\
RSV                    & \textsc{n} & \textsc{n} & \textsc{n} & \textsc{p} \\
RSV ($\nu$)            & \textsc{n} & \textsc{n} & \textsc{n} & \textsc{p} \\
RMU                    & \textsc{p} & \textsc{p} & \textsc{n} & \textsc{c} \\
RMU ($\nu$)            & \textsc{p} & \textsc{p} & \textsc{n} & \textsc{c} \\
Adaptive-RMU           & \textsc{c} & \textsc{c} & \textsc{n} & \textsc{p} \\
Adaptive-RMU ($\nu$)   & \textsc{c} & \textsc{c} & \textsc{n} & \textsc{p} \\
NPO                    & \textsc{c} & \textsc{c} & \textsc{c} & \textsc{c} \\
NPO ($\nu$)            & \textsc{c} & \textsc{c} & \textsc{c} & \textsc{c} \\
SimNPO                 & \textsc{c} & \textsc{c} & \textsc{c} & \textsc{c} \\
SimNPO ($\nu$)         & \textsc{c} & \textsc{c} & \textsc{c} & \textsc{c} \\
DPO                    & \textsc{c} & \textsc{c} & \textsc{c} & \textsc{c} \\
DPO ($\nu$)            & \textsc{c} & \textsc{c} & \textsc{c} & \textsc{c} \\
LoKU+FILA              & \textsc{g} & \textsc{g} & \textendash & \textsc{g} \\
LoKU (no FILA)         & \textendash & \textendash & \textsc{c} & \textendash \\
GA                     & \textsc{g} & \textsc{g} & \textsc{c} & \textsc{g} \\
GA ($\nu$)             & \textsc{g} & \textsc{g} & \textsc{c} & \textsc{g} \\
GD                     & \textsc{g} & \textsc{g} & \textsc{c} & \textsc{g} \\
GD ($\nu$)             & \textsc{g} & \textsc{g} & \textsc{c} & \textsc{g} \\
SPUL                   & \textsc{g} & \textsc{g} & \textsc{g} & \textsc{g} \\
\bottomrule
\end{tabular}
\end{table}

\subsection{TOFU}
\label{app:tofu-results}

Tables~\ref{tab:tofu-cofi-chess-llama8b}--\ref{tab:tofu-cofi-chess-zephyr} give the
per-corpus CoFi and CHess relative shifts for Llama-3.1-8B, Llama-3.2-3B and
Zephyr-7B-$\beta$; Table~\ref{tab:tofu-ppl-1} gives the perplexity ratios and
Table~\ref{tab:tofu-class-summary} the class assignments. The corresponding heatmaps
are in Section~\ref{app:heatmaps}.

\paragraph{Most methods do not register.}
TOFU is the benchmark on which CoFi is quietest. Of the 20 method variants, 15 fall in
the no-op class on Llama-3.1-8B, 19 on Llama-3.2-3B and 15 on Zephyr-7B-$\beta$, with
corpus shift below $0.8\%$ and perplexity ratio below $1.1$. The exceptions
are the ascent-based methods and LoKU. On Llama-3.1-8B, GA and GD are partially
localized, moving the forget split by $11.64\%$ and $9.60\%$ against $11.50\%$ and
$8.99\%$ on retain-90, while leaving WikiText untouched ($0.44\%$ and $0.33\%$). On
Llama-3.2-3B the same two methods collapse to the no-op class, shifting the forget split
by only $0.36\%$; the 3B model apparently absorbs the same objective without moving its
answer rankings at all. LoKU+FILA is the one method that is globally destructive on this
benchmark, and it is so on all three models. On Zephyr-7B-$\beta$ GA and GD join the
active set with forget-split shifts above $16\%$, but their damage stays inside the
TOFU corpora rather than reaching general text.

\paragraph{CoFi and CHess separate cleanly here.}
TOFU is the clearest illustration of what the two probes measure differently. GA on
Llama-3.1-8B raises CHess on the forget split to $57.07\%$ against $32.18\%$ for the
untouched ATU baseline, and CoFi agrees that something large happened
($11.64\%$ against $0.30\%$). On Llama-3.2-3B the same method gives CHess $28.27\%$
against ATU's $27.14\%$ and CoFi $0.36\%$ against $0.04\%$: both probes report a small
effect, and they report it consistently. The disagreements to watch are in the opposite
direction, where a method reshapes the distribution without displacing the correct
answer; SPUL is the example, with forget-split perplexity ratios of $4.25$ and $4.29$ on
the two Llama models and a WikiText ratio of $19.3$ and $18.1$, against CoFi shifts that
never leave the floor.

\paragraph{Marginal assignments.}
One call on Zephyr-7B-$\beta$ is close. GA is partially localized and GD collateral
dominant, but the comparison that separates them is inside the confidence intervals for
GD ($\mathcal{C}_F = 16.66 \pm 0.49$ against $\mathcal{C}_A = 16.85 \pm 0.58$), so the
two should be read as occupying the same position with the sign of the asymmetry
undetermined. Both also sit just below the globally destructive threshold, with
globality ratios of $0.69$ and $0.72$.

\paragraph{Effect of noise augmentation.}
Base and $\nu$ variants agree on class in all 24 method--model pairs, and the agreement
in magnitude is the tightest of any benchmark: the median across pairs of the largest
per-corpus CoFi discrepancy is $0.02$ percentage points and the largest single
discrepancy anywhere is $0.97$ points, for GD on Zephyr-7B-$\beta$. On TOFU, RNA is
effectively a no-op on top of whatever the base method does.

\subsection{MUSE-Books}
\label{app:muse-books-results}

Tables~\ref{tab:books-cofi-chess-llama8b}--\ref{tab:books-cofi-chess-zephyr} give the
per-corpus CoFi and CHess shifts for all four models,
Tables~\ref{tab:books-ppl-1} and~\ref{tab:books-ppl-2} the perplexity ratios, and
Table~\ref{tab:books-class-summary} the class assignments. The corresponding heatmaps
are in Section~\ref{app:heatmaps}.

\paragraph{Confidence intervals on MUSE-Books.}
For every benchmark we estimate uncertainty by evaluating each metric on three
independent subsets per corpus. Each subset is a seeded random sample of 200 documents,
drawn once and shared across all models and unlearning methods, and 95\% confidence
intervals are computed across the three. This works when a corpus contains many more
documents than the subset size, as in WMDP, MUSE-News and TOFU. MUSE-Books is different:
its forget, retain-1 and retain-2 splits contain only 4, 12 and 13 documents, each an
entire book or a long excerpt. With so few documents every subset contains the whole
split, so the three subsets will be identical. We therefore report point
estimates for the three MUSE-Books splits. The
WikiText column is drawn from a corpus large enough to subsample, so it is the only
column in these tables that carries a genuine interval, and it is shown with one.

\paragraph{Window selection on the forget split.}
Because the books are much longer than the model context, documents are split into
consecutive 1024-token windows before CoFi and CHess are computed, and the first 200
windows are used. For the MUSE-Books forget split these 200 windows cover about 205k of
973k tokens and all come from the first book in the split, so the CoFi and CHess
estimates for that split reflect a contiguous portion of the forget corpus rather than a
sample spread across all four books. Perplexity is computed with a sliding window over
the full text and does cover the whole corpus, which is worth keeping in mind when the
two disagree.

\paragraph{The partially localized class dominates.}
MUSE-Books produces the opposite structural picture to WMDP, and the reason is the
construction of the splits. The retain corpora are different books, not different
passages of the same subject matter, so a method that damages the forget book has
nowhere to spill: retain-1 and retain-2 sit at the measurement floor around $0.03\%$
while forget-split shifts reach $5$--$29\%$. Ten of the 20 variants on Llama-3.1-8B are
partially localized, as are ten on Llama-3.2-3B and twelve on Zephyr-7B-$\beta$, and the
collateral dominant class is nearly empty across the benchmark. NPO, SimNPO and DPO are
the clearest cases, reaching $10$--$21\%$ on the forget split with retain shifts of
$0.03$--$0.07\%$ and forget-split perplexity ratios between $10^{26}$ and $10^{39}$.
The methods that escape this pattern are the ascent-based ones, LoKU and SPUL, which
move WikiText by $3$--$43\%$; SPUL is globally destructive on every model. Obliviate is
a no-op on three of the four models, moving no corpus by more than $1.5\%$. The only exception is the Zephyr-7B-$\beta$ model, on which this algorithm is classified as partially localized.

\paragraph{Class migration across models.}
RSV is a no-op on the two Llama models and on Qwen3-32B but partially localized on
Zephyr-7B-$\beta$, where it shifts the forget split by $23.55\%$. RMU follows the same
route, from a no-op on Qwen3-32B to a forget-split shift of $28.56\%$ on
Zephyr-7B-$\beta$. LoKU+FILA is globally destructive on Llama-3.1-8B and Llama-3.2-3B
but collateral dominant on Zephyr-7B-$\beta$, while the FILA-free configuration run on
Qwen3-32B is a no-op. Qwen3-32B is the most resistant model here, with nine of 20
variants in the no-op class.

\paragraph{Effect of noise augmentation.}
Base and $\nu$ variants agree on class in 31 of the 32 pairs, with a median largest
per-corpus CoFi discrepancy of $1.66$ percentage points. The exception is Adaptive-RMU
on Zephyr-7B-$\beta$, where the base run leaves the forget split at $0.30\%$ and the
$\nu$ run moves it to $26.34\%$, with the forget-split perplexity ratio rising from
$1.03$ to $6.0 \times 10^{4}$ and the class changing from no-op to partially localized.
No other model shows anything comparable for this method, and the size of the gap
suggests a run-level rather than a metric-level effect; we report it as measured and do
not draw a conclusion from it.

\subsection{MUSE-News}
\label{app:muse-news-results}

Tables~\ref{tab:news-cofi-chess-llama8b}--\ref{tab:news-cofi-chess-zephyr} give the
per-corpus CoFi and CHess shifts, Tables~\ref{tab:news-ppl-1}
and~\ref{tab:news-ppl-2} the perplexity ratios, and
Table~\ref{tab:news-class-summary} the class assignments. The corresponding heatmaps
are in Section~\ref{app:heatmaps}.

\paragraph{Only two classes are populated.}
MUSE-News produces the sparsest and most uniform taxonomy of the three benchmarks: every
method variant is either a no-op or globally destructive, neither the partially
localized nor the collateral dominant class appears on any model, and the split is
identical across all three models at 15 no-ops and 5 globally destructive variants. The
active set is the same everywhere --- LoKU+FILA, GA, GD and the $\nu$ variants of the
latter two --- while the remaining 15 variants leave every corpus below $1.5\%$, with
forget-split shifts of $0.02$--$0.29\%$. This uniformity is a consequence of how the
methods fail rather than of how they succeed: nothing on this benchmark manages to move
the news forget split without moving WikiText at least as much. GA on Llama-3.1-8B is
the extreme case, with $\mathcal{C}_F = 1.34\%$ against $\mathcal{C}_G = 22.87\%$ and
perplexity ratios above $10^{53}$ on every corpus, a model that has been destroyed
rather than edited.

\paragraph{The clearest CoFi--perplexity disagreement in our results.}
SPUL on MUSE-News is worth isolating. Its CoFi shifts never leave the floor
($0.03\%$ on the forget split for both Llama models) and its CHess shifts are
indistinguishable from the untouched baselines, yet its perplexity ratios are $11.0$ and
$24.1$ on the forget split and $21.3$ and $23.1$ on WikiText. The method has measurably
degraded the model's fluency without displacing a single correct-answer ranking. This is
precisely the case that motivates reporting perplexity as a check rather than as a
classification input: on the criterion the taxonomy is built on, the run did nothing,
and the perplexity column is what tells us the model was nonetheless touched.

\paragraph{Effect of noise augmentation.}
Base and $\nu$ variants agree on class in all 24 pairs. The median largest per-corpus
CoFi discrepancy is $0.03$ percentage points, and the largest anywhere is $3.44$ points,
for GA on Llama-3.1-8B, where the WikiText shift moves from $22.87\%$ to $19.43\%$ ---
a difference well inside that entry's confidence interval of $\pm 6.02$.

\begin{table}[t]
\centering
\footnotesize
\setlength{\tabcolsep}{3pt}
\caption{TOFU per-corpus CoFi and CHess relative shifts (\%) on \textbf{Llama-3.1-8B}, mean $\pm$ 95\% CI over three random subsets of 200 samples. $\uparrow$ on the forget split = stronger unlearning; $\downarrow$ on the retain splits and on \textit{wiki} = less collateral damage. Methods are grouped by the TRIAGE class they fall into \emph{for this model and this benchmark}; the assignment uses the CoFi aggregates alone, with CHess reported alongside as a complementary view of how far the answer distribution has moved. ($\nu$) marks the randomly-perturbed variant.}
\label{tab:tofu-cofi-chess-llama8b}
\begin{tabular}{l rrr @{\hspace{6pt}} rrr}
\toprule
& \multicolumn{3}{c}{\textbf{CoFi (\%)}} & \multicolumn{3}{c}{\textbf{CHess (\%)}} \\
\cmidrule(lr){2-4}\cmidrule(lr){5-7}
\textbf{Method} & \textit{forget10} & \textit{retain90} & \textit{wiki} & \textit{forget10} & \textit{retain90} & \textit{wiki} \\
\midrule
\multicolumn{7}{l}{\textit{No-op class}} \\
ATU                    & 0.30\ci{0.02} & 0.33\ci{0.02} & 0.43\ci{0.23} & 32.2\ci{0.5} & 31.7\ci{0.8} & 28.1\ci{1.0} \\
Obliviate              & 0.68\ci{0.06} & 0.74\ci{0.04} & 0.76\ci{0.33} & 32.6\ci{0.3} & 32.1\ci{0.3} & 28.2\ci{1.1} \\
RSV                    & 0.24\ci{0.01} & 0.26\ci{0.00} & 0.32\ci{0.22} & 32.2\ci{0.6} & 31.7\ci{0.9} & 28.1\ci{1.3} \\
RSV ($\nu$)            & 0.24\ci{0.03} & 0.27\ci{0.01} & 0.31\ci{0.15} & 32.2\ci{0.5} & 31.6\ci{0.7} & 28.1\ci{1.3} \\
RMU                    & 0.23\ci{0.04} & 0.26\ci{0.02} & 0.32\ci{0.19} & 32.2\ci{0.4} & 31.7\ci{0.7} & 28.1\ci{1.0} \\
RMU ($\nu$)            & 0.23\ci{0.01} & 0.25\ci{0.01} & 0.30\ci{0.18} & 32.2\ci{0.5} & 31.7\ci{0.7} & 28.0\ci{1.0} \\
Adaptive-RMU           & 0.25\ci{0.02} & 0.27\ci{0.01} & 0.35\ci{0.18} & 32.2\ci{0.5} & 31.8\ci{0.7} & 28.2\ci{1.2} \\
Adaptive-RMU ($\nu$)   & 0.25\ci{0.03} & 0.28\ci{0.02} & 0.34\ci{0.17} & 32.2\ci{0.6} & 31.7\ci{0.9} & 28.1\ci{0.9} \\
NPO                    & 0.33\ci{0.04} & 0.31\ci{0.03} & 0.32\ci{0.16} & 32.2\ci{0.6} & 31.7\ci{0.8} & 28.3\ci{1.3} \\
NPO ($\nu$)            & 0.30\ci{0.04} & 0.24\ci{0.00} & 0.27\ci{0.18} & 32.2\ci{0.5} & 31.6\ci{0.8} & 28.0\ci{1.0} \\
SimNPO                 & 0.27\ci{0.03} & 0.24\ci{0.01} & 0.29\ci{0.19} & 32.2\ci{0.7} & 31.6\ci{0.9} & 28.0\ci{1.1} \\
SimNPO ($\nu$)         & 0.33\ci{0.08} & 0.26\ci{0.01} & 0.29\ci{0.17} & 32.2\ci{0.6} & 31.7\ci{0.7} & 28.1\ci{1.5} \\
DPO                    & 0.36\ci{0.05} & 0.34\ci{0.11} & 0.39\ci{0.21} & 32.3\ci{0.5} & 31.7\ci{0.7} & 28.2\ci{1.2} \\
DPO ($\nu$)            & 0.35\ci{0.03} & 0.30\ci{0.01} & 0.38\ci{0.19} & 32.3\ci{0.5} & 31.6\ci{0.9} & 28.1\ci{1.1} \\
SPUL                   & 0.34\ci{0.02} & 0.33\ci{0.02} & 0.49\ci{0.73} & 32.4\ci{0.7} & 31.8\ci{0.9} & 28.1\ci{0.9} \\
\midrule
\multicolumn{7}{l}{\textit{Partially-localized class}} \\
GA                     & 11.64\ci{0.37} & 11.50\ci{0.12} & 0.44\ci{0.45} & 57.1\ci{0.4} & 56.7\ci{0.9} & 28.6\ci{1.4} \\
GA ($\nu$)             & 11.17\ci{0.34} & 10.87\ci{0.04} & 0.40\ci{0.32} & 56.1\ci{0.8} & 56.1\ci{1.3} & 28.5\ci{1.6} \\
GD                     & 9.60\ci{0.26} & 8.99\ci{0.26} & 0.33\ci{0.22} & 53.2\ci{0.8} & 52.5\ci{0.9} & 28.8\ci{1.6} \\
GD ($\nu$)             & 9.79\ci{0.36} & 9.33\ci{0.29} & 0.33\ci{0.24} & 53.2\ci{0.7} & 52.8\ci{0.9} & 28.8\ci{2.7} \\
\midrule
\multicolumn{7}{l}{\textit{Globally destructive class}} \\
LoKU+FILA              & 2.70\ci{0.02} & 3.27\ci{1.71} & 3.43\ci{1.04} & 36.1\ci{0.5} & 35.4\ci{1.2} & 36.8\ci{5.1} \\
\bottomrule
\end{tabular}
\end{table}

\begin{table}[t]
\centering
\footnotesize
\setlength{\tabcolsep}{3pt}
\caption{TOFU per-corpus CoFi and CHess relative shifts (\%) on \textbf{Llama-3.2-3B}, mean $\pm$ 95\% CI over three random subsets of 200 samples. $\uparrow$ on the forget split = stronger unlearning; $\downarrow$ on the retain splits and on \textit{wiki} = less collateral damage. Methods are grouped by the TRIAGE class they fall into \emph{for this model and this benchmark}; the assignment uses the CoFi aggregates alone, with CHess reported alongside as a complementary view of how far the answer distribution has moved. ($\nu$) marks the randomly-perturbed variant.}
\label{tab:tofu-cofi-chess-llama3b}
\begin{tabular}{l rrr @{\hspace{6pt}} rrr}
\toprule
& \multicolumn{3}{c}{\textbf{CoFi (\%)}} & \multicolumn{3}{c}{\textbf{CHess (\%)}} \\
\cmidrule(lr){2-4}\cmidrule(lr){5-7}
\textbf{Method} & \textit{forget10} & \textit{retain90} & \textit{wiki} & \textit{forget10} & \textit{retain90} & \textit{wiki} \\
\midrule
\multicolumn{7}{l}{\textit{No-op class}} \\
ATU                    & 0.04\ci{0.00} & 0.07\ci{0.04} & 0.10\ci{0.18} & 27.1\ci{0.6} & 26.8\ci{0.9} & 26.4\ci{1.3} \\
Obliviate              & 0.15\ci{0.00} & 0.15\ci{0.01} & 0.06\ci{0.10} & 27.3\ci{0.4} & 26.9\ci{0.8} & 26.3\ci{1.2} \\
RSV                    & 0.04\ci{0.00} & 0.07\ci{0.09} & 0.06\ci{0.03} & 27.1\ci{0.6} & 26.8\ci{0.8} & 26.3\ci{1.2} \\
RSV ($\nu$)            & 0.04\ci{0.00} & 0.05\ci{0.03} & 0.08\ci{0.11} & 27.1\ci{0.6} & 26.8\ci{0.8} & 26.3\ci{1.1} \\
RMU                    & 0.04\ci{0.00} & 0.05\ci{0.02} & 0.06\ci{0.04} & 27.1\ci{0.5} & 26.8\ci{0.8} & 26.4\ci{1.2} \\
RMU ($\nu$)            & 0.04\ci{0.00} & 0.05\ci{0.01} & 0.08\ci{0.09} & 27.1\ci{0.5} & 26.8\ci{0.8} & 26.3\ci{1.1} \\
Adaptive-RMU           & 0.04\ci{0.02} & 0.04\ci{0.01} & 0.07\ci{0.06} & 27.1\ci{0.7} & 26.8\ci{0.8} & 26.4\ci{1.3} \\
Adaptive-RMU ($\nu$)   & 0.04\ci{0.02} & 0.04\ci{0.01} & 0.08\ci{0.10} & 27.1\ci{0.6} & 26.8\ci{0.8} & 26.3\ci{1.2} \\
NPO                    & 0.07\ci{0.02} & 0.06\ci{0.04} & 0.08\ci{0.10} & 27.1\ci{0.6} & 26.8\ci{0.8} & 26.4\ci{1.2} \\
NPO ($\nu$)            & 0.08\ci{0.02} & 0.05\ci{0.03} & 0.07\ci{0.06} & 27.2\ci{0.6} & 26.8\ci{0.8} & 26.4\ci{1.2} \\
SimNPO                 & 0.07\ci{0.02} & 0.06\ci{0.05} & 0.08\ci{0.09} & 27.1\ci{0.6} & 26.8\ci{0.7} & 26.4\ci{1.1} \\
SimNPO ($\nu$)         & 0.07\ci{0.02} & 0.05\ci{0.03} & 0.08\ci{0.09} & 27.2\ci{0.5} & 26.8\ci{0.8} & 26.3\ci{1.2} \\
DPO                    & 0.11\ci{0.02} & 0.07\ci{0.04} & 0.15\ci{0.33} & 27.2\ci{0.5} & 26.8\ci{1.0} & 26.3\ci{1.2} \\
DPO ($\nu$)            & 0.11\ci{0.02} & 0.07\ci{0.05} & 0.13\ci{0.23} & 27.2\ci{0.6} & 26.8\ci{0.8} & 26.3\ci{1.2} \\
GA                     & 0.36\ci{0.01} & 0.29\ci{0.05} & 0.11\ci{0.19} & 28.3\ci{0.4} & 27.8\ci{0.4} & 26.3\ci{1.2} \\
GA ($\nu$)             & 0.36\ci{0.02} & 0.29\ci{0.05} & 0.10\ci{0.18} & 28.3\ci{0.4} & 27.8\ci{0.7} & 26.3\ci{1.2} \\
GD                     & 0.36\ci{0.04} & 0.28\ci{0.05} & 0.11\ci{0.18} & 28.3\ci{0.3} & 27.8\ci{0.5} & 26.5\ci{1.2} \\
GD ($\nu$)             & 0.35\ci{0.02} & 0.29\ci{0.06} & 0.12\ci{0.19} & 28.3\ci{0.5} & 27.8\ci{0.7} & 26.4\ci{1.3} \\
SPUL                   & 0.26\ci{0.01} & 0.22\ci{0.02} & 0.41\ci{0.36} & 27.4\ci{0.5} & 26.9\ci{0.8} & 26.5\ci{1.2} \\
\midrule
\multicolumn{7}{l}{\textit{Globally destructive class}} \\
LoKU+FILA              & 2.74\ci{0.07} & 2.71\ci{0.11} & 3.13\ci{0.16} & 37.1\ci{1.7} & 36.0\ci{0.4} & 36.5\ci{0.4} \\
\bottomrule
\end{tabular}
\end{table}

\begin{table}[t]
\centering
\footnotesize
\setlength{\tabcolsep}{3pt}
\caption{TOFU per-corpus CoFi and CHess relative shifts (\%) on \textbf{Zephyr-7B-$\beta$}, mean $\pm$ 95\% CI over three random subsets of 200 samples. $\uparrow$ on the forget split = stronger unlearning; $\downarrow$ on the retain splits and on \textit{wiki} = less collateral damage. Methods are grouped by the TRIAGE class they fall into \emph{for this model and this benchmark}; the assignment uses the CoFi aggregates alone, with CHess reported alongside as a complementary view of how far the answer distribution has moved. ($\nu$) marks the randomly-perturbed variant.}
\label{tab:tofu-cofi-chess-zephyr}
\begin{tabular}{l rrr @{\hspace{6pt}} rrr}
\toprule
& \multicolumn{3}{c}{\textbf{CoFi (\%)}} & \multicolumn{3}{c}{\textbf{CHess (\%)}} \\
\cmidrule(lr){2-4}\cmidrule(lr){5-7}
\textbf{Method} & \textit{forget10} & \textit{retain90} & \textit{wiki} & \textit{forget10} & \textit{retain90} & \textit{wiki} \\
\midrule
\multicolumn{7}{l}{\textit{No-op class}} \\
ATU                    & 0.20\ci{0.01} & 0.21\ci{0.08} & 0.19\ci{0.06} & 60.7\ci{1.7} & 60.6\ci{2.0} & 55.8\ci{2.6} \\
Obliviate              & 0.42\ci{0.01} & 0.42\ci{0.03} & 0.34\ci{0.07} & 60.6\ci{1.6} & 60.4\ci{2.4} & 55.5\ci{2.1} \\
RSV                    & 0.04\ci{0.02} & 0.04\ci{0.01} & 0.07\ci{0.02} & 60.6\ci{1.6} & 60.4\ci{2.1} & 55.5\ci{2.0} \\
RSV ($\nu$)            & 0.04\ci{0.00} & 0.04\ci{0.01} & 0.09\ci{0.07} & 60.6\ci{1.6} & 60.4\ci{2.2} & 55.5\ci{2.1} \\
RMU                    & 0.12\ci{0.04} & 0.04\ci{0.01} & 0.07\ci{0.02} & 60.7\ci{1.8} & 60.5\ci{2.2} & 55.5\ci{1.9} \\
RMU ($\nu$)            & 0.13\ci{0.02} & 0.04\ci{0.01} & 0.07\ci{0.02} & 60.6\ci{1.6} & 60.4\ci{2.2} & 55.5\ci{2.2} \\
Adaptive-RMU           & 0.18\ci{0.03} & 0.04\ci{0.01} & 0.07\ci{0.01} & 60.6\ci{1.7} & 60.4\ci{2.2} & 55.5\ci{2.2} \\
Adaptive-RMU ($\nu$)   & 0.18\ci{0.02} & 0.04\ci{0.01} & 0.08\ci{0.01} & 60.7\ci{1.7} & 60.5\ci{2.6} & 55.6\ci{2.3} \\
NPO                    & 0.53\ci{0.07} & 0.17\ci{0.19} & 0.11\ci{0.03} & 60.6\ci{1.6} & 60.4\ci{2.1} & 55.5\ci{2.0} \\
NPO ($\nu$)            & 0.52\ci{0.07} & 0.16\ci{0.13} & 0.12\ci{0.04} & 60.6\ci{1.6} & 60.4\ci{2.2} & 55.5\ci{2.0} \\
SimNPO                 & 0.51\ci{0.06} & 0.17\ci{0.20} & 0.11\ci{0.04} & 60.7\ci{1.6} & 60.4\ci{2.2} & 55.8\ci{2.9} \\
SimNPO ($\nu$)         & 0.52\ci{0.07} & 0.18\ci{0.16} & 0.12\ci{0.05} & 60.6\ci{1.6} & 60.4\ci{2.1} & 55.5\ci{2.2} \\
DPO                    & 0.53\ci{0.10} & 0.20\ci{0.06} & 0.20\ci{0.08} & 60.7\ci{1.6} & 60.5\ci{2.3} & 55.5\ci{1.9} \\
DPO ($\nu$)            & 0.52\ci{0.09} & 0.19\ci{0.08} & 0.20\ci{0.12} & 60.7\ci{1.8} & 60.5\ci{2.4} & 55.7\ci{2.4} \\
SPUL                   & 0.70\ci{0.22} & 0.55\ci{0.07} & 1.34\ci{0.57} & 61.1\ci{1.6} & 60.9\ci{2.0} & 55.9\ci{2.8} \\
\midrule
\multicolumn{7}{l}{\textit{Partially-localized class}} \\
GA                     & 22.46\ci{0.76} & 21.47\ci{0.43} & 15.42\ci{2.30} & 91.0\ci{2.9} & 89.7\ci{2.9} & 81.7\ci{1.9} \\
GA ($\nu$)             & 22.69\ci{0.76} & 21.78\ci{0.38} & 15.97\ci{2.44} & 91.5\ci{1.4} & 91.9\ci{2.6} & 83.1\ci{5.6} \\
\midrule
\multicolumn{7}{l}{\textit{Collateral-dominant class}} \\
GD                     & 16.66\ci{0.49} & 16.85\ci{0.58} & 12.17\ci{2.07} & 77.2\ci{1.3} & 76.5\ci{2.5} & 68.3\ci{2.6} \\
GD ($\nu$)             & 15.69\ci{0.40} & 16.01\ci{0.41} & 11.32\ci{1.95} & 75.1\ci{2.4} & 74.2\ci{1.3} & 65.2\ci{3.0} \\
\midrule
\multicolumn{7}{l}{\textit{Globally destructive class}} \\
LoKU+FILA              & 12.88\ci{7.27} & 11.25\ci{11.95} & 12.36\ci{6.89} & 65.0\ci{1.4} & 64.8\ci{3.5} & 59.3\ci{2.3} \\
\bottomrule
\end{tabular}
\end{table}

\begin{table}[t]
\centering
\small
\setlength{\tabcolsep}{4pt}
\caption{TOFU perplexity ratios (unlearned/original) for Llama-3.1-8B, Llama-3.2-3B, Zephyr-7B-$\beta$, reported as $\log_{10}$ of the ratio. $0.00$ means fluency is unchanged; $\uparrow$ on the forget split is intended, while any large value on the retain splits or on \textit{wiki} is pure damage. Perplexity is a fluency sanity check and takes no part in the TRIAGE class assignment, which is made from CoFi. Rows follow a fixed method order so the models can be read off against each other; class membership is given in Table~\ref{tab:tofu-class-summary}.}
\label{tab:tofu-ppl-1}
\begin{tabular}{l rrr @{\hspace{6pt}} rrr @{\hspace{6pt}} rrr}
\toprule
& \multicolumn{3}{c}{\textbf{Llama-3.1-8B}} & \multicolumn{3}{c}{\textbf{Llama-3.2-3B}} & \multicolumn{3}{c}{\textbf{Zephyr-7B-$\beta$}} \\
\cmidrule(lr){2-4}\cmidrule(lr){5-7}\cmidrule(lr){8-10}
\textbf{Method} & \textit{forget10} & \textit{retain90} & \textit{wiki} & \textit{forget10} & \textit{retain90} & \textit{wiki} & \textit{forget10} & \textit{retain90} & \textit{wiki} \\
\midrule
ATU                    & 0.00 & 0.00 & 0.00 & 0.00 & 0.00 & 0.00 & 0.00 & 0.00 & -0.01 \\
Obliviate              & 0.02 & 0.03 & 0.00 & 0.01 & 0.02 & 0.00 & 0.02 & 0.03 & 0.03 \\
RSV                    & 0.00 & 0.00 & 0.00 & 0.00 & 0.00 & 0.00 & 0.00 & 0.00 & 0.00 \\
RSV ($\nu$)            & 0.00 & 0.00 & 0.00 & 0.00 & 0.00 & 0.00 & 0.00 & 0.00 & 0.00 \\
RMU                    & 0.00 & 0.00 & 0.00 & 0.00 & 0.00 & 0.00 & 0.00 & 0.00 & 0.00 \\
RMU ($\nu$)            & 0.00 & 0.00 & 0.00 & 0.00 & 0.00 & 0.00 & 0.00 & 0.00 & 0.00 \\
Adaptive-RMU           & 0.00 & 0.00 & 0.00 & 0.00 & 0.00 & 0.00 & 0.00 & 0.00 & 0.00 \\
Adaptive-RMU ($\nu$)   & 0.00 & 0.00 & 0.00 & 0.00 & 0.00 & 0.00 & 0.00 & 0.00 & 0.00 \\
NPO                    & 0.00 & 0.00 & 0.00 & 0.00 & 0.00 & 0.00 & 0.01 & 0.00 & 0.00 \\
NPO ($\nu$)            & 0.00 & 0.00 & 0.00 & 0.00 & 0.00 & 0.00 & 0.01 & 0.00 & 0.00 \\
SimNPO                 & 0.00 & 0.00 & 0.00 & 0.00 & 0.00 & 0.00 & 0.01 & 0.00 & 0.00 \\
SimNPO ($\nu$)         & 0.00 & 0.00 & 0.00 & 0.00 & 0.00 & 0.00 & 0.01 & 0.00 & 0.00 \\
DPO                    & 0.00 & 0.00 & 0.00 & 0.00 & 0.00 & 0.00 & 0.01 & 0.00 & 0.00 \\
DPO ($\nu$)            & 0.00 & 0.00 & 0.00 & 0.00 & 0.00 & 0.00 & 0.01 & 0.00 & 0.00 \\
LoKU+FILA              & 3.12 & 3.01 & 3.09 & 2.76 & 2.72 & 3.23 & 2.21 & 2.18 & 2.12 \\
GA                     & 0.11 & 0.16 & 0.01 & 0.01 & 0.01 & 0.00 & 2.08 & 2.50 & 1.04 \\
GA ($\nu$)             & 0.10 & 0.15 & 0.01 & 0.01 & 0.01 & 0.00 & 2.21 & 2.63 & 1.11 \\
GD                     & 0.08 & 0.12 & 0.01 & 0.01 & 0.01 & 0.00 & 1.16 & 1.43 & 0.82 \\
GD ($\nu$)             & 0.07 & 0.10 & 0.00 & 0.01 & 0.01 & 0.00 & 1.05 & 1.27 & 0.75 \\
SPUL                   & 0.63 & 0.67 & 1.29 & 0.63 & 0.68 & 1.26 & 0.07 & 0.08 & 0.08 \\
\bottomrule
\end{tabular}
\end{table}

\begin{table}[t]
\centering
\small
\caption{TRIAGE class assigned to each method on each model for TOFU. \textsc{n} = no-op, \textsc{p} = partially localized, \textsc{c} = collateral dominant, \textsc{g} = globally destructive; \textendash{} = not run. Classes are assigned from the CoFi aggregates alone; CHess and perplexity take no part in the assignment.}
\label{tab:tofu-class-summary}
\begin{tabular}{l ccc}
\toprule
\textbf{Method} & \textbf{Llama-3.1-8B} & \textbf{Llama-3.2-3B} & \textbf{Zephyr-7B-$\beta$} \\
\midrule
ATU                    & \textsc{n} & \textsc{n} & \textsc{n} \\
Obliviate              & \textsc{n} & \textsc{n} & \textsc{n} \\
RSV                    & \textsc{n} & \textsc{n} & \textsc{n} \\
RSV ($\nu$)            & \textsc{n} & \textsc{n} & \textsc{n} \\
RMU                    & \textsc{n} & \textsc{n} & \textsc{n} \\
RMU ($\nu$)            & \textsc{n} & \textsc{n} & \textsc{n} \\
Adaptive-RMU           & \textsc{n} & \textsc{n} & \textsc{n} \\
Adaptive-RMU ($\nu$)   & \textsc{n} & \textsc{n} & \textsc{n} \\
NPO                    & \textsc{n} & \textsc{n} & \textsc{n} \\
NPO ($\nu$)            & \textsc{n} & \textsc{n} & \textsc{n} \\
SimNPO                 & \textsc{n} & \textsc{n} & \textsc{n} \\
SimNPO ($\nu$)         & \textsc{n} & \textsc{n} & \textsc{n} \\
DPO                    & \textsc{n} & \textsc{n} & \textsc{n} \\
DPO ($\nu$)            & \textsc{n} & \textsc{n} & \textsc{n} \\
LoKU+FILA              & \textsc{g} & \textsc{g} & \textsc{g} \\
GA                     & \textsc{p} & \textsc{n} & \textsc{p} \\
GA ($\nu$)             & \textsc{p} & \textsc{n} & \textsc{p} \\
GD                     & \textsc{p} & \textsc{n} & \textsc{c} \\
GD ($\nu$)             & \textsc{p} & \textsc{n} & \textsc{c} \\
SPUL                   & \textsc{n} & \textsc{n} & \textsc{n} \\
\bottomrule
\end{tabular}
\end{table}

\begin{table}[t]
\centering
\footnotesize
\setlength{\tabcolsep}{3pt}
\caption{MUSE-Books per-corpus CoFi and CHess relative shifts (\%) on \textbf{Llama-3.1-8B}. Point estimates are reported without confidence intervals on the three MUSE-Books splits, whose document counts (4, 12 and 13) are smaller than the 200-document subset size, so the three subsets coincide; only the \textit{wiki} column, drawn from WikiText, supports a genuine 95\% CI and is the one column shown with one; the small spreads CHess shows on the book splits reflect the variance of its estimator, not sampling variation. $\uparrow$ on the forget split = stronger unlearning; $\downarrow$ elsewhere = less collateral damage. Methods are grouped by the TRIAGE class they fall into \emph{for this model and this benchmark}, from the CoFi aggregates alone. ($\nu$) marks the randomly-perturbed variant.}
\label{tab:books-cofi-chess-llama8b}
\begin{tabular}{l rrrr @{\hspace{6pt}} rrrr}
\toprule
& \multicolumn{4}{c}{\textbf{CoFi (\%)}} & \multicolumn{4}{c}{\textbf{CHess (\%)}} \\
\cmidrule(lr){2-5}\cmidrule(lr){6-9}
\textbf{Method} & \textit{books-F} & \textit{books-R1} & \textit{books-R2} & \textit{wiki} & \textit{books-F} & \textit{books-R1} & \textit{books-R2} & \textit{wiki} \\
\midrule
\multicolumn{9}{l}{\textit{No-op class}} \\
ATU                    & 0.04 & 0.03 & 0.03 & 0.42\ci{0.63} & 29.3 & 26.2 & 26.2 & 29.4\ci{1.6} \\
Obliviate              & 0.38 & 0.42 & 0.39 & 1.45\ci{0.81} & 30.3 & 28.3 & 27.8 & 30.3\ci{1.4} \\
RSV                    & 0.03 & 0.03 & 0.03 & 0.39\ci{0.57} & 29.2 & 26.1 & 26.1 & 29.6\ci{1.6} \\
RSV ($\nu$)            & 0.03 & 0.03 & 0.03 & 0.29\ci{0.27} & 29.2 & 26.1 & 26.1 & 29.5\ci{1.4} \\
\midrule
\multicolumn{9}{l}{\textit{Partially-localized class}} \\
RMU                    & 4.93 & 0.03 & 0.03 & 0.71\ci{0.69} & 39.3 & 26.1 & 26.1 & 29.5\ci{1.4} \\
RMU ($\nu$)            & 4.81 & 0.03 & 0.03 & 0.38\ci{0.15} & 39.0 & 26.1 & 26.1 & 29.6\ci{1.3} \\
Adaptive-RMU           & 5.41 & 0.03 & 0.03 & 0.27\ci{0.17} & 39.6 & 26.1 & 26.1 & 29.6\ci{1.5} \\
Adaptive-RMU ($\nu$)   & 5.34 & 0.03 & 0.03 & 0.32\ci{0.25} & 39.6 & 26.1 & 26.1 & 29.5\ci{1.4} \\
NPO                    & 13.29 & 0.03 & 0.03 & 0.31\ci{0.23} & 70.7 & 26.1 & 26.1 & 29.5\ci{1.5} \\
NPO ($\nu$)            & 13.42 & 0.03 & 0.03 & 0.26\ci{0.12} & 70.4 & 26.1 & 26.1 & 29.4\ci{1.2} \\
SimNPO                 & 13.16 & 0.03 & 0.03 & 0.26\ci{0.15} & 69.9 & 26.1 & 26.1 & 29.6\ci{1.3} \\
SimNPO ($\nu$)         & 16.39 & 0.03 & 0.03 & 0.30\ci{0.20} & 73.7 & 26.1 & 26.1 & 29.5\ci{1.3} \\
DPO                    & 13.38 & 0.03 & 0.03 & 0.57\ci{1.26} & 68.8 & 26.1 & 26.1 & 29.4\ci{1.3} \\
DPO ($\nu$)            & 11.32 & 0.03 & 0.03 & 0.26\ci{0.17} & 59.0 & 26.1 & 26.2 & 29.4\ci{1.4} \\
\midrule
\multicolumn{9}{l}{\textit{Globally destructive class}} \\
LoKU+FILA              & 22.03 & 20.68 & 19.91 & 27.00\ci{2.87} & 66.3 & 67.3 & 67.2 & 70.5\ci{7.5} \\
GA                     & 4.32 & 1.27 & 2.51 & 16.44\ci{8.04} & 56.0 & 42.0 & 38.0 & 56.5\ci{23.4} \\
GA ($\nu$)             & 3.27 & 5.74 & 3.68 & 23.02\ci{5.06} & 53.5 & 46.0 & 41.7 & 67.0\ci{5.2} \\
GD                     & 4.34 & 3.37 & 1.72 & 14.76\ci{9.25} & 56.6 & 37.5 & 37.5 & 57.5\ci{15.8} \\
GD ($\nu$)             & 7.85 & 5.78 & 3.80 & 21.11\ci{4.34} & 54.5 & 47.1 & 42.6 & 69.8\ci{12.3} \\
SPUL                   & 7.97 & 6.60 & 5.99 & 18.82\ci{4.73} & 71.2 & 45.4 & 60.2 & 62.8\ci{7.6} \\
\bottomrule
\end{tabular}
\end{table}

\begin{table}[t]
\centering
\footnotesize
\setlength{\tabcolsep}{3pt}
\caption{MUSE-Books per-corpus CoFi and CHess relative shifts (\%) on \textbf{Llama-3.2-3B}. Point estimates are reported without confidence intervals on the three MUSE-Books splits, whose document counts (4, 12 and 13) are smaller than the 200-document subset size, so the three subsets coincide; only the \textit{wiki} column, drawn from WikiText, supports a genuine 95\% CI and is the one column shown with one; the small spreads CHess shows on the book splits reflect the variance of its estimator, not sampling variation. $\uparrow$ on the forget split = stronger unlearning; $\downarrow$ elsewhere = less collateral damage. Methods are grouped by the TRIAGE class they fall into \emph{for this model and this benchmark}, from the CoFi aggregates alone. ($\nu$) marks the randomly-perturbed variant.}
\label{tab:books-cofi-chess-llama3b}
\begin{tabular}{l rrrr @{\hspace{6pt}} rrrr}
\toprule
& \multicolumn{4}{c}{\textbf{CoFi (\%)}} & \multicolumn{4}{c}{\textbf{CHess (\%)}} \\
\cmidrule(lr){2-5}\cmidrule(lr){6-9}
\textbf{Method} & \textit{books-F} & \textit{books-R1} & \textit{books-R2} & \textit{wiki} & \textit{books-F} & \textit{books-R1} & \textit{books-R2} & \textit{wiki} \\
\midrule
\multicolumn{9}{l}{\textit{No-op class}} \\
ATU                    & 0.03 & 0.03 & 0.03 & 0.10\ci{0.05} & 22.0 & 23.3 & 23.1 & 27.2\ci{1.2} \\
Obliviate              & 0.25 & 0.37 & 0.35 & 1.01\ci{0.46} & 24.3 & 25.8 & 25.6 & 28.7\ci{1.3} \\
RSV                    & 0.03 & 0.03 & 0.03 & 0.21\ci{0.10} & 22.1 & 23.3 & 23.1 & 27.8\ci{1.6} \\
RSV ($\nu$)            & 0.03 & 0.03 & 0.03 & 0.22\ci{0.16} & 22.1 & 23.3 & 23.1 & 28.1\ci{1.7} \\
\midrule
\multicolumn{9}{l}{\textit{Partially-localized class}} \\
RMU                    & 5.26 & 0.03 & 0.03 & 0.21\ci{0.11} & 40.4 & 23.3 & 23.1 & 28.2\ci{1.6} \\
RMU ($\nu$)            & 4.99 & 0.03 & 0.03 & 0.20\ci{0.11} & 39.2 & 23.3 & 23.1 & 27.9\ci{1.8} \\
Adaptive-RMU           & 3.31 & 0.03 & 0.03 & 0.28\ci{0.25} & 33.9 & 23.3 & 23.1 & 28.0\ci{1.1} \\
Adaptive-RMU ($\nu$)   & 3.26 & 0.03 & 0.03 & 0.31\ci{0.42} & 33.6 & 23.3 & 23.1 & 27.9\ci{0.3} \\
NPO                    & 21.03 & 0.03 & 0.03 & 0.20\ci{0.11} & 74.6 & 23.3 & 23.1 & 27.9\ci{0.7} \\
NPO ($\nu$)            & 11.96 & 0.03 & 0.04 & 0.21\ci{0.11} & 66.8 & 23.3 & 23.1 & 27.9\ci{1.4} \\
SimNPO                 & 16.17 & 0.03 & 0.03 & 0.20\ci{0.10} & 70.4 & 23.3 & 23.1 & 27.9\ci{1.7} \\
SimNPO ($\nu$)         & 17.08 & 0.03 & 0.04 & 0.19\ci{0.10} & 67.2 & 23.3 & 23.1 & 27.4\ci{1.3} \\
DPO                    & 9.62 & 0.03 & 0.10 & 0.12\ci{0.06} & 61.9 & 23.3 & 23.1 & 27.2\ci{1.2} \\
DPO ($\nu$)            & 8.25 & 0.03 & 0.04 & 0.20\ci{0.09} & 56.7 & 23.3 & 23.1 & 27.4\ci{1.0} \\
\midrule
\multicolumn{9}{l}{\textit{Globally destructive class}} \\
LoKU+FILA              & 3.18 & 2.83 & 2.97 & 3.64\ci{0.57} & 40.0 & 38.4 & 41.1 & 39.3\ci{2.3} \\
GA                     & 3.58 & 3.50 & 2.97 & 25.16\ci{2.13} & 52.1 & 46.9 & 43.4 & 75.3\ci{9.2} \\
GA ($\nu$)             & 3.33 & 3.79 & 2.27 & 24.38\ci{4.25} & 53.5 & 47.4 & 44.2 & 73.7\ci{9.3} \\
GD                     & 1.59 & 2.34 & 1.08 & 18.06\ci{3.10} & 51.4 & 40.3 & 39.7 & 69.1\ci{11.1} \\
GD ($\nu$)             & 3.89 & 4.37 & 2.28 & 23.18\ci{6.33} & 52.8 & 46.8 & 45.7 & 72.3\ci{7.1} \\
SPUL                   & 8.12 & 9.59 & 5.27 & 31.31\ci{2.39} & 60.3 & 47.2 & 46.9 & 76.5\ci{2.5} \\
\bottomrule
\end{tabular}
\end{table}

\begin{table}[t]
\centering
\footnotesize
\setlength{\tabcolsep}{3pt}
\caption{MUSE-Books per-corpus CoFi and CHess relative shifts (\%) on \textbf{Qwen3-32B}. Point estimates are reported without confidence intervals on the three MUSE-Books splits, whose document counts (4, 12 and 13) are smaller than the 200-document subset size, so the three subsets coincide; only the \textit{wiki} column, drawn from WikiText, supports a genuine 95\% CI and is the one column shown with one; the small spreads CHess shows on the book splits reflect the variance of its estimator, not sampling variation. $\uparrow$ on the forget split = stronger unlearning; $\downarrow$ elsewhere = less collateral damage. Methods are grouped by the TRIAGE class they fall into \emph{for this model and this benchmark}, from the CoFi aggregates alone. ($\nu$) marks the randomly-perturbed variant.}
\label{tab:books-cofi-chess-qwen32b}
\begin{tabular}{l rrrr @{\hspace{6pt}} rrrr}
\toprule
& \multicolumn{4}{c}{\textbf{CoFi (\%)}} & \multicolumn{4}{c}{\textbf{CHess (\%)}} \\
\cmidrule(lr){2-5}\cmidrule(lr){6-9}
\textbf{Method} & \textit{books-F} & \textit{books-R1} & \textit{books-R2} & \textit{wiki} & \textit{books-F} & \textit{books-R1} & \textit{books-R2} & \textit{wiki} \\
\midrule
\multicolumn{9}{l}{\textit{No-op class}} \\
ATU                    & 0.04 & 0.03 & 0.04 & 0.04\ci{0.00} & 19.2 & 21.8 & 22.4 & 25.6\ci{0.7} \\
Obliviate              & 0.06 & 0.11 & 0.09 & 0.12\ci{0.08} & 21.1 & 23.1 & 23.5 & 25.9\ci{0.7} \\
RSV                    & 0.04 & 0.03 & 0.04 & 0.08\ci{0.06} & 19.2 & 21.5 & 22.1 & 25.6\ci{0.8} \\
RSV ($\nu$)            & 0.04 & 0.03 & 0.03 & 0.04\ci{0.03} & 19.3 & 21.7 & 22.3 & 25.6\ci{0.9} \\
RMU                    & 0.09 & 0.03 & 0.03 & 0.04\ci{0.02} & 22.8 & 21.7 & 22.3 & 25.5\ci{0.9} \\
RMU ($\nu$)            & 0.08 & 0.06 & 0.08 & 0.04\ci{0.06} & 27.0 & 21.7 & 22.3 & 25.5\ci{0.8} \\
Adaptive-RMU           & 0.85 & 0.03 & 0.03 & 0.03\ci{0.01} & 36.4 & 21.7 & 22.2 & 25.6\ci{0.9} \\
Adaptive-RMU ($\nu$)   & 0.77 & 0.12 & 0.04 & 0.05\ci{0.03} & 34.9 & 21.8 & 22.2 & 25.5\ci{0.9} \\
LoKU (no FILA)         & 1.10 & 0.07 & 0.09 & 0.66\ci{0.35} & 28.8 & 22.6 & 23.2 & 29.3\ci{3.5} \\
\midrule
\multicolumn{9}{l}{\textit{Partially-localized class}} \\
NPO                    & 10.35 & 0.03 & 0.05 & 0.05\ci{0.01} & 67.5 & 21.6 & 22.9 & 26.0\ci{0.8} \\
NPO ($\nu$)            & 7.28 & 0.03 & 0.04 & 0.06\ci{0.03} & 63.8 & 21.6 & 22.8 & 26.0\ci{0.8} \\
SimNPO                 & 11.27 & 0.03 & 0.03 & 0.05\ci{0.02} & 68.4 & 21.8 & 22.5 & 25.8\ci{0.8} \\
SimNPO ($\nu$)         & 9.88 & 0.03 & 0.03 & 0.05\ci{0.02} & 66.5 & 21.6 & 22.3 & 25.9\ci{0.6} \\
DPO                    & 5.02 & 0.03 & 0.03 & 0.07\ci{0.01} & 59.3 & 21.7 & 22.7 & 26.1\ci{0.7} \\
DPO ($\nu$)            & 7.11 & 0.04 & 0.03 & 0.07\ci{0.02} & 63.2 & 21.6 & 23.4 & 26.2\ci{1.0} \\
\midrule
\multicolumn{9}{l}{\textit{Globally destructive class}} \\
GA                     & 2.71 & 1.18 & 1.99 & 6.02\ci{6.12} & 55.1 & 30.5 & 36.7 & 50.2\ci{16.8} \\
GA ($\nu$)             & 2.11 & 1.88 & 2.80 & 6.44\ci{6.57} & 54.6 & 27.6 & 33.3 & 49.7\ci{20.4} \\
GD                     & 3.33 & 0.79 & 0.82 & 9.10\ci{6.57} & 54.9 & 27.1 & 29.8 & 51.7\ci{28.2} \\
GD ($\nu$)             & 7.02 & 0.57 & 0.92 & 7.64\ci{3.42} & 58.7 & 27.5 & 33.4 & 48.0\ci{28.1} \\
SPUL                   & 7.00 & 4.33 & 1.01 & 23.86\ci{7.36} & 69.1 & 32.6 & 37.6 & 71.8\ci{1.1} \\
\bottomrule
\end{tabular}
\end{table}

\begin{table}[t]
\centering
\footnotesize
\setlength{\tabcolsep}{3pt}
\caption{MUSE-Books per-corpus CoFi and CHess relative shifts (\%) on \textbf{Zephyr-7B-$\beta$}. Point estimates are reported without confidence intervals on the three MUSE-Books splits, whose document counts (4, 12 and 13) are smaller than the 200-document subset size, so the three subsets coincide; only the \textit{wiki} column, drawn from WikiText, supports a genuine 95\% CI and is the one column shown with one; the small spreads CHess shows on the book splits reflect the variance of its estimator, not sampling variation. $\uparrow$ on the forget split = stronger unlearning; $\downarrow$ elsewhere = less collateral damage. Methods are grouped by the TRIAGE class they fall into \emph{for this model and this benchmark}, from the CoFi aggregates alone. ($\nu$) marks the randomly-perturbed variant.}
\label{tab:books-cofi-chess-zephyr}
\begin{tabular}{l rrrr @{\hspace{6pt}} rrrr}
\toprule
& \multicolumn{4}{c}{\textbf{CoFi (\%)}} & \multicolumn{4}{c}{\textbf{CHess (\%)}} \\
\cmidrule(lr){2-5}\cmidrule(lr){6-9}
\textbf{Method} & \textit{books-F} & \textit{books-R1} & \textit{books-R2} & \textit{wiki} & \textit{books-F} & \textit{books-R1} & \textit{books-R2} & \textit{wiki} \\
\midrule
\multicolumn{9}{l}{\textit{No-op class}} \\
ATU                    & 0.11 & 0.11 & 0.08 & 0.79\ci{0.25} & 55.6 & 56.1 & 55.3 & 58.8\ci{1.7} \\
Adaptive-RMU           & 0.30 & 0.03 & 0.03 & 0.37\ci{0.21} & 55.8 & 56.1 & 55.5 & 58.6\ci{1.8} \\
\midrule
\multicolumn{9}{l}{\textit{Partially-localized class}} \\
Obliviate              & 6.18 & 0.11 & 0.08 & 0.90\ci{0.38} & 56.1 & 56.3 & 55.9 & 59.0\ci{1.3} \\
RSV                    & 23.55 & 0.02 & 0.03 & 0.22\ci{0.26} & 55.6 & 56.2 & 55.3 & 58.6\ci{1.7} \\
RSV ($\nu$)            & 23.15 & 0.03 & 0.03 & 0.26\ci{0.32} & 55.6 & 56.1 & 55.3 & 58.6\ci{1.5} \\
RMU                    & 28.56 & 0.13 & 0.15 & 0.67\ci{0.81} & 55.6 & 56.1 & 55.5 & 58.6\ci{1.7} \\
RMU ($\nu$)            & 28.61 & 0.07 & 1.49 & 4.05\ci{5.73} & 58.2 & 56.3 & 55.5 & 58.6\ci{1.7} \\
Adaptive-RMU ($\nu$)   & 26.34 & 0.04 & 0.15 & 3.68\ci{5.77} & 56.7 & 56.1 & 55.5 & 58.6\ci{1.5} \\
NPO                    & 6.71 & 0.05 & 0.17 & 1.59\ci{2.67} & 119.6 & 94.9 & 99.5 & 96.4\ci{11.1} \\
NPO ($\nu$)            & 10.33 & 0.05 & 0.23 & 1.32\ci{2.13} & 108.4 & 77.0 & 78.3 & 78.7\ci{9.3} \\
SimNPO                 & 5.91 & 0.05 & 0.19 & 2.52\ci{6.14} & 122.5 & 107.7 & 106.4 & 104.5\ci{5.1} \\
SimNPO ($\nu$)         & 7.87 & 0.05 & 0.32 & 2.81\ci{6.59} & 121.6 & 98.1 & 97.6 & 100.0\ci{8.6} \\
DPO                    & 10.04 & 0.08 & 0.37 & 1.25\ci{2.82} & 111.8 & 80.5 & 89.3 & 80.7\ci{6.9} \\
DPO ($\nu$)            & 8.38 & 0.09 & 0.45 & 1.23\ci{1.53} & 124.1 & 95.5 & 104.6 & 99.5\ci{7.5} \\
\midrule
\multicolumn{9}{l}{\textit{Collateral-dominant class}} \\
LoKU+FILA              & 32.20 & 36.73 & 34.18 & 24.60\ci{3.60} & 79.7 & 70.9 & 74.8 & 78.8\ci{6.0} \\
\midrule
\multicolumn{9}{l}{\textit{Globally destructive class}} \\
GA                     & 7.97 & 4.57 & 4.93 & 37.88\ci{9.81} & 168.7 & 155.3 & 156.7 & 162.8\ci{8.1} \\
GA ($\nu$)             & 5.07 & 10.22 & 14.64 & 44.86\ci{4.71} & 169.4 & 155.1 & 157.6 & 169.5\ci{4.9} \\
GD                     & 8.93 & 6.71 & 9.23 & 42.62\ci{9.36} & 160.4 & 144.8 & 145.1 & 153.3\ci{5.6} \\
GD ($\nu$)             & 14.23 & 1.11 & 2.10 & 24.10\ci{17.99} & 151.7 & 125.1 & 127.7 & 137.3\ci{7.1} \\
SPUL                   & 16.81 & 3.49 & 3.29 & 23.02\ci{0.86} & 87.5 & 85.4 & 87.6 & 87.7\ci{12.3} \\
\bottomrule
\end{tabular}
\end{table}

\begin{table}[t]
\centering
\small
\setlength{\tabcolsep}{4pt}
\caption{MUSE-Books perplexity ratios (unlearned/original) for Llama-3.1-8B, Llama-3.2-3B, reported as $\log_{10}$ of the ratio. $0.00$ means fluency is unchanged; $\uparrow$ on the forget split is intended, while any large value on the retain splits or on \textit{wiki} is pure damage. Perplexity is a fluency sanity check and takes no part in the TRIAGE class assignment, which is made from CoFi. Rows follow a fixed method order so the models can be read off against each other; class membership is given in Table~\ref{tab:books-class-summary}.}
\label{tab:books-ppl-1}
\begin{tabular}{l rrrr @{\hspace{6pt}} rrrr}
\toprule
& \multicolumn{4}{c}{\textbf{Llama-3.1-8B}} & \multicolumn{4}{c}{\textbf{Llama-3.2-3B}} \\
\cmidrule(lr){2-5}\cmidrule(lr){6-9}
\textbf{Method} & \textit{books-F} & \textit{books-R1} & \textit{books-R2} & \textit{wiki} & \textit{books-F} & \textit{books-R1} & \textit{books-R2} & \textit{wiki} \\
\midrule
ATU                    & 0.00 & 0.00 & 0.00 & 0.00 & 0.00 & 0.00 & 0.00 & 0.00 \\
Obliviate              & 0.19 & 0.35 & 0.32 & 0.13 & 0.19 & 0.37 & 0.34 & 0.12 \\
RSV                    & 0.00 & 0.00 & 0.00 & 0.00 & 0.00 & 0.00 & 0.00 & 0.00 \\
RSV ($\nu$)            & 0.00 & 0.00 & 0.00 & 0.00 & 0.00 & 0.00 & 0.00 & 0.00 \\
RMU                    & 3.29 & 0.00 & 0.00 & 0.00 & 2.53 & 0.00 & 0.00 & 0.00 \\
RMU ($\nu$)            & 3.18 & 0.00 & 0.00 & 0.00 & 2.39 & 0.00 & 0.00 & 0.00 \\
Adaptive-RMU           & 5.34 & 0.00 & 0.00 & 0.00 & 3.65 & 0.00 & 0.00 & 0.00 \\
Adaptive-RMU ($\nu$)   & 5.35 & 0.00 & 0.00 & 0.01 & 3.63 & 0.00 & 0.00 & 0.00 \\
NPO                    & 32.66 & 0.00 & 0.00 & 0.00 & 31.89 & 0.00 & 0.00 & 0.00 \\
NPO ($\nu$)            & 30.18 & 0.00 & 0.00 & 0.00 & 24.31 & 0.00 & 0.00 & 0.00 \\
SimNPO                 & 30.86 & 0.00 & 0.00 & 0.00 & 25.86 & 0.00 & 0.00 & 0.00 \\
SimNPO ($\nu$)         & 34.17 & 0.00 & 0.00 & 0.00 & 24.45 & 0.00 & 0.00 & 0.00 \\
DPO                    & 34.17 & 0.00 & 0.00 & 0.00 & 29.52 & 0.00 & 0.00 & 0.00 \\
DPO ($\nu$)            & 42.07 & 0.00 & 0.00 & 0.00 & 26.47 & 0.00 & 0.00 & 0.01 \\
LoKU+FILA              & 3.92 & 3.72 & 3.70 & 3.62 & 3.40 & 3.29 & 3.40 & 3.69 \\
GA                     & 55.17 & 0.09 & 0.07 & 0.20 & 34.69 & 0.06 & 0.04 & 0.25 \\
GA ($\nu$)             & 55.25 & 0.11 & 0.08 & 0.31 & 34.66 & 0.05 & 0.04 & 0.21 \\
GD                     & 55.20 & 0.09 & 0.07 & 0.16 & 34.65 & 0.06 & 0.04 & 0.14 \\
GD ($\nu$)             & 55.21 & 0.09 & 0.08 & 0.32 & 34.61 & 0.05 & 0.04 & 0.26 \\
SPUL                   & 26.05 & 14.77 & 14.27 & 14.98 & 16.08 & 1.29 & 1.18 & 2.69 \\
\bottomrule
\end{tabular}
\end{table}

\begin{table}[t]
\centering
\small
\setlength{\tabcolsep}{4pt}
\caption{MUSE-Books perplexity ratios (unlearned/original) for Qwen3-32B, Zephyr-7B-$\beta$, reported as $\log_{10}$ of the ratio. $0.00$ means fluency is unchanged; $\uparrow$ on the forget split is intended, while any large value on the retain splits or on \textit{wiki} is pure damage. Perplexity is a fluency sanity check and takes no part in the TRIAGE class assignment, which is made from CoFi. Rows follow a fixed method order so the models can be read off against each other; class membership is given in Table~\ref{tab:books-class-summary}.}
\label{tab:books-ppl-2}
\begin{tabular}{l rrrr @{\hspace{6pt}} rrrr}
\toprule
& \multicolumn{4}{c}{\textbf{Qwen3-32B}} & \multicolumn{4}{c}{\textbf{Zephyr-7B-$\beta$}} \\
\cmidrule(lr){2-5}\cmidrule(lr){6-9}
\textbf{Method} & \textit{books-F} & \textit{books-R1} & \textit{books-R2} & \textit{wiki} & \textit{books-F} & \textit{books-R1} & \textit{books-R2} & \textit{wiki} \\
\midrule
ATU                    & 0.00 & 0.00 & 0.00 & 0.00 & 0.00 & 0.00 & 0.00 & -0.07 \\
Obliviate              & 0.23 & 0.42 & 0.39 & 0.14 & 2.63 & 0.00 & 0.00 & 0.09 \\
RSV                    & 0.00 & 0.00 & 0.00 & 0.00 & 4.48 & 0.00 & 0.00 & 0.00 \\
RSV ($\nu$)            & 0.00 & 0.00 & 0.00 & 0.00 & 4.50 & 0.00 & 0.00 & 0.00 \\
RMU                    & 0.03 & 0.00 & 0.00 & 0.00 & 5.13 & 0.00 & 0.00 & 0.02 \\
RMU ($\nu$)            & 0.03 & 0.00 & 0.00 & 0.00 & 5.06 & 0.00 & 0.00 & 0.17 \\
Adaptive-RMU           & 3.35 & 0.00 & 0.00 & 0.00 & 0.01 & 0.00 & 0.00 & 0.00 \\
Adaptive-RMU ($\nu$)   & 3.31 & 0.00 & 0.00 & 0.00 & 4.78 & 0.00 & 0.00 & 0.31 \\
NPO                    & 39.10 & 0.00 & 0.00 & 0.02 & 28.09 & 0.00 & 0.00 & 0.57 \\
NPO ($\nu$)            & 34.88 & 0.00 & 0.00 & 0.03 & 29.54 & 0.00 & 0.00 & 0.21 \\
SimNPO                 & 39.23 & 0.00 & 0.00 & 0.02 & 28.08 & 0.00 & 0.00 & 0.80 \\
SimNPO ($\nu$)         & 39.09 & 0.00 & 0.00 & 0.02 & 27.92 & 0.00 & 0.00 & 0.59 \\
DPO                    & 38.28 & 0.00 & 0.00 & 0.03 & 30.15 & 0.00 & 0.00 & 0.11 \\
DPO ($\nu$)            & 37.75 & 0.00 & 0.00 & 0.03 & 28.10 & 0.00 & 0.00 & 0.66 \\
LoKU+FILA              & \textendash & \textendash & \textendash & \textendash & 4.39 & 4.39 & 4.43 & 3.18 \\
LoKU (no FILA)         & 9.79 & -0.42 & 0.04 & 0.08 & \textendash & \textendash & \textendash & \textendash \\
GA                     & 92.80 & 0.06 & 0.02 & 0.30 & 113.40 & 0.04 & 0.03 & 5.75 \\
GA ($\nu$)             & 92.81 & 0.05 & 0.02 & 0.29 & 113.42 & 0.07 & 0.04 & 11.97 \\
GD                     & 92.85 & 0.05 & 0.02 & 0.37 & 113.10 & 0.05 & 0.03 & 9.60 \\
GD ($\nu$)             & 78.41 & 0.04 & 0.01 & 0.25 & 113.00 & 0.03 & 0.03 & 1.32 \\
SPUL                   & 78.98 & -0.03 & -0.03 & 0.56 & 13.05 & 0.10 & 0.10 & 0.09 \\
\bottomrule
\end{tabular}
\end{table}

\begin{table}[t]
\centering
\small
\caption{TRIAGE class assigned to each method on each model for MUSE-Books. \textsc{n} = no-op, \textsc{p} = partially localized, \textsc{c} = collateral dominant, \textsc{g} = globally destructive; \textendash{} = not run. Classes are assigned from the CoFi aggregates alone; CHess and perplexity take no part in the assignment.}
\label{tab:books-class-summary}
\begin{tabular}{l cccc}
\toprule
\textbf{Method} & \textbf{Llama-3.1-8B} & \textbf{Llama-3.2-3B} & \textbf{Qwen3-32B} & \textbf{Zephyr-7B-$\beta$} \\
\midrule
ATU                    & \textsc{n} & \textsc{n} & \textsc{n} & \textsc{n} \\
Obliviate              & \textsc{n} & \textsc{n} & \textsc{n} & \textsc{p} \\
RSV                    & \textsc{n} & \textsc{n} & \textsc{n} & \textsc{p} \\
RSV ($\nu$)            & \textsc{n} & \textsc{n} & \textsc{n} & \textsc{p} \\
RMU                    & \textsc{p} & \textsc{p} & \textsc{n} & \textsc{p} \\
RMU ($\nu$)            & \textsc{p} & \textsc{p} & \textsc{n} & \textsc{p} \\
Adaptive-RMU           & \textsc{p} & \textsc{p} & \textsc{n} & \textsc{n} \\
Adaptive-RMU ($\nu$)   & \textsc{p} & \textsc{p} & \textsc{n} & \textsc{p} \\
NPO                    & \textsc{p} & \textsc{p} & \textsc{p} & \textsc{p} \\
NPO ($\nu$)            & \textsc{p} & \textsc{p} & \textsc{p} & \textsc{p} \\
SimNPO                 & \textsc{p} & \textsc{p} & \textsc{p} & \textsc{p} \\
SimNPO ($\nu$)         & \textsc{p} & \textsc{p} & \textsc{p} & \textsc{p} \\
DPO                    & \textsc{p} & \textsc{p} & \textsc{p} & \textsc{p} \\
DPO ($\nu$)            & \textsc{p} & \textsc{p} & \textsc{p} & \textsc{p} \\
LoKU+FILA              & \textsc{g} & \textsc{g} & \textendash & \textsc{c} \\
LoKU (no FILA)         & \textendash & \textendash & \textsc{n} & \textendash \\
GA                     & \textsc{g} & \textsc{g} & \textsc{g} & \textsc{g} \\
GA ($\nu$)             & \textsc{g} & \textsc{g} & \textsc{g} & \textsc{g} \\
GD                     & \textsc{g} & \textsc{g} & \textsc{g} & \textsc{g} \\
GD ($\nu$)             & \textsc{g} & \textsc{g} & \textsc{g} & \textsc{g} \\
SPUL                   & \textsc{g} & \textsc{g} & \textsc{g} & \textsc{g} \\
\bottomrule
\end{tabular}
\end{table}

\begin{table}[t]
\centering
\footnotesize
\setlength{\tabcolsep}{3pt}
\caption{MUSE-News per-corpus CoFi and CHess relative shifts (\%) on \textbf{Llama-3.1-8B}, mean $\pm$ 95\% CI over three random subsets of 200 samples. $\uparrow$ on the forget split = stronger unlearning; $\downarrow$ on the retain splits and on \textit{wiki} = less collateral damage. Methods are grouped by the TRIAGE class they fall into \emph{for this model and this benchmark}; the assignment uses the CoFi aggregates alone, with CHess reported alongside as a complementary view of how far the answer distribution has moved. ($\nu$) marks the randomly-perturbed variant.}
\label{tab:news-cofi-chess-llama8b}
\begin{tabular}{l rrrr @{\hspace{6pt}} rrrr}
\toprule
& \multicolumn{4}{c}{\textbf{CoFi (\%)}} & \multicolumn{4}{c}{\textbf{CHess (\%)}} \\
\cmidrule(lr){2-5}\cmidrule(lr){6-9}
\textbf{Method} & \textit{news-F} & \textit{news-R1} & \textit{news-R2} & \textit{wiki} & \textit{news-F} & \textit{news-R1} & \textit{news-R2} & \textit{wiki} \\
\midrule
\multicolumn{9}{l}{\textit{No-op class}} \\
ATU                    & 0.03\ci{0.01} & 0.03\ci{0.01} & 0.06\ci{0.13} & 0.46\ci{0.86} & 25.8\ci{0.2} & 26.0\ci{0.6} & 26.4\ci{1.0} & 30.3\ci{1.3} \\
Obliviate              & 0.29\ci{0.00} & 0.30\ci{0.01} & 0.30\ci{0.03} & 0.89\ci{0.84} & 27.5\ci{0.0} & 28.1\ci{0.3} & 28.3\ci{0.5} & 31.2\ci{1.5} \\
RSV                    & 0.03\ci{0.00} & 0.03\ci{0.01} & 0.03\ci{0.00} & 0.24\ci{0.23} & 25.8\ci{0.2} & 26.0\ci{0.5} & 26.4\ci{1.0} & 30.3\ci{1.1} \\
RSV ($\nu$)            & 0.03\ci{0.01} & 0.03\ci{0.00} & 0.03\ci{0.00} & 0.26\ci{0.27} & 25.8\ci{0.2} & 26.0\ci{0.6} & 26.4\ci{1.1} & 30.2\ci{1.3} \\
RMU                    & 0.03\ci{0.00} & 0.03\ci{0.02} & 0.09\ci{0.25} & 0.19\ci{0.11} & 25.8\ci{0.2} & 26.0\ci{0.5} & 26.4\ci{1.0} & 30.3\ci{1.2} \\
RMU ($\nu$)            & 0.03\ci{0.00} & 0.05\ci{0.05} & 0.05\ci{0.04} & 0.31\ci{0.53} & 25.8\ci{0.2} & 26.0\ci{0.5} & 26.4\ci{0.9} & 30.3\ci{1.1} \\
Adaptive-RMU           & 0.03\ci{0.00} & 0.03\ci{0.01} & 0.04\ci{0.01} & 0.39\ci{0.64} & 25.8\ci{0.2} & 26.0\ci{0.5} & 26.4\ci{1.0} & 30.3\ci{1.4} \\
Adaptive-RMU ($\nu$)   & 0.03\ci{0.00} & 0.03\ci{0.00} & 0.03\ci{0.01} & 0.28\ci{0.30} & 25.8\ci{0.2} & 26.0\ci{0.5} & 26.4\ci{0.9} & 30.3\ci{1.4} \\
NPO                    & 0.03\ci{0.00} & 0.03\ci{0.01} & 0.03\ci{0.01} & 0.24\ci{0.26} & 25.8\ci{0.2} & 26.0\ci{0.5} & 26.4\ci{1.1} & 30.2\ci{1.3} \\
NPO ($\nu$)            & 0.03\ci{0.00} & 0.03\ci{0.00} & 0.03\ci{0.00} & 0.30\ci{0.46} & 25.8\ci{0.2} & 26.0\ci{0.6} & 26.4\ci{1.0} & 30.2\ci{1.3} \\
SimNPO                 & 0.03\ci{0.01} & 0.03\ci{0.01} & 0.03\ci{0.00} & 0.22\ci{0.17} & 25.8\ci{0.2} & 26.0\ci{0.5} & 26.4\ci{1.0} & 30.4\ci{0.9} \\
SimNPO ($\nu$)         & 0.03\ci{0.00} & 0.05\ci{0.07} & 0.03\ci{0.00} & 0.24\ci{0.24} & 25.8\ci{0.2} & 26.0\ci{0.6} & 26.4\ci{1.0} & 30.3\ci{1.1} \\
DPO                    & 0.03\ci{0.00} & 0.04\ci{0.01} & 0.03\ci{0.01} & 0.48\ci{0.91} & 25.8\ci{0.2} & 26.0\ci{0.5} & 26.4\ci{1.0} & 30.2\ci{1.3} \\
DPO ($\nu$)            & 0.03\ci{0.00} & 0.04\ci{0.03} & 0.03\ci{0.01} & 0.39\ci{0.57} & 25.8\ci{0.2} & 26.0\ci{0.5} & 26.4\ci{1.0} & 30.2\ci{1.2} \\
SPUL                   & 0.03\ci{0.00} & 0.03\ci{0.00} & 0.04\ci{0.04} & 0.65\ci{1.10} & 25.8\ci{0.2} & 26.1\ci{0.5} & 26.4\ci{1.1} & 30.3\ci{1.1} \\
\midrule
\multicolumn{9}{l}{\textit{Globally destructive class}} \\
LoKU+FILA              & 10.38\ci{0.78} & 11.21\ci{3.25} & 11.77\ci{2.50} & 18.46\ci{6.89} & 50.3\ci{0.8} & 50.4\ci{2.3} & 51.4\ci{1.7} & 55.7\ci{2.0} \\
GA                     & 1.34\ci{1.20} & 1.22\ci{1.00} & 1.10\ci{0.50} & 22.87\ci{6.02} & 37.9\ci{0.1} & 38.2\ci{2.0} & 38.1\ci{2.2} & 61.5\ci{2.6} \\
GA ($\nu$)             & 1.60\ci{0.32} & 1.58\ci{0.33} & 1.42\ci{0.11} & 19.43\ci{2.45} & 37.0\ci{0.7} & 36.8\ci{0.5} & 36.4\ci{0.9} & 55.9\ci{2.8} \\
GD                     & 1.17\ci{1.18} & 0.93\ci{0.11} & 0.93\ci{0.09} & 12.19\ci{8.38} & 35.0\ci{0.7} & 35.2\ci{1.1} & 35.2\ci{0.4} & 44.4\ci{6.4} \\
GD ($\nu$)             & 0.81\ci{0.25} & 0.74\ci{0.15} & 0.83\ci{0.14} & 10.47\ci{3.14} & 34.4\ci{0.2} & 34.6\ci{0.6} & 34.6\ci{1.0} & 43.7\ci{3.9} \\
\bottomrule
\end{tabular}
\end{table}

\begin{table}[t]
\centering
\footnotesize
\setlength{\tabcolsep}{3pt}
\caption{MUSE-News per-corpus CoFi and CHess relative shifts (\%) on \textbf{Llama-3.2-3B}, mean $\pm$ 95\% CI over three random subsets of 200 samples. $\uparrow$ on the forget split = stronger unlearning; $\downarrow$ on the retain splits and on \textit{wiki} = less collateral damage. Methods are grouped by the TRIAGE class they fall into \emph{for this model and this benchmark}; the assignment uses the CoFi aggregates alone, with CHess reported alongside as a complementary view of how far the answer distribution has moved. ($\nu$) marks the randomly-perturbed variant.}
\label{tab:news-cofi-chess-llama3b}
\begin{tabular}{l rrrr @{\hspace{6pt}} rrrr}
\toprule
& \multicolumn{4}{c}{\textbf{CoFi (\%)}} & \multicolumn{4}{c}{\textbf{CHess (\%)}} \\
\cmidrule(lr){2-5}\cmidrule(lr){6-9}
\textbf{Method} & \textit{news-F} & \textit{news-R1} & \textit{news-R2} & \textit{wiki} & \textit{news-F} & \textit{news-R1} & \textit{news-R2} & \textit{wiki} \\
\midrule
\multicolumn{9}{l}{\textit{No-op class}} \\
ATU                    & 0.03\ci{0.00} & 0.03\ci{0.00} & 0.03\ci{0.00} & 0.41\ci{0.20} & 22.7\ci{0.3} & 22.8\ci{0.2} & 23.1\ci{0.8} & 27.6\ci{1.1} \\
Obliviate              & 0.25\ci{0.00} & 0.25\ci{0.00} & 0.25\ci{0.01} & 0.89\ci{0.45} & 24.4\ci{0.1} & 24.5\ci{0.1} & 24.7\ci{0.1} & 28.7\ci{1.3} \\
RSV                    & 0.04\ci{0.00} & 0.04\ci{0.00} & 0.04\ci{0.00} & 0.63\ci{0.33} & 22.8\ci{0.2} & 22.9\ci{0.2} & 23.1\ci{0.7} & 28.0\ci{1.0} \\
RSV ($\nu$)            & 0.04\ci{0.00} & 0.04\ci{0.00} & 0.04\ci{0.00} & 0.63\ci{0.33} & 22.8\ci{0.3} & 22.9\ci{0.2} & 23.2\ci{0.7} & 28.1\ci{1.3} \\
RMU                    & 0.04\ci{0.00} & 0.04\ci{0.00} & 0.04\ci{0.00} & 0.61\ci{0.26} & 22.8\ci{0.2} & 22.9\ci{0.2} & 23.2\ci{0.7} & 27.8\ci{1.5} \\
RMU ($\nu$)            & 0.04\ci{0.00} & 0.04\ci{0.00} & 0.04\ci{0.00} & 0.63\ci{0.35} & 22.8\ci{0.2} & 22.8\ci{0.3} & 23.1\ci{0.8} & 27.5\ci{1.1} \\
Adaptive-RMU           & 0.04\ci{0.00} & 0.04\ci{0.01} & 0.04\ci{0.00} & 0.63\ci{0.30} & 22.8\ci{0.2} & 22.9\ci{0.2} & 23.2\ci{0.7} & 28.1\ci{1.4} \\
Adaptive-RMU ($\nu$)   & 0.04\ci{0.00} & 0.04\ci{0.01} & 0.04\ci{0.00} & 0.64\ci{0.30} & 22.8\ci{0.2} & 22.9\ci{0.2} & 23.2\ci{0.7} & 28.0\ci{1.2} \\
NPO                    & 0.03\ci{0.00} & 0.03\ci{0.00} & 0.03\ci{0.00} & 0.43\ci{0.20} & 22.8\ci{0.1} & 22.8\ci{0.2} & 23.1\ci{0.9} & 27.5\ci{1.3} \\
NPO ($\nu$)            & 0.03\ci{0.00} & 0.03\ci{0.00} & 0.03\ci{0.00} & 0.41\ci{0.21} & 22.7\ci{0.3} & 22.8\ci{0.2} & 23.1\ci{0.8} & 27.5\ci{1.2} \\
SimNPO                 & 0.03\ci{0.00} & 0.03\ci{0.00} & 0.03\ci{0.00} & 0.37\ci{0.18} & 22.7\ci{0.3} & 22.8\ci{0.2} & 23.1\ci{0.8} & 27.4\ci{1.5} \\
SimNPO ($\nu$)         & 0.03\ci{0.00} & 0.03\ci{0.00} & 0.03\ci{0.00} & 0.40\ci{0.21} & 22.7\ci{0.3} & 22.8\ci{0.2} & 23.1\ci{0.8} & 27.4\ci{1.2} \\
DPO                    & 0.03\ci{0.00} & 0.03\ci{0.00} & 0.03\ci{0.00} & 0.44\ci{0.25} & 22.8\ci{0.3} & 22.8\ci{0.2} & 23.1\ci{0.8} & 27.5\ci{1.1} \\
DPO ($\nu$)            & 0.04\ci{0.00} & 0.04\ci{0.00} & 0.04\ci{0.00} & 0.53\ci{0.27} & 22.8\ci{0.3} & 22.9\ci{0.3} & 23.1\ci{0.8} & 27.9\ci{1.1} \\
SPUL                   & 0.03\ci{0.00} & 0.03\ci{0.00} & 0.03\ci{0.00} & 0.28\ci{0.28} & 22.7\ci{0.3} & 22.8\ci{0.2} & 23.1\ci{0.7} & 27.0\ci{1.3} \\
\midrule
\multicolumn{9}{l}{\textit{Globally destructive class}} \\
LoKU+FILA              & 7.93\ci{0.28} & 8.08\ci{0.45} & 8.07\ci{0.28} & 7.37\ci{0.49} & 55.8\ci{2.4} & 55.9\ci{3.2} & 56.9\ci{2.4} & 50.7\ci{3.7} \\
GA                     & 1.64\ci{0.13} & 1.88\ci{0.56} & 1.72\ci{0.17} & 13.58\ci{3.82} & 34.5\ci{1.0} & 34.5\ci{1.9} & 34.4\ci{1.1} & 52.7\ci{2.9} \\
GA ($\nu$)             & 1.68\ci{0.12} & 2.05\ci{1.15} & 1.82\ci{0.26} & 14.31\ci{2.56} & 35.5\ci{1.0} & 35.5\ci{2.4} & 35.5\ci{1.0} & 55.1\ci{3.3} \\
GD                     & 2.59\ci{0.09} & 2.63\ci{0.28} & 2.73\ci{0.71} & 11.21\ci{2.55} & 42.4\ci{1.5} & 42.5\ci{0.5} & 42.1\ci{0.1} & 52.2\ci{1.6} \\
GD ($\nu$)             & 2.95\ci{0.07} & 2.92\ci{0.10} & 2.94\ci{0.14} & 9.13\ci{4.79} & 45.9\ci{0.3} & 45.4\ci{0.6} & 45.6\ci{0.6} & 49.1\ci{2.5} \\
\bottomrule
\end{tabular}
\end{table}

\begin{table}[t]
\centering
\footnotesize
\setlength{\tabcolsep}{3pt}
\caption{MUSE-News per-corpus CoFi and CHess relative shifts (\%) on \textbf{Zephyr-7B-$\beta$}, mean $\pm$ 95\% CI over three random subsets of 200 samples. $\uparrow$ on the forget split = stronger unlearning; $\downarrow$ on the retain splits and on \textit{wiki} = less collateral damage. Methods are grouped by the TRIAGE class they fall into \emph{for this model and this benchmark}; the assignment uses the CoFi aggregates alone, with CHess reported alongside as a complementary view of how far the answer distribution has moved. ($\nu$) marks the randomly-perturbed variant.}
\label{tab:news-cofi-chess-zephyr}
\begin{tabular}{l rrrr @{\hspace{6pt}} rrrr}
\toprule
& \multicolumn{4}{c}{\textbf{CoFi (\%)}} & \multicolumn{4}{c}{\textbf{CHess (\%)}} \\
\cmidrule(lr){2-5}\cmidrule(lr){6-9}
\textbf{Method} & \textit{news-F} & \textit{news-R1} & \textit{news-R2} & \textit{wiki} & \textit{news-F} & \textit{news-R1} & \textit{news-R2} & \textit{wiki} \\
\midrule
\multicolumn{9}{l}{\textit{No-op class}} \\
ATU                    & 0.03\ci{0.00} & 0.05\ci{0.01} & 0.06\ci{0.10} & 0.32\ci{0.19} & 52.5\ci{2.2} & 53.1\ci{3.0} & 53.1\ci{3.1} & 57.0\ci{3.7} \\
Obliviate              & 0.10\ci{0.01} & 0.11\ci{0.02} & 0.10\ci{0.01} & 1.35\ci{0.38} & 52.6\ci{2.3} & 53.0\ci{2.2} & 53.1\ci{2.9} & 57.1\ci{3.5} \\
RSV                    & 0.02\ci{0.00} & 0.02\ci{0.00} & 0.02\ci{0.00} & 0.09\ci{0.04} & 52.5\ci{2.2} & 52.9\ci{2.7} & 53.0\ci{3.3} & 57.1\ci{3.8} \\
RSV ($\nu$)            & 0.02\ci{0.00} & 0.02\ci{0.00} & 0.05\ci{0.10} & 0.08\ci{0.04} & 52.5\ci{2.6} & 52.9\ci{2.8} & 53.1\ci{3.1} & 57.1\ci{3.5} \\
RMU                    & 0.02\ci{0.00} & 0.02\ci{0.00} & 0.04\ci{0.07} & 0.10\ci{0.05} & 52.5\ci{2.2} & 53.0\ci{2.5} & 53.0\ci{3.3} & 57.1\ci{3.5} \\
RMU ($\nu$)            & 0.02\ci{0.00} & 0.02\ci{0.00} & 0.04\ci{0.04} & 0.10\ci{0.05} & 52.6\ci{2.1} & 52.9\ci{2.8} & 53.0\ci{3.5} & 57.0\ci{3.7} \\
Adaptive-RMU           & 0.02\ci{0.00} & 0.02\ci{0.00} & 0.03\ci{0.05} & 0.08\ci{0.02} & 52.5\ci{2.7} & 53.0\ci{3.1} & 53.2\ci{3.9} & 57.1\ci{3.7} \\
Adaptive-RMU ($\nu$)   & 0.02\ci{0.00} & 0.02\ci{0.00} & 0.03\ci{0.04} & 0.12\ci{0.14} & 52.5\ci{2.5} & 53.0\ci{2.2} & 53.0\ci{3.3} & 57.1\ci{3.5} \\
NPO                    & 0.04\ci{0.01} & 0.04\ci{0.00} & 0.04\ci{0.03} & 0.21\ci{0.07} & 52.5\ci{2.1} & 53.0\ci{2.3} & 53.1\ci{3.2} & 57.1\ci{3.6} \\
NPO ($\nu$)            & 0.03\ci{0.00} & 0.04\ci{0.01} & 0.03\ci{0.01} & 0.19\ci{0.08} & 52.5\ci{2.6} & 52.9\ci{2.8} & 53.1\ci{3.1} & 57.0\ci{3.7} \\
SimNPO                 & 0.03\ci{0.00} & 0.04\ci{0.01} & 0.04\ci{0.04} & 0.20\ci{0.07} & 52.5\ci{2.4} & 53.1\ci{2.9} & 53.0\ci{3.2} & 57.1\ci{3.7} \\
SimNPO ($\nu$)         & 0.03\ci{0.00} & 0.04\ci{0.01} & 0.05\ci{0.11} & 0.19\ci{0.05} & 52.4\ci{2.4} & 53.0\ci{2.9} & 53.0\ci{3.3} & 57.1\ci{3.8} \\
DPO                    & 0.03\ci{0.00} & 0.04\ci{0.01} & 0.03\ci{0.01} & 0.27\ci{0.21} & 52.5\ci{2.2} & 53.1\ci{2.1} & 53.1\ci{3.1} & 57.0\ci{3.7} \\
DPO ($\nu$)            & 0.04\ci{0.01} & 0.04\ci{0.01} & 0.04\ci{0.03} & 0.26\ci{0.09} & 52.5\ci{2.3} & 53.0\ci{2.6} & 53.3\ci{3.0} & 57.1\ci{3.8} \\
SPUL                   & 0.05\ci{0.01} & 0.05\ci{0.00} & 0.08\ci{0.10} & 0.85\ci{0.36} & 52.6\ci{2.7} & 53.2\ci{3.4} & 53.2\ci{3.5} & 57.1\ci{3.8} \\
\midrule
\multicolumn{9}{l}{\textit{Globally destructive class}} \\
LoKU+FILA              & 35.68\ci{1.79} & 34.19\ci{8.75} & 35.79\ci{14.38} & 31.79\ci{19.22} & 76.5\ci{16.9} & 75.3\ci{3.3} & 79.0\ci{10.9} & 76.3\ci{12.9} \\
GA                     & 1.15\ci{0.05} & 1.12\ci{0.06} & 1.11\ci{0.02} & 2.97\ci{0.95} & 58.4\ci{1.1} & 57.9\ci{1.6} & 58.1\ci{0.3} & 61.1\ci{3.7} \\
GA ($\nu$)             & 1.13\ci{0.05} & 1.10\ci{0.05} & 1.08\ci{0.02} & 2.98\ci{0.99} & 59.9\ci{2.5} & 59.4\ci{2.1} & 59.0\ci{4.3} & 60.9\ci{0.8} \\
GD                     & 1.52\ci{0.07} & 1.50\ci{0.13} & 1.51\ci{0.03} & 3.10\ci{0.93} & 62.9\ci{7.6} & 60.0\ci{2.0} & 60.1\ci{6.1} & 63.2\ci{1.0} \\
GD ($\nu$)             & 1.19\ci{0.09} & 1.18\ci{0.10} & 1.18\ci{0.03} & 2.91\ci{0.71} & 59.5\ci{1.7} & 59.4\ci{1.9} & 58.9\ci{3.6} & 62.2\ci{1.5} \\
\bottomrule
\end{tabular}
\end{table}

\begin{table}[t]
\centering
\small
\setlength{\tabcolsep}{4pt}
\caption{MUSE-News perplexity ratios (unlearned/original) for Llama-3.1-8B, Llama-3.2-3B, reported as $\log_{10}$ of the ratio. $0.00$ means fluency is unchanged; $\uparrow$ on the forget split is intended, while any large value on the retain splits or on \textit{wiki} is pure damage. Perplexity is a fluency sanity check and takes no part in the TRIAGE class assignment, which is made from CoFi. Rows follow a fixed method order so the models can be read off against each other; class membership is given in Table~\ref{tab:news-class-summary}.}
\label{tab:news-ppl-1}
\begin{tabular}{l rrrr @{\hspace{6pt}} rrrr}
\toprule
& \multicolumn{4}{c}{\textbf{Llama-3.1-8B}} & \multicolumn{4}{c}{\textbf{Llama-3.2-3B}} \\
\cmidrule(lr){2-5}\cmidrule(lr){6-9}
\textbf{Method} & \textit{news-F} & \textit{news-R1} & \textit{news-R2} & \textit{wiki} & \textit{news-F} & \textit{news-R1} & \textit{news-R2} & \textit{wiki} \\
\midrule
ATU                    & 0.00 & 0.00 & 0.00 & 0.00 & 0.00 & 0.00 & 0.00 & 0.00 \\
Obliviate              & 0.25 & 0.25 & 0.25 & 0.19 & 0.30 & 0.30 & 0.30 & 0.17 \\
RSV                    & 0.00 & 0.00 & 0.00 & 0.00 & 0.00 & 0.00 & 0.00 & 0.00 \\
RSV ($\nu$)            & 0.00 & 0.00 & 0.00 & 0.00 & 0.00 & 0.00 & 0.00 & 0.00 \\
RMU                    & 0.00 & 0.00 & 0.00 & 0.00 & 0.00 & 0.00 & 0.00 & 0.00 \\
RMU ($\nu$)            & 0.00 & 0.00 & 0.00 & 0.00 & 0.00 & 0.00 & 0.00 & 0.00 \\
Adaptive-RMU           & 0.00 & 0.00 & 0.00 & 0.00 & 0.00 & 0.00 & 0.00 & 0.00 \\
Adaptive-RMU ($\nu$)   & 0.00 & 0.00 & 0.00 & 0.00 & 0.00 & 0.00 & 0.00 & 0.00 \\
NPO                    & 0.00 & 0.00 & 0.00 & 0.00 & 0.00 & 0.00 & 0.00 & 0.00 \\
NPO ($\nu$)            & 0.00 & 0.00 & 0.00 & 0.00 & 0.00 & 0.00 & 0.00 & 0.00 \\
SimNPO                 & 0.00 & 0.00 & 0.00 & 0.00 & 0.00 & 0.00 & 0.00 & 0.00 \\
SimNPO ($\nu$)         & 0.00 & 0.00 & 0.00 & 0.00 & 0.00 & 0.00 & 0.00 & 0.00 \\
DPO                    & 0.00 & 0.00 & 0.00 & 0.00 & 0.00 & 0.00 & 0.00 & 0.00 \\
DPO ($\nu$)            & 0.00 & 0.00 & 0.00 & 0.00 & 0.00 & 0.00 & 0.00 & 0.00 \\
LoKU+FILA              & 3.45 & 3.45 & 3.45 & 3.40 & 3.75 & 3.78 & 3.76 & 3.43 \\
GA                     & 54.75 & 54.64 & 54.63 & 53.98 & 28.84 & 28.73 & 28.76 & 28.10 \\
GA ($\nu$)             & 54.61 & 54.46 & 54.46 & 53.13 & 28.67 & 28.57 & 28.60 & 27.99 \\
GD                     & 54.76 & 54.66 & 54.65 & 54.08 & 25.68 & 25.64 & 25.65 & 26.08 \\
GD ($\nu$)             & 54.75 & 54.65 & 54.64 & 54.17 & 30.90 & 30.89 & 30.90 & 30.63 \\
SPUL                   & 1.04 & 1.06 & 1.07 & 1.33 & 1.38 & 1.33 & 1.37 & 1.36 \\
\bottomrule
\end{tabular}
\end{table}

\begin{table}[t]
\centering
\small
\setlength{\tabcolsep}{4pt}
\caption{MUSE-News perplexity ratios (unlearned/original) for Zephyr-7B-$\beta$, reported as $\log_{10}$ of the ratio. $0.00$ means fluency is unchanged; $\uparrow$ on the forget split is intended, while any large value on the retain splits or on \textit{wiki} is pure damage. Perplexity is a fluency sanity check and takes no part in the TRIAGE class assignment, which is made from CoFi. Rows follow a fixed method order so the models can be read off against each other; class membership is given in Table~\ref{tab:news-class-summary}.}
\label{tab:news-ppl-2}
\begin{tabular}{l rrrr}
\toprule
& \multicolumn{4}{c}{\textbf{Zephyr-7B-$\beta$}} \\
\cmidrule(lr){2-5}
\textbf{Method} & \textit{news-F} & \textit{news-R1} & \textit{news-R2} & \textit{wiki} \\
\midrule
ATU                    & 0.00 & 0.01 & 0.00 & -0.01 \\
Obliviate              & 0.00 & 0.00 & 0.00 & 0.13 \\
RSV                    & 0.00 & 0.00 & 0.00 & 0.00 \\
RSV ($\nu$)            & 0.00 & 0.00 & 0.00 & 0.00 \\
RMU                    & 0.00 & 0.00 & 0.00 & 0.00 \\
RMU ($\nu$)            & 0.00 & 0.00 & 0.00 & 0.00 \\
Adaptive-RMU           & 0.00 & 0.00 & 0.00 & 0.00 \\
Adaptive-RMU ($\nu$)   & 0.00 & 0.00 & 0.00 & 0.00 \\
NPO                    & 0.00 & 0.00 & 0.00 & 0.00 \\
NPO ($\nu$)            & 0.00 & 0.00 & 0.00 & 0.00 \\
SimNPO                 & 0.00 & 0.00 & 0.00 & 0.00 \\
SimNPO ($\nu$)         & 0.00 & 0.00 & 0.00 & 0.00 \\
DPO                    & 0.00 & 0.00 & 0.00 & 0.00 \\
DPO ($\nu$)            & 0.00 & 0.00 & 0.00 & 0.00 \\
LoKU+FILA              & 3.61 & 3.57 & 3.60 & 2.57 \\
GA                     & 28.96 & 28.85 & 28.84 & 28.10 \\
GA ($\nu$)             & 28.97 & 28.86 & 28.85 & 28.11 \\
GD                     & 29.04 & 28.93 & 28.92 & 28.18 \\
GD ($\nu$)             & 28.97 & 28.86 & 28.85 & 28.14 \\
SPUL                   & 0.01 & 0.01 & 0.01 & 0.00 \\
\bottomrule
\end{tabular}
\end{table}

\begin{table}[t]
\centering
\small
\caption{TRIAGE class assigned to each method on each model for MUSE-News. \textsc{n} = no-op, \textsc{p} = partially localized, \textsc{c} = collateral dominant, \textsc{g} = globally destructive; \textendash{} = not run. Classes are assigned from the CoFi aggregates alone; CHess and perplexity take no part in the assignment.}
\label{tab:news-class-summary}
\begin{tabular}{l ccc}
\toprule
\textbf{Method} & \textbf{Llama-3.1-8B} & \textbf{Llama-3.2-3B} & \textbf{Zephyr-7B-$\beta$} \\
\midrule
ATU                    & \textsc{n} & \textsc{n} & \textsc{n} \\
Obliviate              & \textsc{n} & \textsc{n} & \textsc{n} \\
RSV                    & \textsc{n} & \textsc{n} & \textsc{n} \\
RSV ($\nu$)            & \textsc{n} & \textsc{n} & \textsc{n} \\
RMU                    & \textsc{n} & \textsc{n} & \textsc{n} \\
RMU ($\nu$)            & \textsc{n} & \textsc{n} & \textsc{n} \\
Adaptive-RMU           & \textsc{n} & \textsc{n} & \textsc{n} \\
Adaptive-RMU ($\nu$)   & \textsc{n} & \textsc{n} & \textsc{n} \\
NPO                    & \textsc{n} & \textsc{n} & \textsc{n} \\
NPO ($\nu$)            & \textsc{n} & \textsc{n} & \textsc{n} \\
SimNPO                 & \textsc{n} & \textsc{n} & \textsc{n} \\
SimNPO ($\nu$)         & \textsc{n} & \textsc{n} & \textsc{n} \\
DPO                    & \textsc{n} & \textsc{n} & \textsc{n} \\
DPO ($\nu$)            & \textsc{n} & \textsc{n} & \textsc{n} \\
LoKU+FILA              & \textsc{g} & \textsc{g} & \textsc{g} \\
GA                     & \textsc{g} & \textsc{g} & \textsc{g} \\
GA ($\nu$)             & \textsc{g} & \textsc{g} & \textsc{g} \\
GD                     & \textsc{g} & \textsc{g} & \textsc{g} \\
GD ($\nu$)             & \textsc{g} & \textsc{g} & \textsc{g} \\
SPUL                   & \textsc{n} & \textsc{n} & \textsc{n} \\
\bottomrule
\end{tabular}
\end{table}

\subsection{Heatmap Visualizations}
\label{app:heatmaps}

This subsection collects the per-model heatmaps for every benchmark. Each figure shows
one row per method, grouped by TRIAGE class, one column per evaluation corpus, and the
three metrics as separate panels with independent colour scales. The class signatures
described at the start of this section --- a uniformly dark grid for no-op methods, a
left-to-right decay for partially localized ones, an inverted gradient with a dark
final column for collateral dominant ones, and an illuminated WikiText column for
globally destructive ones --- are visible directly in the CoFi panel, and the figures
are intended as a fast qualitative check on the tables rather than as a substitute for
them.

Two readings are worth pointing out. First, the CHess panel is generally brighter and
flatter than the CoFi panel, because CHess responds to any change in the answer
distribution whereas CoFi responds only to changes that shift the correct-answer
ranking; a method can therefore look active in CHess and inert in CoFi. Adaptive-RMU on
Qwen3-32B is the clearest instance, and it is why that method is classified as a no-op
despite a visibly non-trivial CHess row. Second, the PPL panel highlights the forget
columns for the globally destructive methods far more strongly than the WikiText
column, which is an artefact of the log scale: a WikiText perplexity ratio of $10^{2}$
is catastrophic in absolute terms but sits near the bottom of a scale whose top is
$10^{70}$.

\begin{figure}[p]
\centering
\includegraphics[width=0.8\textwidth]{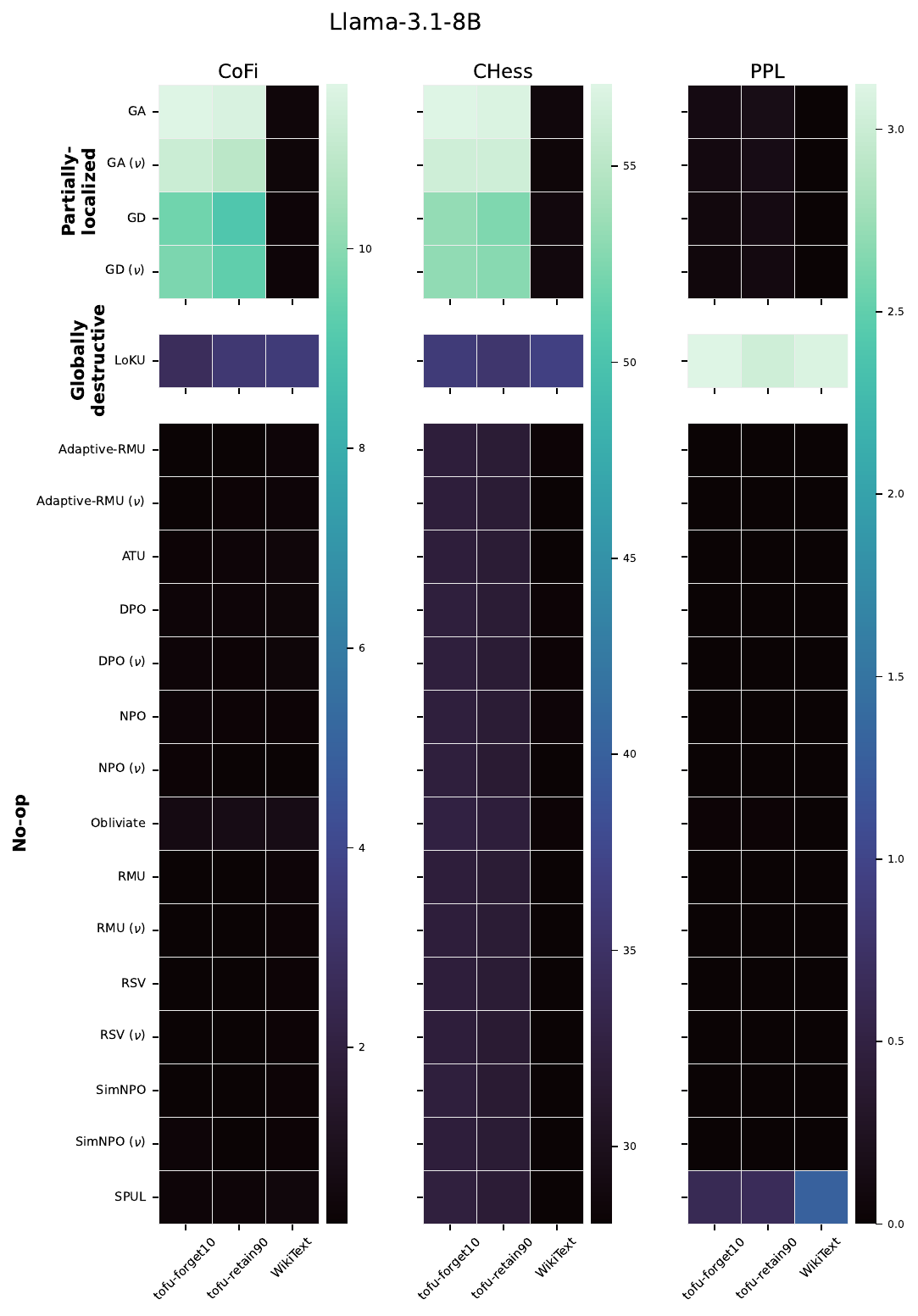}
\caption{TOFU TRIAGE heatmap for Llama-3.1-8B. Rows are grouped by TRIAGE class;
columns are the three evaluation corpora. Colour scales are independent per metric
panel. PPL is shown as $\log_{10}$ of the unlearned/original ratio.}
\label{fig:heatmap-tofu-llama8b}
\end{figure}

\begin{figure}[p]
\centering
\includegraphics[width=0.8\textwidth]{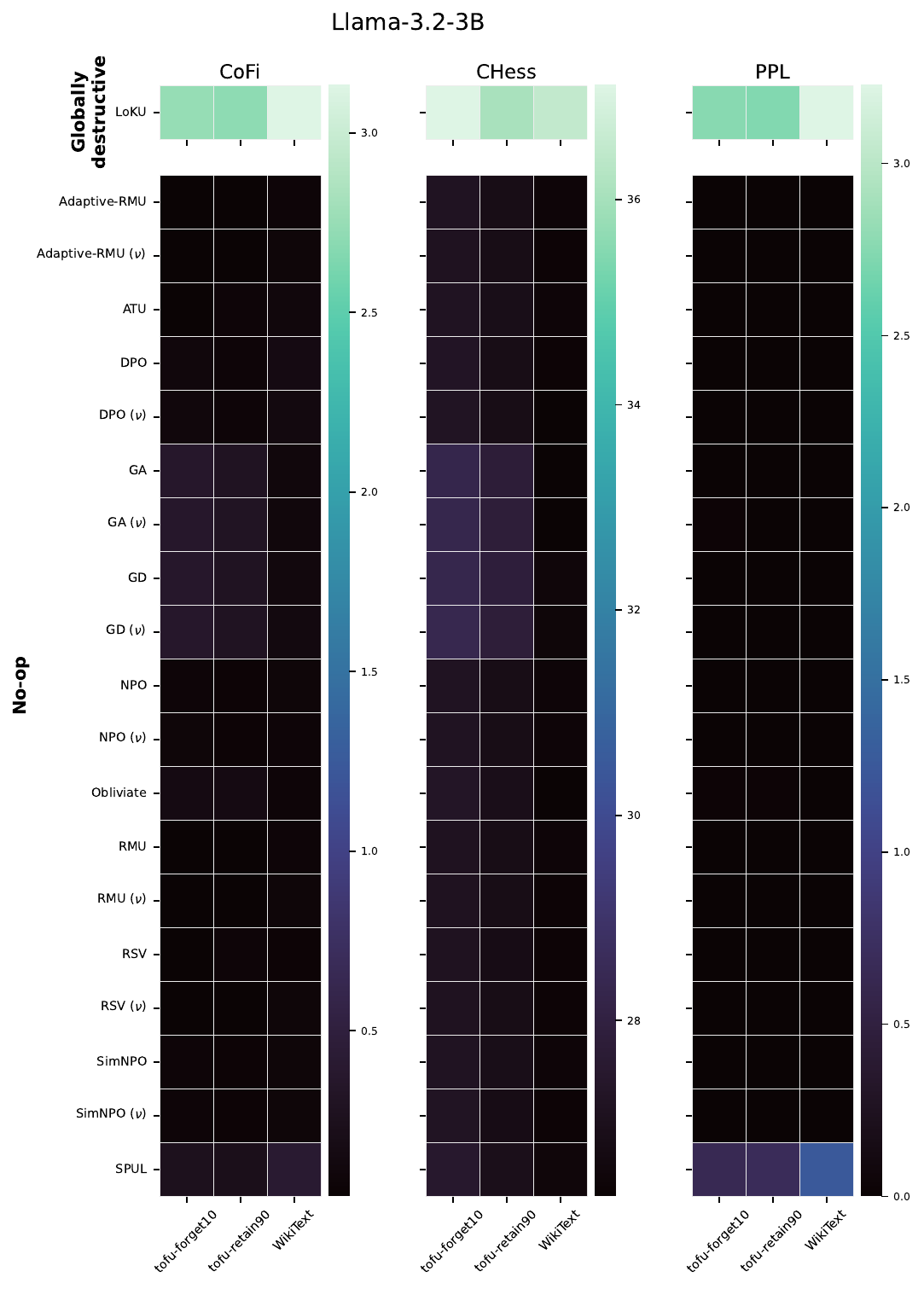}
\caption{TOFU TRIAGE heatmap for Llama-3.2-3B. Rows are grouped by TRIAGE class;
columns are the three evaluation corpora. Colour scales are independent per metric
panel. PPL is shown as $\log_{10}$ of the unlearned/original ratio.}
\label{fig:heatmap-tofu-llama3b}
\end{figure}

\begin{figure}[p]
\centering
\includegraphics[width=0.8\textwidth]{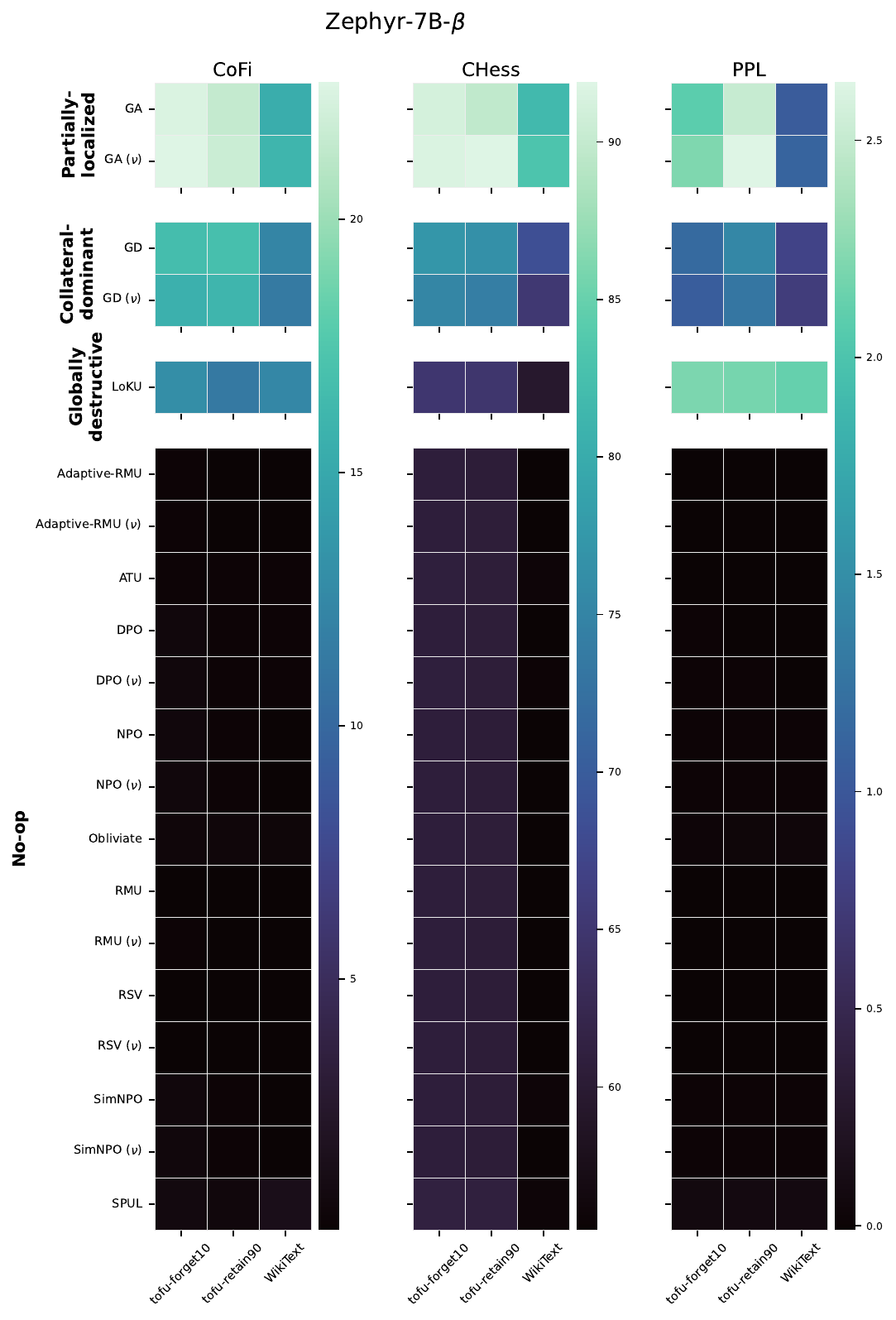}
\caption{TOFU TRIAGE heatmap for Zephyr-7B-$\beta$. Rows are grouped by TRIAGE class;
columns are the three evaluation corpora. Colour scales are independent per metric
panel. PPL is shown as $\log_{10}$ of the unlearned/original ratio.}
\label{fig:heatmap-tofu-zephyr}
\end{figure}

\begin{figure}[p]
\centering
\includegraphics[width=\textwidth]{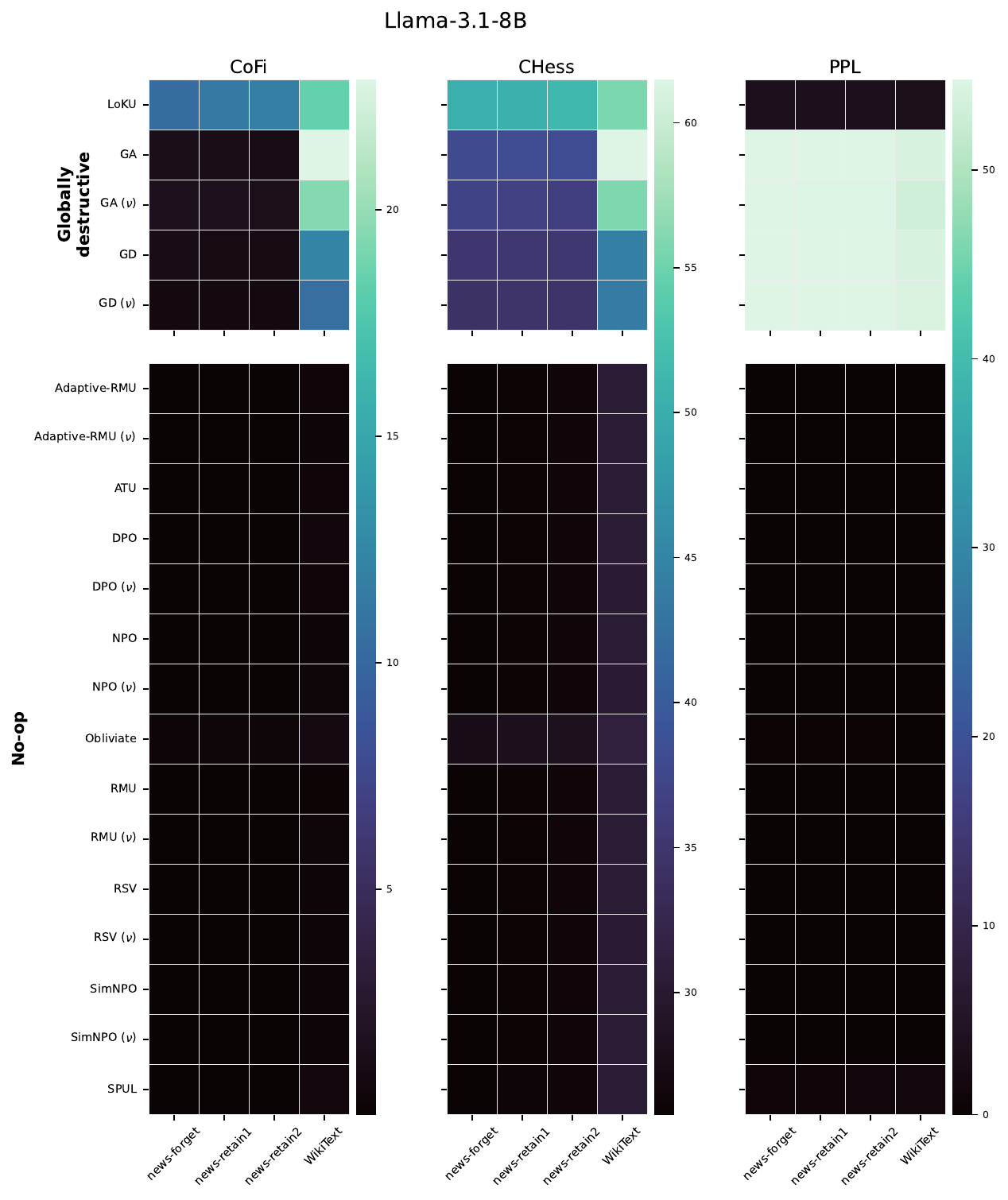}
\caption{MUSE-News TRIAGE heatmap for Llama-3.1-8B. Rows are grouped by TRIAGE class;
columns are the four evaluation corpora. Colour scales are independent per metric
panel. PPL is shown as $\log_{10}$ of the unlearned/original ratio.}
\label{fig:heatmap-news-llama8b}
\end{figure}

\begin{figure}[p]
\centering
\includegraphics[width=\textwidth]{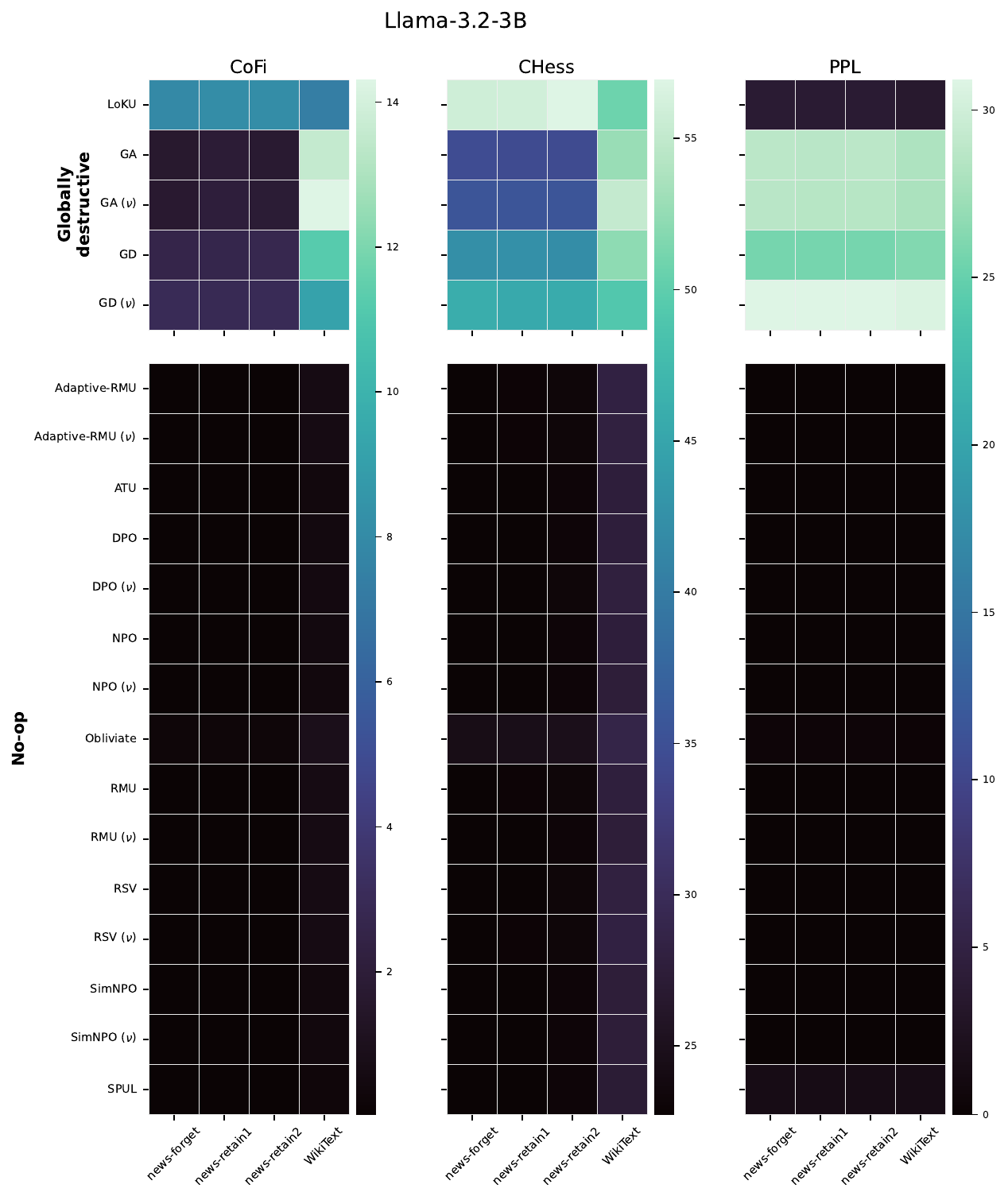}
\caption{MUSE-News TRIAGE heatmap for Llama-3.2-3B. Rows are grouped by TRIAGE class;
columns are the four evaluation corpora. Colour scales are independent per metric
panel. PPL is shown as $\log_{10}$ of the unlearned/original ratio.}
\label{fig:heatmap-news-llama3b}
\end{figure}

\begin{figure}[p]
\centering
\includegraphics[width=\textwidth]{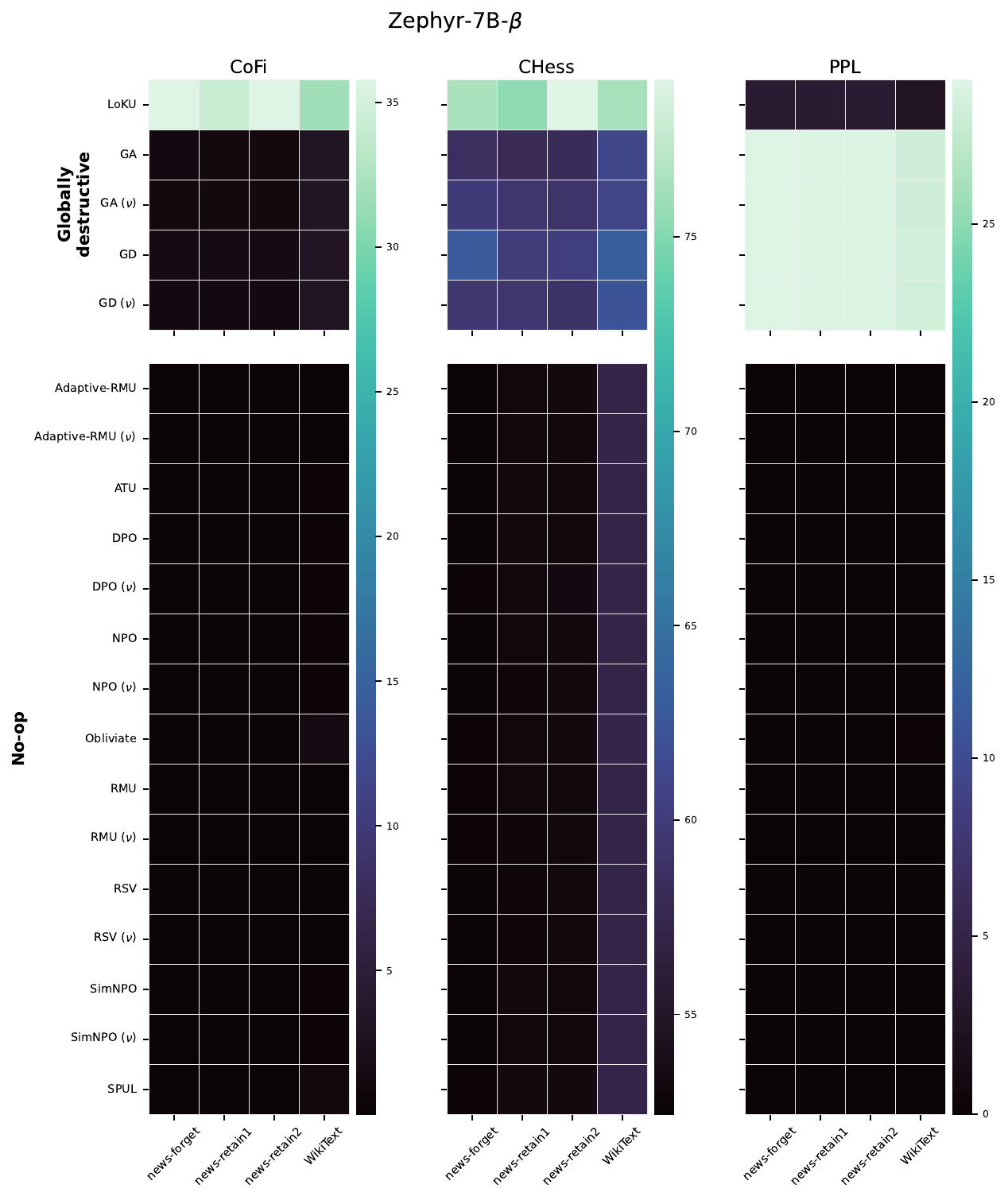}
\caption{MUSE-News TRIAGE heatmap for Zephyr-7B-$\beta$. Rows are grouped by TRIAGE class;
columns are the four evaluation corpora. Colour scales are independent per metric
panel. PPL is shown as $\log_{10}$ of the unlearned/original ratio.}
\label{fig:heatmap-news-zephyr}
\end{figure}

\begin{figure}[p]
\centering
\includegraphics[width=\textwidth]{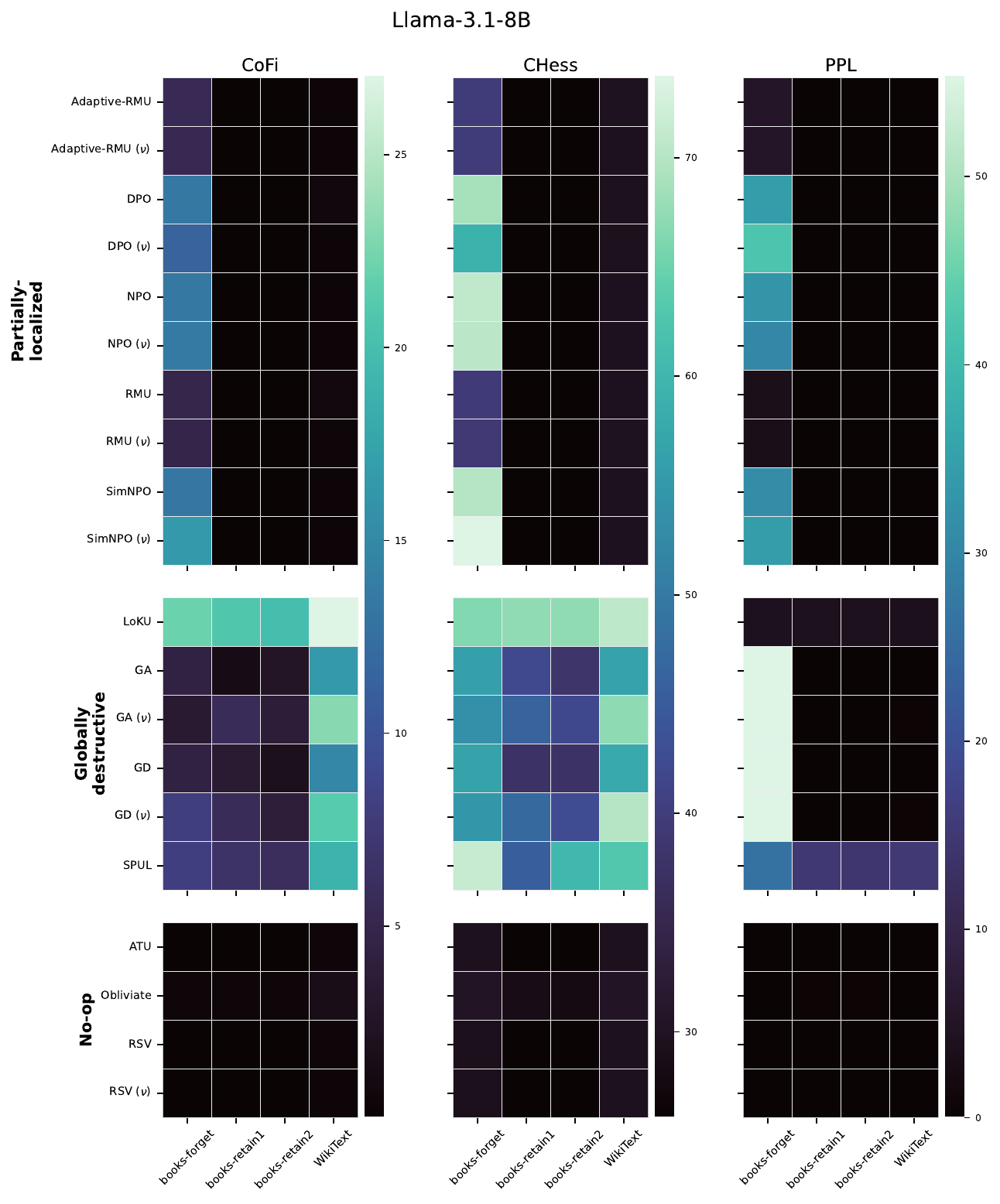}
\caption{MUSE-Books TRIAGE heatmap for Llama-3.1-8B. Rows are grouped by TRIAGE class;
columns are the four evaluation corpora. Colour scales are independent per metric
panel. PPL is shown as $\log_{10}$ of the unlearned/original ratio.}
\label{fig:heatmap-books-llama8b}
\end{figure}

\begin{figure}[p]
\centering
\includegraphics[width=\textwidth]{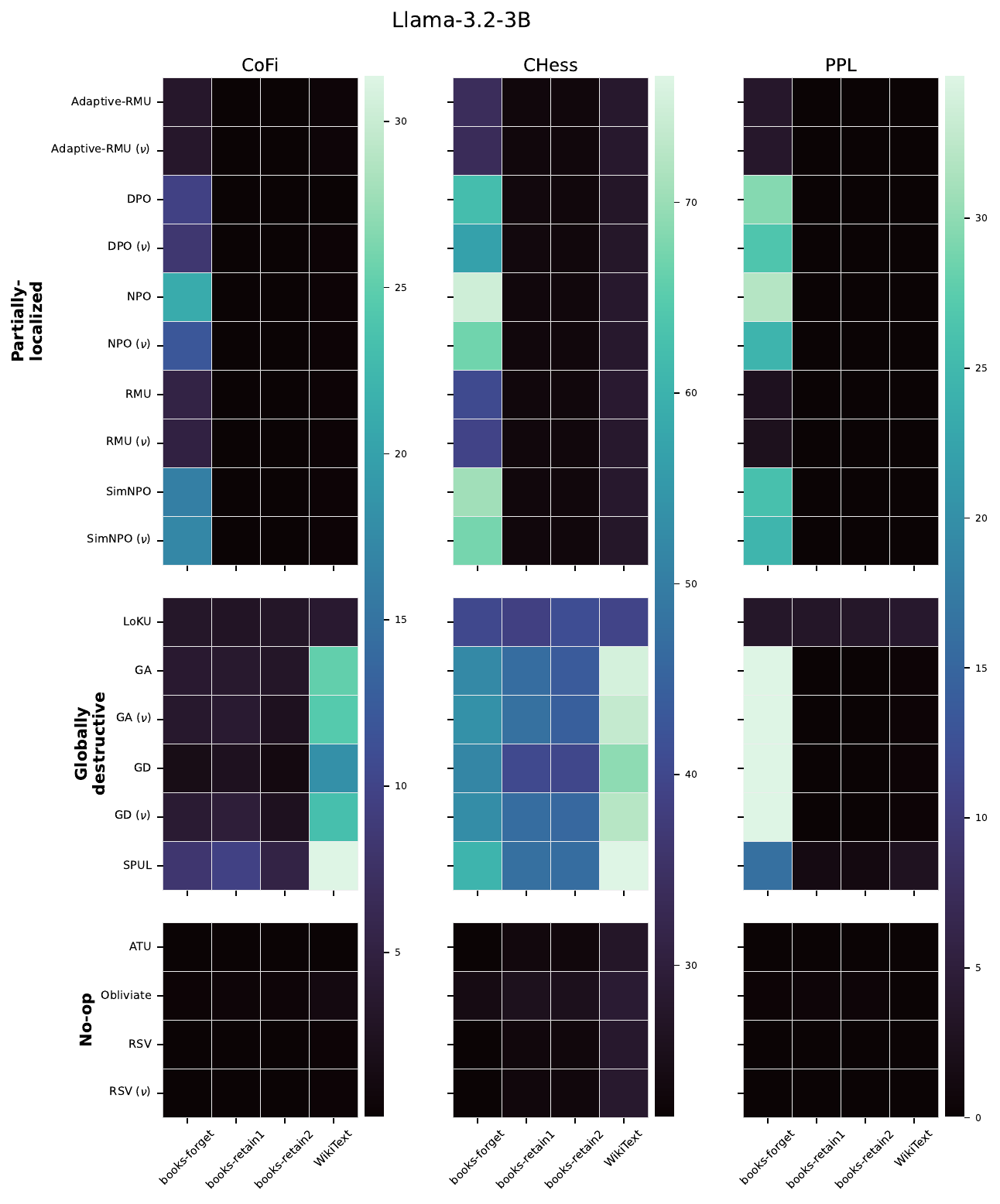}
\caption{MUSE-Books TRIAGE heatmap for Llama-3.2-3B. Rows are grouped by TRIAGE class;
columns are the four evaluation corpora. Colour scales are independent per metric
panel. PPL is shown as $\log_{10}$ of the unlearned/original ratio.}
\label{fig:heatmap-books-llama3b}
\end{figure}

\begin{figure}[p]
\centering
\includegraphics[width=\textwidth]{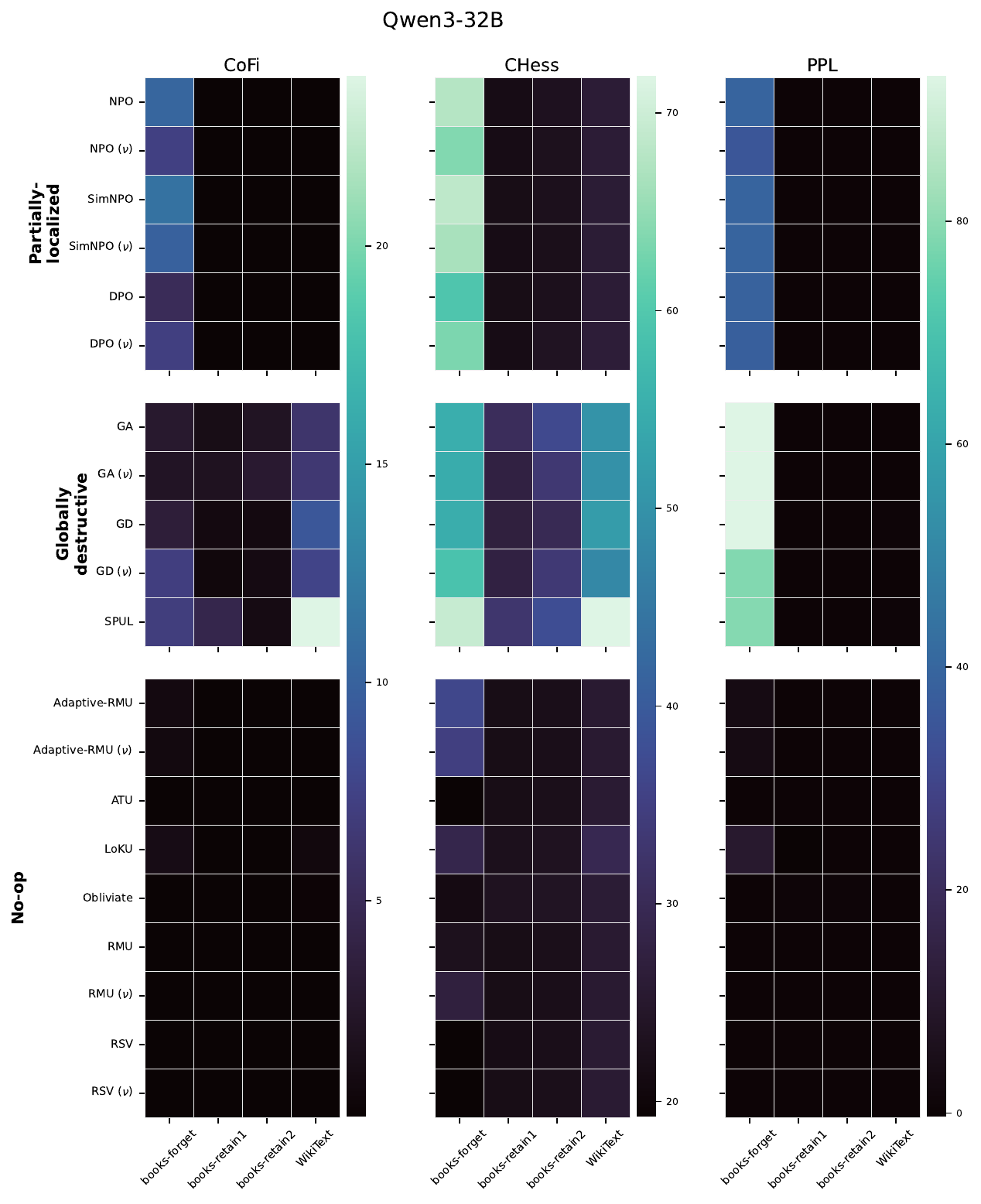}
\caption{MUSE-Books TRIAGE heatmap for Qwen3-32b. Rows are grouped by TRIAGE class;
columns are the four evaluation corpora. Colour scales are independent per metric
panel. PPL is shown as $\log_{10}$ of the unlearned/original ratio..}
\label{fig:heatmap-books-qwen}
\end{figure}

\begin{figure}[p]
\centering
\includegraphics[width=\textwidth]{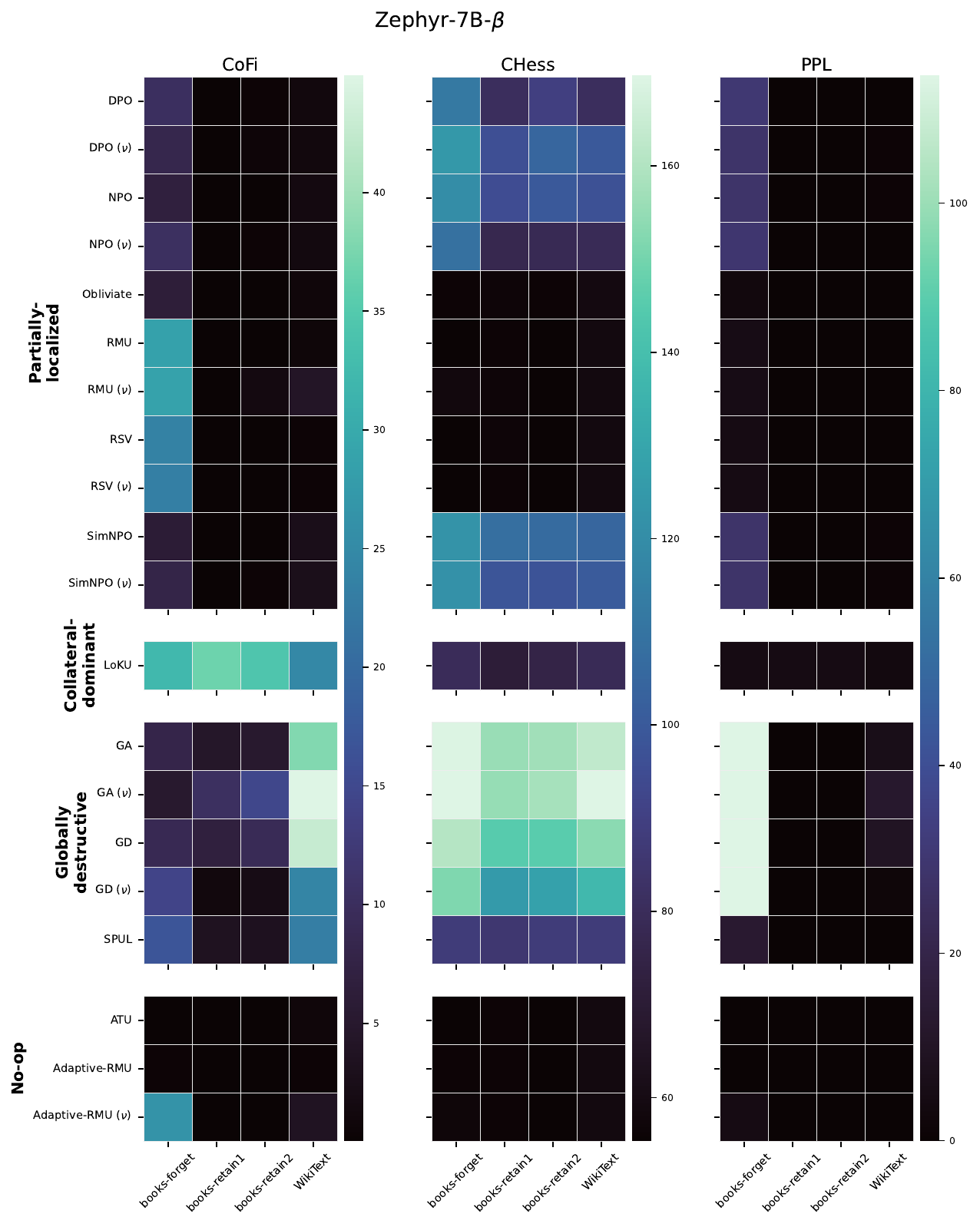}
\caption{MUSE-Books TRIAGE heatmap for Zephyr-7B-$\beta$. Rows are grouped by TRIAGE class;
columns are the four evaluation corpora. Colour scales are independent per metric
panel. PPL is shown as $\log_{10}$ of the unlearned/original ratio.}
\label{fig:heatmap-books-zephyr}
\end{figure}

\section{Additional Behavioral Evaluations}
\label{app:tbe}

This section reports the behavioural evaluations behind Sections~\ref{sec:agreement}
and~\ref{sec:relearning} in full: the matched tripartite accuracies for all four models,
the conventional WMDP/MMLU view for comparison with prior work, and the complete
relearning results. The main text uses Llama-3.1-8B throughout; the other three models
are here.

\subsection{Matched tripartite accuracies}

Tables~\ref{tab:tbe-llama} and~\ref{tab:tbe-qwen-zephyr} give raw accuracies on the same
$\mathcal{C}_F$, $\mathcal{C}_A$, $\mathcal{C}_G$ partition used by TRIAGE. Forget is
WMDP-Bio and WMDP-Cyber, adjacent-retain is the MMLU \texttt{virology},
\texttt{college\_biology} and \texttt{computer\_security} subsets, and general is the
remaining MMLU subsets. The evaluation is limited to a randomly selected subset of 40 questions from each set of questions to maximize computation efficiency.

\begin{table}[htbp]
\centering
\footnotesize
\setlength{\tabcolsep}{4pt}
\caption{Matched tripartite behavioural accuracies on WMDP for the two Llama models.
Forget: lower is more forgetting. Retain and general: higher is less collateral damage.
Methods are grouped by the TRIAGE class they occupy on that model, which is identical for
these two models.}
\label{tab:tbe-llama}
\begin{tabular}{l rrrrr @{\hspace{8pt}} rrrrr}
\toprule
& \multicolumn{5}{c}{\textbf{Llama-3.1-8B}} & \multicolumn{5}{c}{\textbf{Llama-3.2-3B}} \\
\cmidrule(lr){2-6}\cmidrule(lr){7-11}
\textbf{Method} & \textit{F-Bio} & \textit{R-Bio} & \textit{F-Cyb} & \textit{R-Cyb} & \textit{Gen}
& \textit{F-Bio} & \textit{R-Bio} & \textit{F-Cyb} & \textit{R-Cyb} & \textit{Gen} \\
\midrule
Base model   & 0.698 & 0.775 & 0.442 & 0.750 & 0.657 & 0.623 & 0.675 & 0.395 & 0.700 & 0.545 \\
\midrule
\multicolumn{11}{l}{\textit{No-op class}} \\
ATU          & 0.698 & 0.750 & 0.441 & 0.750 & 0.657 & 0.623 & 0.675 & 0.393 & 0.700 & 0.543 \\
Obliviate    & 0.700 & 0.775 & 0.433 & 0.750 & 0.659 & 0.616 & 0.650 & 0.399 & 0.750 & 0.544 \\
RSV          & 0.705 & 0.775 & 0.441 & 0.750 & 0.657 & 0.626 & 0.675 & 0.395 & 0.700 & 0.543 \\
\midrule
\multicolumn{11}{l}{\textit{Partially-localized class}} \\
RMU          & 0.329 & 0.600 & 0.259 & 0.500 & 0.647 & 0.372 & 0.575 & 0.280 & 0.550 & 0.546 \\
\midrule
\multicolumn{11}{l}{\textit{Collateral-dominant class}} \\
Adaptive-RMU & 0.269 & 0.475 & 0.256 & 0.300 & 0.614 & 0.263 & 0.475 & 0.251 & 0.150 & 0.494 \\
NPO          & 0.529 & 0.725 & 0.276 & 0.550 & 0.629 & 0.395 & 0.650 & 0.283 & 0.500 & 0.507 \\
SimNPO       & 0.502 & 0.675 & 0.273 & 0.500 & 0.608 & 0.399 & 0.675 & 0.283 & 0.550 & 0.513 \\
DPO          & 0.641 & 0.725 & 0.393 & 0.700 & 0.641 & 0.583 & 0.675 & 0.324 & 0.600 & 0.543 \\
\midrule
\multicolumn{11}{l}{\textit{Globally destructive class}} \\
LoKU+FILA    & 0.234 & 0.325 & 0.235 & 0.300 & 0.253 & 0.247 & 0.325 & 0.253 & 0.200 & 0.239 \\
GA           & 0.247 & 0.250 & 0.266 & 0.400 & 0.268 & 0.274 & 0.150 & 0.266 & 0.250 & 0.405 \\
GD           & 0.247 & 0.225 & 0.266 & 0.400 & 0.240 & 0.269 & 0.200 & 0.259 & 0.300 & 0.402 \\
SPUL         & 0.247 & 0.225 & 0.265 & 0.400 & 0.239 & 0.247 & 0.300 & 0.243 & 0.250 & 0.252 \\
\bottomrule
\end{tabular}
\end{table}

\begin{table}[htbp]
\centering
\footnotesize
\setlength{\tabcolsep}{4pt}
\caption{Matched tripartite behavioural accuracies on WMDP for Qwen3-32B and
Zephyr-7B-$\beta$. Conventions as in Table~\ref{tab:tbe-llama}. Methods are listed in a
fixed order here because the two models assign them to different classes; see
Table~\ref{tab:relearning-summary} for the per-model assignments. On Qwen3-32B, LoKU is
run in the FILA-free configuration.}
\label{tab:tbe-qwen-zephyr}
\begin{tabular}{l rrrrr @{\hspace{8pt}} rrrrr}
\toprule
& \multicolumn{5}{c}{\textbf{Qwen3-32B}} & \multicolumn{5}{c}{\textbf{Zephyr-7B-$\beta$}} \\
\cmidrule(lr){2-6}\cmidrule(lr){7-11}
\textbf{Method} & \textit{F-Bio} & \textit{R-Bio} & \textit{F-Cyb} & \textit{R-Cyb} & \textit{Gen}
& \textit{F-Bio} & \textit{R-Bio} & \textit{F-Cyb} & \textit{R-Cyb} & \textit{Gen} \\
\midrule
Base model   & 0.815 & 0.800 & 0.601 & 0.850 & 0.807 & 0.645 & 0.525 & 0.446 & 0.650 & 0.604 \\
\midrule
ATU          & 0.831 & 0.775 & 0.614 & 0.850 & 0.822 & 0.654 & 0.500 & 0.441 & 0.650 & 0.595 \\
Obliviate    & 0.829 & 0.775 & 0.614 & 0.850 & 0.821 & 0.637 & 0.525 & 0.392 & 0.750 & 0.598 \\
RSV          & 0.829 & 0.775 & 0.615 & 0.850 & 0.819 & 0.279 & 0.400 & 0.252 & 0.200 & 0.579 \\
RMU          & 0.830 & 0.775 & 0.608 & 0.850 & 0.819 & 0.274 & 0.350 & 0.255 & 0.400 & 0.519 \\
Adaptive-RMU & 0.745 & 0.725 & 0.565 & 0.850 & 0.823 & 0.639 & 0.500 & 0.426 & 0.600 & 0.606 \\
NPO          & 0.830 & 0.775 & 0.611 & 0.850 & 0.822 & 0.457 & 0.575 & 0.378 & 0.550 & 0.575 \\
SimNPO       & 0.830 & 0.775 & 0.612 & 0.850 & 0.821 & 0.472 & 0.575 & 0.382 & 0.600 & 0.584 \\
DPO          & 0.828 & 0.775 & 0.615 & 0.850 & 0.822 & 0.445 & 0.575 & 0.371 & 0.650 & 0.575 \\
LoKU         & 0.829 & 0.775 & 0.613 & 0.850 & 0.821 & 0.243 & 0.350 & 0.254 & 0.400 & 0.249 \\
GA           & 0.828 & 0.775 & 0.615 & 0.850 & 0.818 & 0.247 & 0.300 & 0.243 & 0.250 & 0.263 \\
GD           & 0.830 & 0.775 & 0.614 & 0.850 & 0.821 & 0.247 & 0.300 & 0.243 & 0.250 & 0.256 \\
SPUL         & 0.247 & 0.225 & 0.266 & 0.400 & 0.238 & 0.240 & 0.150 & 0.238 & 0.100 & 0.257 \\
\bottomrule
\end{tabular}
\end{table}

\paragraph{The two extreme classes agree with behaviour; the middle two do not.}
Across the four models, the no-op class contains 12 checkpoints and 11 of them sit within
roughly one point of the base model on every partition. The exception is Adaptive-RMU on
Qwen3-32B, which drops forget-bio from $0.815$ to $0.745$. The globally
destructive class contains 13 checkpoints and every one shows substantial general-utility
loss, reaching chance in 11 cases; the two exceptions, GA and GD on Llama-3.2-3B, still
fall 14 points to $0.405$ and $0.402$ from a base of $0.545$.

The middle two classes are where the two evaluations come apart, and Qwen3-32B is the
clearest case. Eleven of its twelve checkpoints leave WMDP accuracy at or slightly above
base, yet six of those eleven are collateral dominant with $\mathcal{C}_F$ shifts between
$1.88\%$ and $9.96\%$. Zephyr-7B-$\beta$ shows a milder version: Obliviate moves
$\mathcal{C}_F$ by $6.93\%$ and $\mathcal{C}_A$ by $8.90\%$ while leaving every
behavioural number within two points of base. We report these as divergences, not as
failures of either evaluation. Our data does not settle whether the behavioural
evaluation is missing an effect that would appear under a different probe, or whether
CoFi is registering parameter movement with no behavioural consequence on these
checkpoints. What the divergence does establish is that the two signals are not
substitutes.

\subsection{Conventional WMDP and MMLU accuracies}

Table~\ref{tab:behavioral-conventional} gives the standard forget/utility view for
comparison with prior work, where MMLU is used whole as a general-utility control rather
than split into adjacent and general partitions. It contains no information the matched
evaluation does not, but it is the format most unlearning papers report.

\begin{table}[htbp]
\centering
\footnotesize
\setlength{\tabcolsep}{3pt}
\caption{Conventional WMDP behavioural evaluation across all four models. WMDP-Bio and
WMDP-Cyber report MCQ accuracy on the unlearning targets (lower is more forgetting, $0.25$
is chance); MMLU is the undivided general-utility control (higher is better).}
\label{tab:behavioral-conventional}
\begin{tabular}{l rrr @{\hspace{6pt}} rrr @{\hspace{6pt}} rrr @{\hspace{6pt}} rrr}
\toprule
& \multicolumn{3}{c}{\textbf{Llama-3.1-8B}} & \multicolumn{3}{c}{\textbf{Llama-3.2-3B}}
& \multicolumn{3}{c}{\textbf{Qwen3-32B}} & \multicolumn{3}{c}{\textbf{Zephyr-7B-$\beta$}} \\
\cmidrule(lr){2-4}\cmidrule(lr){5-7}\cmidrule(lr){8-10}\cmidrule(lr){11-13}
\textbf{Method} & \textit{Bio} & \textit{Cyb} & \textit{MMLU} & \textit{Bio} & \textit{Cyb} & \textit{MMLU}
& \textit{Bio} & \textit{Cyb} & \textit{MMLU} & \textit{Bio} & \textit{Cyb} & \textit{MMLU} \\
\midrule
Base model   & 0.698 & 0.442 & 0.663 & 0.623 & 0.395 & 0.553 & 0.815 & 0.601 & 0.808 & 0.645 & 0.446 & 0.602 \\
\midrule
ATU          & 0.698 & 0.441 & 0.661 & 0.623 & 0.393 & 0.550 & 0.831 & 0.614 & 0.821 & 0.654 & 0.441 & 0.593 \\
Obliviate    & 0.700 & 0.433 & 0.665 & 0.616 & 0.399 & 0.551 & 0.829 & 0.614 & 0.820 & 0.637 & 0.392 & 0.598 \\
RSV          & 0.705 & 0.441 & 0.663 & 0.626 & 0.395 & 0.550 & 0.829 & 0.615 & 0.818 & 0.279 & 0.252 & 0.566 \\
RMU          & 0.329 & 0.259 & 0.643 & 0.372 & 0.280 & 0.547 & 0.830 & 0.608 & 0.818 & 0.274 & 0.255 & 0.511 \\
Adaptive-RMU & 0.269 & 0.256 & 0.604 & 0.263 & 0.251 & 0.488 & 0.745 & 0.565 & 0.820 & 0.639 & 0.426 & 0.602 \\
NPO          & 0.529 & 0.276 & 0.631 & 0.395 & 0.283 & 0.512 & 0.830 & 0.611 & 0.821 & 0.457 & 0.378 & 0.575 \\
SimNPO       & 0.502 & 0.273 & 0.609 & 0.399 & 0.283 & 0.519 & 0.830 & 0.612 & 0.820 & 0.472 & 0.382 & 0.584 \\
DPO          & 0.641 & 0.393 & 0.645 & 0.583 & 0.324 & 0.548 & 0.828 & 0.615 & 0.821 & 0.445 & 0.371 & 0.576 \\
LoKU+FILA    & 0.234 & 0.235 & 0.256 & 0.247 & 0.253 & 0.241 & 0.829 & 0.613 & 0.820 & 0.243 & 0.254 & 0.255 \\
GA           & 0.247 & 0.266 & 0.269 & 0.274 & 0.266 & 0.393 & 0.828 & 0.615 & 0.817 & 0.247 & 0.243 & 0.264 \\
GD           & 0.247 & 0.266 & 0.242 & 0.269 & 0.259 & 0.393 & 0.830 & 0.614 & 0.820 & 0.247 & 0.243 & 0.257 \\
SPUL         & 0.247 & 0.265 & 0.241 & 0.247 & 0.243 & 0.254 & 0.247 & 0.266 & 0.240 & 0.240 & 0.238 & 0.251 \\
\bottomrule
\end{tabular}
\end{table}

\subsection{Relearning attack}
\label{app:relearning}

Table~\ref{tab:relearning-summary} gives the recovery
$\Delta\mathrm{Acc}_F^{\mathrm{RL}}$ for every attacked checkpoint on all four models,
alongside its TRIAGE class, and Table~\ref{tab:relearning-tripartite} gives the full
per-partition accuracies for Llama-3.1-8B. SPUL was not attacked on any model. All
attacks use the identical budget specified in Appendix~\ref{app:hyperparams}.

\begin{table}[htbp]
\centering
\small
\setlength{\tabcolsep}{4pt}
\caption{Forget-partition recovery under the fixed-budget relearning attack,
$\Delta\mathrm{Acc}_F^{\mathrm{RL}}$, averaged over the bio and cyber splits, with the
TRIAGE class on each model. \textsc{n} = no-op, \textsc{p} = partially localized,
\textsc{c} = collateral dominant, \textsc{g} = globally destructive. Larger values mean
more of the forgotten content returned.}
\label{tab:relearning-summary}
\begin{tabular}{l cr cr cr cr}
\toprule
& \multicolumn{2}{c}{\textbf{Llama-3.1-8B}} & \multicolumn{2}{c}{\textbf{Llama-3.2-3B}}
& \multicolumn{2}{c}{\textbf{Qwen3-32B}} & \multicolumn{2}{c}{\textbf{Zephyr-7B-$\beta$}} \\
\cmidrule(lr){2-3}\cmidrule(lr){4-5}\cmidrule(lr){6-7}\cmidrule(lr){8-9}
\textbf{Method} & cls & $\Delta$ & cls & $\Delta$ & cls & $\Delta$ & cls & $\Delta$ \\
\midrule
ATU          & \textsc{n} & $0.0028$  & \textsc{n} & $0.0052$  & \textsc{n} & $-0.0015$ & \textsc{n} & $-0.0038$ \\
Obliviate    & \textsc{n} & $0.0056$  & \textsc{n} & $0.0059$  & \textsc{n} & $0.0006$  & \textsc{c} & $0.0362$ \\
RSV          & \textsc{n} & $-0.0017$ & \textsc{n} & $0.0059$  & \textsc{n} & $-0.0004$ & \textsc{p} & $0.2688$ \\
RMU          & \textsc{p} & $0.2738$  & \textsc{p} & $0.1810$  & \textsc{n} & $0.0012$  & \textsc{c} & $0.2778$ \\
Adaptive-RMU & \textsc{c} & $0.3020$  & \textsc{c} & $0.2451$  & \textsc{n} & $0.0604$  & \textsc{p} & $0.0121$ \\
NPO          & \textsc{c} & $0.1594$  & \textsc{c} & $0.1540$  & \textsc{c} & $0.0014$  & \textsc{c} & $0.0916$ \\
SimNPO       & \textsc{c} & $0.1732$  & \textsc{c} & $0.1352$  & \textsc{c} & $0.0019$  & \textsc{c} & $0.1107$ \\
DPO          & \textsc{c} & $0.0586$  & \textsc{c} & $0.0495$  & \textsc{c} & $0.0012$  & \textsc{c} & $0.1330$ \\
LoKU+FILA    & \textsc{g} & $0.0116$  & \textsc{g} & $-0.0016$ & \textsc{c} & $0.0013$  & \textsc{g} & $0.0014$ \\
GA           & \textsc{g} & $0.0000$  & \textsc{g} & $0.0367$  & \textsc{c} & $0.0025$  & \textsc{g} & $0.1652$ \\
GD           & \textsc{g} & $0.0015$  & \textsc{g} & $-0.0067$ & \textsc{c} & $0.0009$  & \textsc{g} & $0.0028$ \\
\bottomrule
\end{tabular}
\end{table}

\begin{table}[htbp]
\centering
\small
\setlength{\tabcolsep}{4pt}
\caption{Llama-3.1-8B WMDP, post-unlearning $\rightarrow$ post-relearning accuracy on all
five partitions. Recovery on the forget partitions is the attack succeeding; the retain
and general partitions should stay roughly flat.}
\label{tab:relearning-tripartite}
\begin{tabular}{l ccccc}
\toprule
\textbf{Method} & \textit{Forget-Bio} & \textit{Retain-Bio} & \textit{Forget-Cyber}
& \textit{Retain-Cyber} & \textit{General} \\
\midrule
ATU          & $0.698\!\to\!0.705$ & $0.750\!\to\!0.800$ & $0.441\!\to\!0.440$ & $0.750\!\to\!0.750$ & $0.657\!\to\!0.665$ \\
Obliviate    & $0.700\!\to\!0.707$ & $0.775\!\to\!0.800$ & $0.433\!\to\!0.437$ & $0.750\!\to\!0.750$ & $0.659\!\to\!0.667$ \\
RSV          & $0.705\!\to\!0.705$ & $0.775\!\to\!0.800$ & $0.441\!\to\!0.437$ & $0.750\!\to\!0.750$ & $0.657\!\to\!0.664$ \\
\midrule
RMU          & $0.329\!\to\!0.694$ & $0.600\!\to\!0.775$ & $0.259\!\to\!0.441$ & $0.500\!\to\!0.750$ & $0.647\!\to\!0.646$ \\
\midrule
Adaptive-RMU & $0.269\!\to\!0.687$ & $0.475\!\to\!0.775$ & $0.256\!\to\!0.442$ & $0.300\!\to\!0.750$ & $0.614\!\to\!0.646$ \\
NPO          & $0.529\!\to\!0.694$ & $0.725\!\to\!0.775$ & $0.276\!\to\!0.429$ & $0.550\!\to\!0.750$ & $0.629\!\to\!0.647$ \\
SimNPO       & $0.502\!\to\!0.697$ & $0.675\!\to\!0.750$ & $0.273\!\to\!0.425$ & $0.500\!\to\!0.750$ & $0.608\!\to\!0.651$ \\
DPO          & $0.641\!\to\!0.698$ & $0.725\!\to\!0.750$ & $0.393\!\to\!0.453$ & $0.700\!\to\!0.800$ & $0.641\!\to\!0.657$ \\
\midrule
LoKU+FILA    & $0.234\!\to\!0.248$ & $0.325\!\to\!0.275$ & $0.235\!\to\!0.244$ & $0.300\!\to\!0.250$ & $0.253\!\to\!0.259$ \\
GA           & $0.247\!\to\!0.247$ & $0.250\!\to\!0.250$ & $0.266\!\to\!0.266$ & $0.400\!\to\!0.400$ & $0.268\!\to\!0.287$ \\
GD           & $0.247\!\to\!0.272$ & $0.225\!\to\!0.250$ & $0.266\!\to\!0.244$ & $0.400\!\to\!0.200$ & $0.240\!\to\!0.257$ \\
\bottomrule
\end{tabular}
\end{table}

\paragraph{Recovery concentrates in the structurally active, non-collapsed checkpoints.}
Sixteen checkpoints recover more than $0.05$ across the four models, and 14 of them are
partially localized or collateral dominant. The no-op checkpoints recover essentially
nothing anywhere, which is the consistency check we would expect when almost nothing was
changed. The pattern that mattered on Llama-3.1-8B repeats on Llama-3.2-3B and
Zephyr-7B-$\beta$: the methods that look best behaviourally before the attack give the
most back after it. RMU and RSV on Zephyr are the extreme case, recovering $0.2778$ and
$0.2688$ from near-chance forget accuracy.

Two exceptions are worth stating. Adaptive-RMU on Qwen3-32B
recovers $0.0604$ despite being classified a no-op, which is consistent with it being the
one no-op checkpoint that moved behaviourally at all; it sits at the boundary of the class
on this model and we would not defend the assignment strongly. More importantly, GA on
Zephyr-7B-$\beta$ recovers $0.1652$ while being globally destructive, and its general
accuracy rises from $0.2630$ to $0.5454$ under the attack. The relearning fine-tuning
partially repaired a collapsed model rather than restoring targeted knowledge. This is a
counterexample to reading low recovery in the globally destructive class as a general
property: on Llama-3.1-8B, where Section~\ref{sec:relearning} makes that argument, all
three destructive checkpoints stay collapsed and recover at most $0.0116$, but the
mechanism is not guaranteed to hold on a more fragile base model. GD and LoKU+FILA on
Zephyr do stay collapsed, so GA is the single case in our data.

\subsection{Partition sizes}
See Table~\ref{tab:adjacent-probe}

\begin{table}[htbp]
\centering
\small
\setlength{\tabcolsep}{5pt}
\caption{Per-question results on the full adjacent probe for Llama-3.1-8B, as
correct/total with accuracy in parentheses. The forget partitions are large; the adjacent
partitions are not, which bounds how finely $\delta_A$ can be read.}
\label{tab:adjacent-probe}
\begin{tabular}{l ccccc}
\toprule
\textbf{Method} & \textit{wmdp\_bio} & \textit{college\_biology} & \textit{virology}
& \textit{wmdp\_cyber} & \textit{computer\_security} \\
\midrule
ATU          & 888/1273 (0.70) & 111/144 (0.77) & 88/166 (0.53) & 877/1987 (0.44) & 82/100 (0.82) \\
Obliviate    & 891/1273 (0.70) & 113/144 (0.78) & 88/166 (0.53) & 861/1987 (0.43) & 79/100 (0.79) \\
RSV          & 897/1273 (0.70) & 112/144 (0.78) & 90/166 (0.54) & 877/1987 (0.44) & 80/100 (0.80) \\
RMU          & 419/1273 (0.33) & 107/144 (0.74) & 81/166 (0.49) & 514/1987 (0.26) & 58/100 (0.58) \\
Adaptive-RMU & 343/1273 (0.27) & 102/144 (0.71) & 52/166 (0.31) & 508/1987 (0.26) & 32/100 (0.32) \\
NPO          & 673/1273 (0.53) & 110/144 (0.76) & 89/166 (0.54) & 549/1987 (0.28) & 58/100 (0.58) \\
SimNPO       & 639/1273 (0.50) & 110/144 (0.76) & 84/166 (0.51) & 543/1987 (0.27) & 48/100 (0.48) \\
DPO          & 816/1273 (0.64) & 108/144 (0.75) & 88/166 (0.53) & 781/1987 (0.39) & 70/100 (0.70) \\
LoKU+FILA    & 298/1273 (0.23) &  38/144 (0.26) & 53/166 (0.32) & 467/1987 (0.24) & 29/100 (0.29) \\
GA           & 314/1273 (0.25) &  40/144 (0.28) & 47/166 (0.28) & 528/1987 (0.27) & 28/100 (0.28) \\
GD           & 314/1273 (0.25) &  37/144 (0.26) & 47/166 (0.28) & 528/1987 (0.27) & 28/100 (0.28) \\
SPUL         & 313/1273 (0.25) &  37/144 (0.26) & 46/166 (0.28) & 529/1987 (0.27) & 29/100 (0.29) \\
\bottomrule
\end{tabular}
\end{table}

\section{Structural Comparison Frameworks}
\label{app:structural_comparisons}

Table~\ref{tab:comparison} details the comparative reach of current structural evaluations discussed in Section~\ref{sec:comparison}. TRIAGE extends existing protocols by introducing multi-axis controls that capture spatial footprint localization, ensuring that destructively scaled updates are not falsely categorized as highly robust unlearning operations.

\begin{table}[htbp]
\centering
\footnotesize
\setlength{\tabcolsep}{4pt}
\caption{Empirical reach of structural unlearning evaluations as reported in their published forms: ConceptVectors~\citep{hong2025intrinsic}, Reversibility~\citep{xu2025unlearning}, Tamper-Resistance~\citep{siddiqui2026dormant}. ``Y'' indicates a capability supported by the framework's evaluation protocol as published; ``---'' indicates the capability is outside the framework's evaluation scope as currently structured. The comparison concerns what each framework's published evaluation surfaces, not in-principle limits.}
\label{tab:comparison}
\begin{tabular}{l c c c c}
\toprule
& \textbf{ConceptVec.} & \textbf{Revers.} & \textbf{Tamper-R.} & \textbf{TRIAGE} \\
\midrule
Concept-level granularity                  & Y    & Y    & ---  & Y \\
Model-wide parameter diagnostics           & ---  & ---  & Y    & Y \\
Adjacent-retain control alongside forget   & ---  & ---  & ---  & Y \\
Generic-retain control alongside forget    & Y    & Y    & ---  & Y \\
Fluency / utility readings on retain content& ---  & ---  & ---  & Y \\
Single-pass evaluation (no relearning attack)& Y    & Y    & ---  & Y \\
Algorithm-agnostic (no per-method instrumentation) & ---  & Y    & Y    & Y \\
\bottomrule
\end{tabular}
\end{table}

\end{document}